\documentclass[letterpaper]{article}
\usepackage[T1]{fontenc}
\usepackage{geometry}
\usepackage{setspace}
\usepackage{xurl}
\usepackage{colortbl}
\usepackage{threeparttable}
\usepackage{float}
\usepackage{adjustbox}
\usepackage{placeins}
\usepackage{multirow}
\usepackage{booktabs}
\usepackage{hyperref}
\usepackage{pifont}
\usepackage{longtable}
\usepackage{seqsplit}
\usepackage{rotating}

\usepackage{pifont}
\usepackage{xcolor}

\definecolor{checkgreen}{RGB}{0,145,145}
\definecolor{crossred}{RGB}{255,0,0}

\newcommand{\cmark}{\textcolor{checkgreen}{\ding{51}}}
\newcommand{\xmark}{\textcolor{crossred}{\ding{55}}}

\usepackage{caption}
\usepackage{array}
\usepackage{tabularx}
\usepackage{xcolor}
\definecolor{VAEbg}{HTML}{EAF2FB}
\definecolor{GANbg}{HTML}{FCEDEA}
\definecolor{Flowbg}{HTML}{ECF5EC}
\definecolor{Diffbg}{HTML}{FFF4D8}
\usepackage[edges]{forest}
\usetikzlibrary{shadows}
\definecolor{nodeblue}{RGB}{135,180,220}
\definecolor{nodebluedeep}{RGB}{100,150,200}
\definecolor{boxbg}{RGB}{245,250,255}
\definecolor{lineblue}{RGB}{70,130,180}
\tikzset{
	mainnode/.style={
		rounded corners=6pt,
		draw=lineblue!80,
		top color=nodeblue,
		bottom color=nodebluedeep!70,
		text=black,
		font=\bfseries\small,
		inner sep=6pt,
		align=center,
		minimum height=3.25ex,
		drop shadow={opacity=0.25, shadow xshift=1pt, shadow yshift=-1pt},
	},
	contentbox/.style={
		rounded corners=5pt,
		draw=lineblue,
		fill=boxbg,
		align=left,
		font=\footnotesize,
		inner sep=7pt,
		text=black,
	},
}
\usepackage{amsmath}
\usepackage{amssymb}
\usepackage[
    backend=biber,
    style=ieee,
    sorting=none,
    doi=true,
    url=false,
    isbn=false,
    maxbibnames=10,
    minbibnames=10
]{biblatex}

\DefineBibliographyStrings{english}{
  andothers = {et\addabbrvspace al\adddot}
}

\renewbibmacro*{volume+number+eid}{
  \printfield{volume}
  \iffieldundef{number}
    {}
    {\printtext[parens]{\printfield{number}}}
  \setunit{\addcomma\space}
  \printfield{eid}
}

\usepackage{xpatch}
\xpatchbibdriver{inproceedings}
  {\usebibmacro{byauthor}
   \setunit{\addspace}
   \usebibmacro{in:}}
  {\usebibmacro{byauthor}
   \newunit\newblock
   \usebibmacro{title}
	\setunit{\addperiod\space}
   \usebibmacro{in:}}
  {}
  {\typeout{Failed to patch inproceedings driver}}

\DeclareFieldFormat[inproceedings]{title}{#1}
\usepackage{graphicx}
\newfloat{scheme}{htbp}{los}
\floatname{scheme}{Scheme}
\floatname{chart}{Chart}
\newfloat{graph}{htbp}{loh}
\usepackage{chemformula}
\usepackage[version = 4]{mhchem}
\usepackage{authblk}
\author[1]{Xinrui Xu}
\author[1]{Xueer Wang}
\author[2]{Dan Luo*}
\author[3]{Sisi Yuan*}
\author[1]{Xuan Lin*}
\affil[1]{School of Computer Science, Xiangtan University, Xiangtan 411105, China}
\affil[2]{College of Computer Science and Electronic Engineering, Hunan University, Changsha 410082, China}
\affil[3]{School of Chinese Medicine, Hong Kong Baptist University,
15 Baptist University Road, Kowloon Tong 999077, Kowloon, Hong Kong SAR, China}

\title{A Systematic Evaluation of Molecule Generation Models for \textit{De Novo} Drug Design: From Benchmarks to Practical Insights}
\date{*Email: \texttt{dluo.ac@gmail.com}; \texttt{sisiyuan@hkbu.edu.hk}; \texttt{jack\_lin@xtu.edu.cn}}

\begin{document}
	\maketitle
	\begin{abstract}
	Molecule generation has emerged as a powerful computational tool for \textit{de novo} drug design, enabling the exploration of chemical space beyond the limits of conventional virtual screening. The field has progressed rapidly, driven by advances in molecular representations, generative architectures, and target-aware modeling strategies. However, existing reviews typically address specific model families or application scenarios in isolation, rather than offering an integrated perspective on how these components collectively form a coherent generation workflow. In this review, we present a comprehensive evaluation of molecule generation models for \textit{de novo} drug design, covering 82 methods across five deep generative frameworks, including recurrent neural network (RNN)- and Transformer-based models, variational autoencoders (VAEs), generative adversarial networks (GANs), flow-based models, and diffusion models. We first summarize widely used benchmarks and molecular representations, and then examine the methodological principles underlying both general and pocket-conditioned generation. A central contribution of this work is a systematic synthesis and comparative analysis of reported performance across commonly used benchmarks and evaluation metrics. We also summarize representative experimentally validated case studies. Looking ahead, we discuss future directions in standardized 3D data, interaction-aware generation, receptor flexibility, and multi-objective molecular design, with the aim of improving the reliability and experimental relevance of molecule generation. All collected benchmark resources, evaluation metrics, and model references are provided in a publicly accessible repository at \url{https://github.com/JacklinGroup/molecule-generation-review}.
	\end{abstract}

	\section*{Keywords}
	molecule generation, deep generative models, \textit{de novo} drug design, molecular representation

	\section{Introduction}
	
	The central challenge of drug discovery lies in identifying novel bioactive molecules within the vast chemical space, which is estimated to contain up to $10^{60}$ molecules~\cite{reymond2015chemical}. Traditional virtual screening strategies rely heavily on predefined compound libraries and high-throughput computational screening, which can be time- and resource-intensive and restrict molecular exploration to available compounds and their close analogues. These limitations have motivated the development of \textit{de novo} drug design, which aims to construct novel molecules from scratch rather than select candidates from existing libraries. Molecule generation has therefore become a central computational step in \textit{de novo} drug design, enabling models to more effectively learn molecular distributions and propose chemically valid and diverse structures for subsequent evaluation, screening, and candidate prioritization.

\begin{figure*}[!t]
		\centering
		\includegraphics[width=1\linewidth]{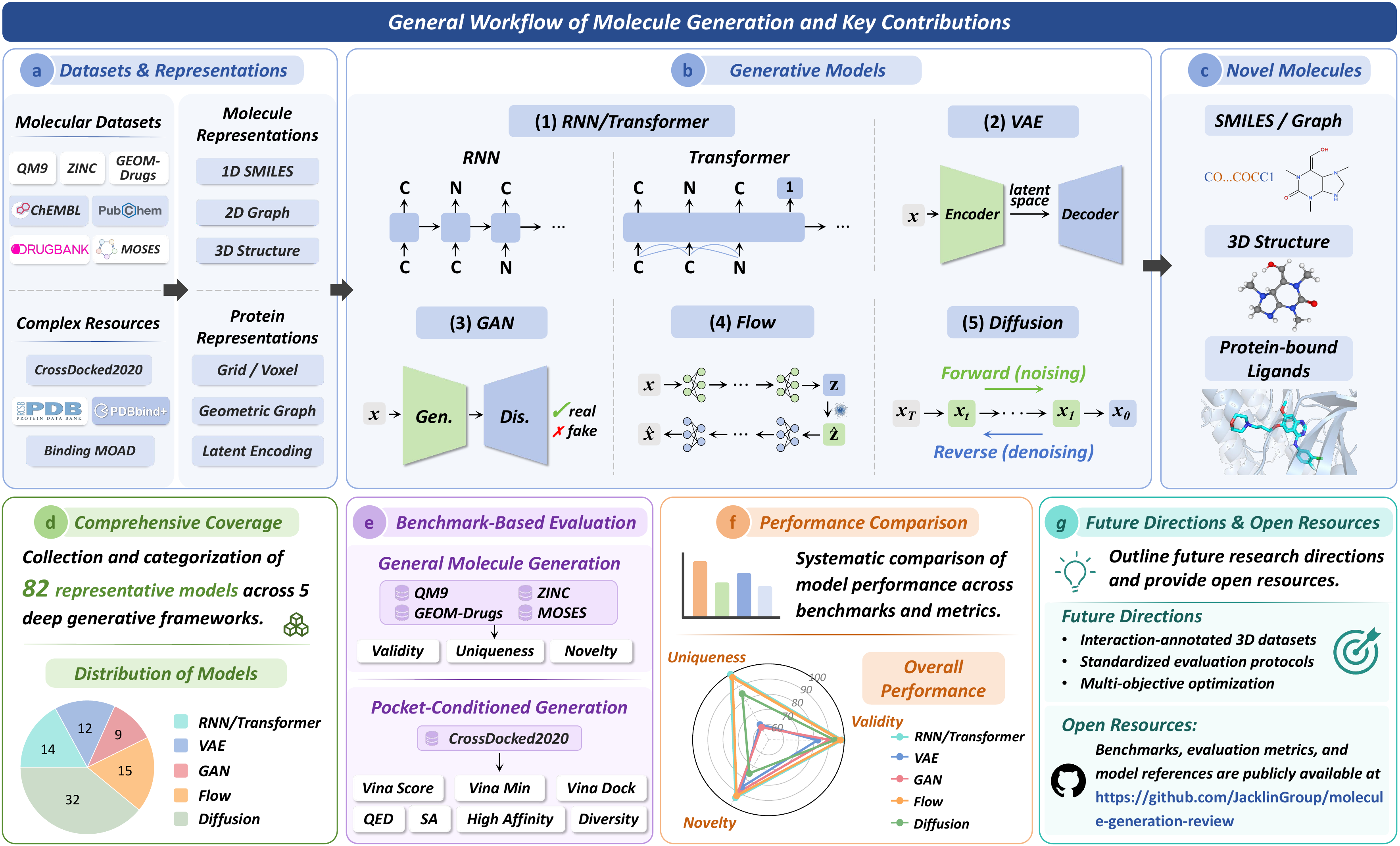}
		\caption{
			Overview of the molecule generation workflow and key contributions of this review.
			(a) Molecular and protein structure data are curated from public databases and encoded into suitable representations, including 1D SMILES sequences, 2D molecular graphs, and 3D geometric structures for molecules, as well as grid- or voxel-based, graph-based, and latent encodings for proteins.
			(b) These representations are then fed into five major classes of generative models which learn the data distribution through distinct mechanisms (e.g., autoregressive sequence modeling, latent-space sampling, adversarial training, invertible transformations, or iterative denoising).
			(c) The trained models generate three types of molecular outputs: SMILES- or graph-based molecules, 3D molecular structures, and protein-bound ligands.
			(d) This review collects and categorizes 82 representative molecule generation models across five major deep generative frameworks.
			(e) Benchmark-based evaluations are organized for both general molecule generation and pocket-conditioned generation tasks using widely adopted datasets and metrics.
			(f) Model performance is systematically compared across benchmarks and evaluation metrics.
			(g) Future research directions are outlined, and benchmarks, evaluation metrics, and model references are made publicly available through an open repository.
		}
		\label{fig:generalpipeline}
	\end{figure*}

	Molecule generation can be organized as a workflow. As shown in Figure~\ref{fig:generalpipeline}, the workflow begins with chemical and structural datasets~\cite{wu2018moleculenet}, from which molecular data and target-related structural information are collected and curated. Molecules can be represented as sequences, molecular graphs, or structures, whereas protein environments can be encoded using grids or voxels, geometric graphs, latent embeddings, and other structural representations~\cite{yang2019analyzing}. These representations are subsequently processed by major deep generative frameworks, including recurrent neural network (RNN)/Transformer-based, variational autoencoder (VAE)-based, generative adversarial network (GAN)-based, flow-based, and diffusion-based models, to learn molecular distributions and generate novel candidate molecules, with pocket-conditioned variants further incorporating protein-pocket information to guide target-specific generation. The generated molecules are then evaluated in terms of chemical validity, novelty, drug-likeness, synthesizability, and, where applicable, target relevance, thereby providing a pool of screenable and computationally tractable candidates for subsequent analysis and prioritization.
    Existing reviews have summarized molecule generation from specific perspectives, including 3D generation, diffusion models, and graph-based generative models~\cite{pang2023deep,sousa2021generative,alakhdar2024diffusion,zhang2025graph}. For example, 3D deep generative models have been reviewed mainly from the perspectives of 3D molecular representations, generation strategies, and associated challenges~\cite{xie2022advances}, while geometric deep learning for structure-based drug design has been surveyed with emphasis on molecular property prediction, ligand binding-site prediction, pose prediction, and structure-based drug design~\cite{isert2023structure}.
	However, most existing reviews organize the field around individual model families or specific application scenarios, providing limited integration across the overall molecule generation workflow. In particular, relatively few studies connect model taxonomy with benchmark frameworks, protocol-aware performance interpretation, and evidence from practical drug design applications. In contrast, this review integrates these aspects to provide a systematic evaluation of molecule generation models from methodological foundations to benchmark performance and practical implications.

	To address this gap, this review is organized around three successive questions: what molecular information generative models learn and how it is represented, how generative frameworks and pocket-conditioned strategies construct novel molecules, and how reported performance and practical utility should be evaluated and interpreted across benchmarks. Accordingly, Section 2 summarizes representative benchmark datasets and evaluation frameworks, while Section 3 reviews molecular and protein-pocket representations. Section 4 introduces RNN/Transformer-based, VAE-based, GAN-based, flow-based, and diffusion-based frameworks, followed by their training objectives and pocket-conditioned extensions. Section 5 presents a comparative performance analysis by first defining commonly used evaluation metrics and then examining reported results on QM9, ZINC, GEOM-Drugs, MOSES, and CrossDocked2020. Section 6 discusses current limitations and future directions. To summarize, the main contributions of this work are as follows:
	\begin{itemize}
		\item[(i)] \textit{Model taxonomy.} We provide a systematic taxonomy covering 82 representative molecule generation models across five generative frameworks, offering a structured entry point for researchers navigating this rapidly expanding field.
		\item[(ii)] \textit{Benchmark-based performance synthesis.} We systematically compile and compare literature-reported results across commonly used benchmarks and evaluation metrics. For pocket-conditioned generation, we further summarize the docking evaluation protocols used across representative models to facilitate careful interpretation of reported docking performance.
		\item[(iii)] \textit{From in silico to in vitro: a curated collection of experimentally validated case studies.} While most reviews focus exclusively on computational metrics, we bridge the gap between virtual design and experimental practice by compiling eight representative case studies in which computationally generated molecules have been synthesized and experimentally tested (Table~\ref{tbl:representative_case_studies}). These cases span diverse targets and diseases, and we analyze them along three axes: (a) the generative strategy employed; (b) the approach and level of experimental validation; and (c) lessons learned that can be drawn regarding the translation of computationally generated molecules into experimentally validated candidates. This analysis provides a practical perspective on the extent to which current molecular generative methods have progressed beyond computational benchmarking toward experimental evaluation.
		\item[(iv)] \textit{Abundant resources.} We have compiled an organized collection of resources related to the reviewed models, including links to benchmark datasets, original papers, and available code repositories. These resources can be accessed at \href{https://github.com/JacklinGroup/molecule-generation-review}{https://github.com/JacklinGroup/molecule-generation-review}, which will be continuously updated.
		\item[(v)] \textit{Future directions.} We discuss the limitations of existing studies and suggest future directions toward more reliable and decision-oriented molecule generation.
	\end{itemize}

	\section{Benchmarking and Evaluation of Molecule Generation}

	Benchmarking molecular generative models requires both representative datasets and evaluation protocols that capture different aspects of model performance. This section reviews commonly used benchmark datasets and evaluation frameworks for molecule generation, optimization, 3D structural quality, and structure-based generation.
	
	\subsection{Benchmark Datasets}
	
	Datasets provide the empirical basis for training, validating, and comparing molecule generation models. As summarized in Table~\ref{tbl:dataset_summary}, existing resources cover different levels of molecular information, ranging from ligand-only molecular corpora and quantum-chemical datasets to conformer ensembles, protein--ligand complexes, structural biology repositories, and drug-centered databases. These datasets determine what molecular, geometric, biological, and pharmacological information can be learned, and they also influence the choice of generation tasks, evaluation metrics, and generalization settings in molecule generation.

	\noindent\textbf{QM9.} QM9 is a quantum-chemical dataset of approximately 134,000 small neutral organic molecules containing C, H, N, O, and F, with SMILES strings, density-functional-theory-optimized equilibrium geometries, and computed geometric, energetic, electronic, and thermodynamic properties~\cite{ramakrishnan2014quantum}. It is widely used for molecular property prediction and as a benchmark for two- and three-dimensional molecule generation, including the evaluation of the chemical validity and geometric quality of generated small molecules.

	\noindent\textbf{ZINC-22.} ZINC-22 is a free, multi-billion-scale database of commercially available compounds, searchable in two-dimensional form, with a large subset prepared in biologically relevant, ready-to-dock three-dimensional formats~\cite{tingle2023zinc}. ZINC-derived subsets are widely used for virtual screening, ligand-centric molecule generation, molecular optimization, and exploration of drug-like chemical space.

	\noindent\textbf{GEOM-Drugs.} GEOM-Drugs contains drug-like molecules represented by molecular graphs and ensembles of three-dimensional conformers annotated with energies and statistical weights~\cite{axelrod2022geom}. It is mainly used for conformer generation, ensemble-aware property prediction, three-dimensional molecule generation, geometry modeling, and conformational diversity analysis.

	\begin{table*}[!t]
		\centering
		\footnotesize
		\caption{Summary of Representative Data Resources and Benchmarks for Molecule Generation. Data were collected in May 2026.}
		\label{tbl:dataset_summary}
		\begin{tabular}{l l l c l}
			\hline
			Dataset & Version & Entries & Latest Update & Link \\
			\hline
			QM9~\cite{ramakrishnan2014quantum} & -- & 133,885 & 2014 & \href{https://quantum-machine.org/datasets/}{link} \\
			ZINC-22~\cite{tingle2023zinc} & v22 & 54,900,000,000 & 2024 & \href{https://cartblanche.docking.org/}{link} \\
			GEOM-Drugs~\cite{axelrod2022geom} & -- & 304,466 & 2022 & \href{https://github.com/learningmatter-mit/geom}{link} \\
			ChEMBL~\cite{zdrazil2024chembl,davies2015chembl} & v37 & 2,921,148 & 2026 & \href{https://www.ebi.ac.uk/chembl/}{link} \\
			PubChem~\cite{kim2025pubchem} & -- & 123,000,000 & 2026 & \href{https://pubchem.ncbi.nlm.nih.gov}{link} \\
			DrugBank~\cite{knox2024drugbank} & v6.0 & 19,872 & 2026 & \href{https://go.drugbank.com}{link} \\
			MOSES~\cite{polykovskiy2020molecular} & -- & 1,936,962 & 2020 & \href{https://github.com/molecularsets/moses}{link} \\
			CrossDocked2020~\cite{francoeur2020three} & v1.3 & 22,500,000 & 2024 & \href{https://bits.csb.pitt.edu/files/crossdock2020/}{link} \\
			RCSB PDB~\cite{10.1093/nar/gkae1091} & -- & 254,978 & 2026 & \href{https://www.rcsb.org}{link} \\
			PDBbind~\cite{liu2017forging} & v2020.R1 & 29,001 & 2020 & \href{https://www.pdbbind-plus.org.cn}{link} \\
			Binding MOAD~\cite{wagle2023sunsetting} & -- & 41,409 & 2023 & \href{https://www.bindingmoad.org/}{link} \\
			\hline
		\end{tabular}
	\end{table*}
	\noindent\textbf{ChEMBL.} ChEMBL is a manually curated bioactivity database that links molecular structures, biological targets, assay records, and quantitative activity measurements~\cite{davies2015chembl,zdrazil2024chembl}. It is widely used for bioactivity prediction, quantitative structure--activity relationship modeling, target--ligand modeling, and bioactivity-guided molecule generation and optimization.

	\noindent\textbf{PubChem.} PubChem is a large, integrated public chemical database containing standardized compound structures, deposited substances, identifiers, physicochemical annotations, literature and patent information, and bioassay data~\cite{kim2025pubchem}. It is commonly used for large-scale molecular pretraining and representation learning, property and bioactivity prediction, chemical similarity and substructure searches, and the construction of molecular datasets at scale.

	\noindent\textbf{DrugBank.} DrugBank is a drug-centered knowledgebase integrating molecular structures, approved and investigational drugs, biological targets, mechanisms of action, pharmacological annotations, and drug interaction information~\cite{knox2024drugbank}. It is mainly used for drug repurposing, drug--target and drug--drug interaction prediction, pharmacological and mechanism-of-action modeling, and reference-based analysis of approved and clinically investigated compounds.

	\noindent\textbf{MOSES.} MOSES is a SMILES-based molecule generation benchmark derived from the ZINC Clean Leads collection and designed to standardize the training and evaluation of generative models using unified datasets and metrics~\cite{polykovskiy2020molecular}. It is mainly used in practice for ligand-centric distribution learning, validity, novelty, and diversity evaluation, distribution-similarity assessment, and evaluation of the ability to generate scaffolds absent from the training set.

	\noindent\textbf{CrossDocked2020.} CrossDocked2020 is a structure-based machine-learning dataset containing protein structures and approximately 22.5 million docked ligand poses generated across structurally related binding sites~\cite{francoeur2020three}. In pocket-conditioned molecule generation, studies commonly use filtered subsets that retain high-quality docked poses and separate training and test pockets according to protein-similarity criteria. These subsets support pocket-conditioned generation, docking-based evaluation, and assessment of generalization to unseen or dissimilar binding pockets.

	\noindent\textbf{RCSB PDB.} RCSB PDB provides access to experimentally determined three-dimensional structures of proteins, nucleic acids, and other biological macromolecules and their complexes, together with atomic coordinates and associated structural and experimental metadata~\cite{10.1093/nar/gkae1091}. It is commonly used to obtain protein structures, ligand-bound complexes, binding pockets, and structural templates for protein-conditioned generation, molecular docking, and structure-based modeling.

	\noindent\textbf{PDBbind.} PDBbind collects three-dimensional biomolecular complex structures from the PDB and links them to experimentally measured binding-affinity data~\cite{liu2017forging}. It is widely used for binding-affinity prediction, scoring-function development, docking and pose scoring, and protein--ligand interaction modeling.

	\noindent\textbf{Binding MOAD.} Binding MOAD is a curated database of protein--ligand complex structures with biologically relevant ligands and experimentally measured binding affinities when available, and is now maintained as a final archived resource~\cite{wagle2023sunsetting}. It has been used as a source of curated complexes for molecular docking, protein--ligand recognition, binding-site analysis, affinity modeling, and the training or evaluation of structure-based molecule generation methods.

\begin{figure*}[!t]
		\centering
		\includegraphics[width=\textwidth]{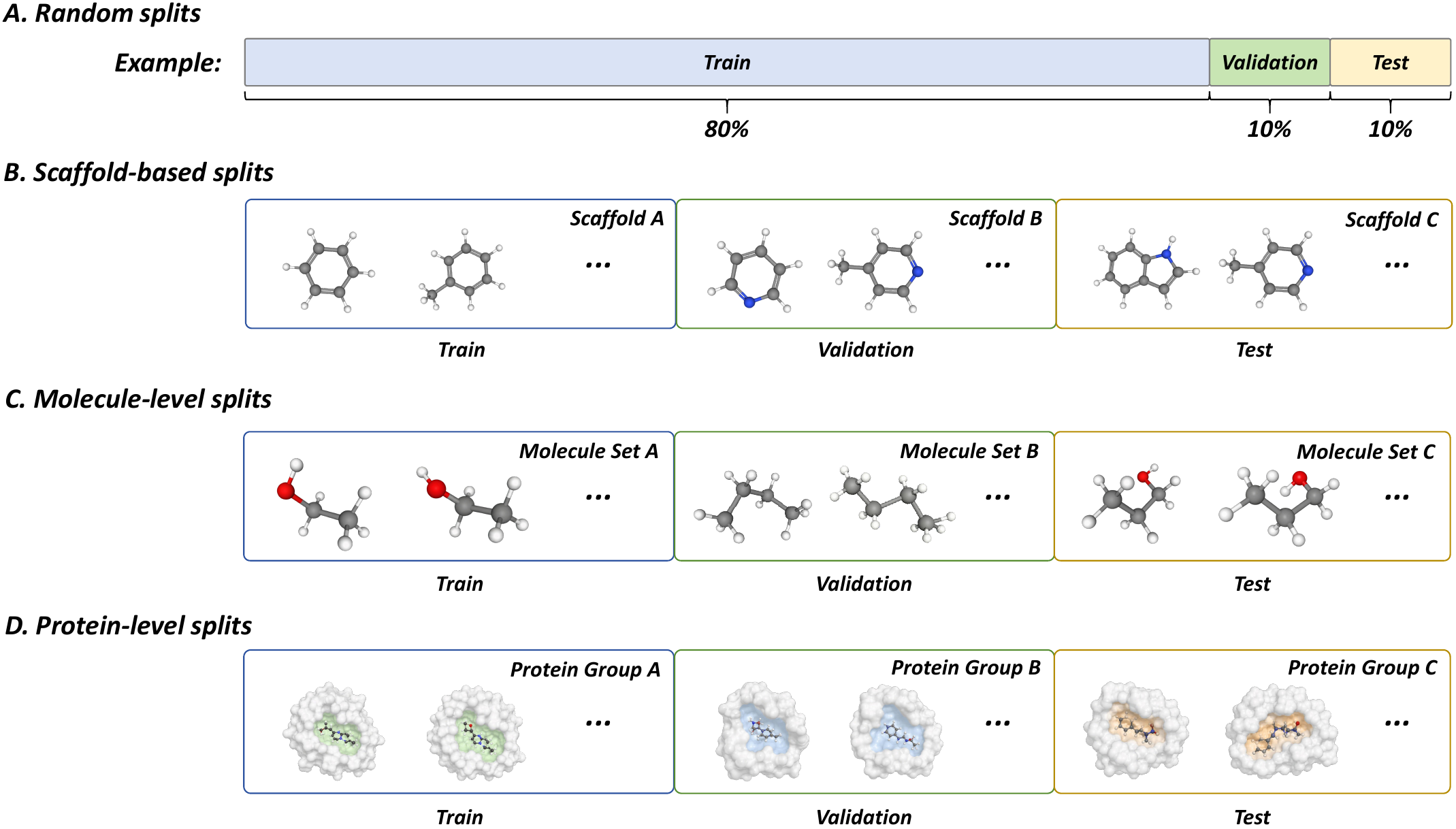}
		\caption{
		Representative dataset splitting strategies used in molecule generation, including (a) random splits, (b) scaffold-based splits, (c) molecule-level splits, and (d) protein-level splits.
		}
		\label{fig:data_split_strategies}
	\end{figure*}

	\noindent\textbf{Dataset splitting protocols.} Common dataset splitting strategies used in molecule generation are illustrated in Figure~\ref{fig:data_split_strategies}. Specifically, (a) random splits, such as an 80:10:10 division into training, validation, and test sets, are widely used for model training and evaluation~\cite{wu2018moleculenet}. (b) Scaffold-based splits separate molecules according to their molecular scaffolds and are commonly used to assess generalization to unseen chemotypes~\cite{polykovskiy2020molecular}. (c) Molecule-level splits assign different molecules to separate subsets and are particularly suitable for conformer and 3D datasets, ensuring that all conformers of the same molecule remain in the same subset~\cite{ganea2021geomol}. (d) Protein-level splits separate protein--ligand complexes according to proteins, typically using protein clustering or sequence identity, to reduce information leakage and evaluate generalization to unseen proteins or protein families~\cite{NEURIPS2021_31445061}. Overall, random splits mainly assess performance on molecules similar to the training data, whereas scaffold- and protein-level splits provide stricter assessments of molecular and biological generalization.

	\subsection{Evaluation Benchmarks}

	Beyond benchmark datasets, several evaluation benchmarks have been developed to provide standardized procedures for assessing molecular generative models. They define task-specific evaluation settings and metrics, facilitating more consistent and reproducible comparisons across different methods.

	\noindent\textbf{GuacaMol.} GuacaMol is an open-source Python package for benchmarking models in \textit{de novo} drug design~\cite{brown2019guacamol}. It comprises five distribution-learning benchmarks and 20 goal-directed benchmarks, with six and five baseline methods evaluated, respectively. For distribution-learning evaluation, a generative model is trained on a standardized subset of ChEMBL and then generates a specified number of molecules for evaluation by GuacaMol. The generated molecules are assessed for validity, uniqueness, and novelty, while Kullback--Leibler (KL) divergence and Fréchet ChemNet Distance (FCD) compare the generated distributions with training distributions. GuacaMol also provides goal-directed benchmarks, in which the model is given a predefined scoring function to generate molecules that maximize the corresponding objective. The code is available at \url{https://github.com/BenevolentAI/guacamol}.

	\noindent\textbf{Practical Molecular Optimization (PMO).} PMO is an open-source benchmark designed to facilitate the transparent and reproducible evaluation of algorithmic advances in molecular optimization~\cite{gao2022sample}. It provides 25 molecular design algorithms across 23 tasks, with a special emphasis on sample efficiency. To reduce dataset bias, PMO uses ZINC for pretraining and initialization where applicable. For each of its 23 single-objective optimization tasks, an algorithm iteratively proposes molecules, which are evaluated by a task-specific oracle. The resulting scores guide subsequent proposals, forming a repeated generation--evaluation--optimization process. Performance is summarized by the area under the curve (AUC) of the Top-10 molecules, abbreviated as AUC Top-10. The curve plots the average score of the top-10 molecules found so far against the cumulative number of oracle calls. Thus, methods that reach high objective scores earlier within the oracle budget achieve larger AUC values, reflecting greater sample efficiency. The complete codebase is publicly available at \url{https://github.com/wenhao-gao/mol_opt}.

	\noindent\textbf{PoseBusters.} PoseBusters is a test suite for assessing the chemical and physical validity of protein--ligand poses~\cite{buttenschoen2024posebusters}. Generated poses are subjected to checks of chemical consistency, intramolecular geometry, and intermolecular interactions. A pose that passes all applicable checks is classified as PB-valid, and different methods can be compared by the proportion of PB-valid poses they produce. In the original study, seven docking methods were evaluated on the Astex Diverse set (85 complexes) and the more challenging PoseBusters Benchmark set (308 unseen complexes). The code is available at \url{https://github.com/maabuu/posebusters}.

	\noindent\textbf{PoseCheck.} PoseCheck is a package for assessing the quality of generated protein--ligand complexes from 3D target-conditioned generative models~\cite{harris2023benchmarking}. For each generated protein--ligand pose, it evaluates steric clashes, ligand strain energy, and protein--ligand interaction fingerprints. The ligand is then redocked, and changes in pose and interactions are examined to determine whether redocking substantially corrects the original generated pose. In the original study, seven generative models were evaluated on 100 test protein pockets from CrossDocked2020, with 100 molecules generated per target. The code is available at \url{https://github.com/cch1999/posecheck}.

	\noindent\textbf{GenBench3D.} GenBench3D is a benchmark designed for evaluating deep learning models that generate 3D molecules~\cite{baillif2024benchmarking}. It integrates molecular graph metrics, molecular conformation metrics, pocket-based metrics, and binding affinity metrics. Its central metric, Validity3D, assesses whether generated conformations exhibit plausible bond lengths and valence angles based on reference distributions from the Cambridge Structural Database (CSD), while also checking intramolecular steric clashes and puckered aromatic rings. The framework also evaluates conformation diversity and novelty, protein--ligand clashes, and binding affinity using Vina, Glide, and GOLD PLP. In the original study, GenBench3D was used to benchmark six structure-based 3D molecular generative models. All the code to benchmark models is available at \url{https://github.com/bbaillif/genbench3d}.

	\section{Representation}
	
\begin{figure*}[!t]
		\centering
		\includegraphics[width=0.9\linewidth]{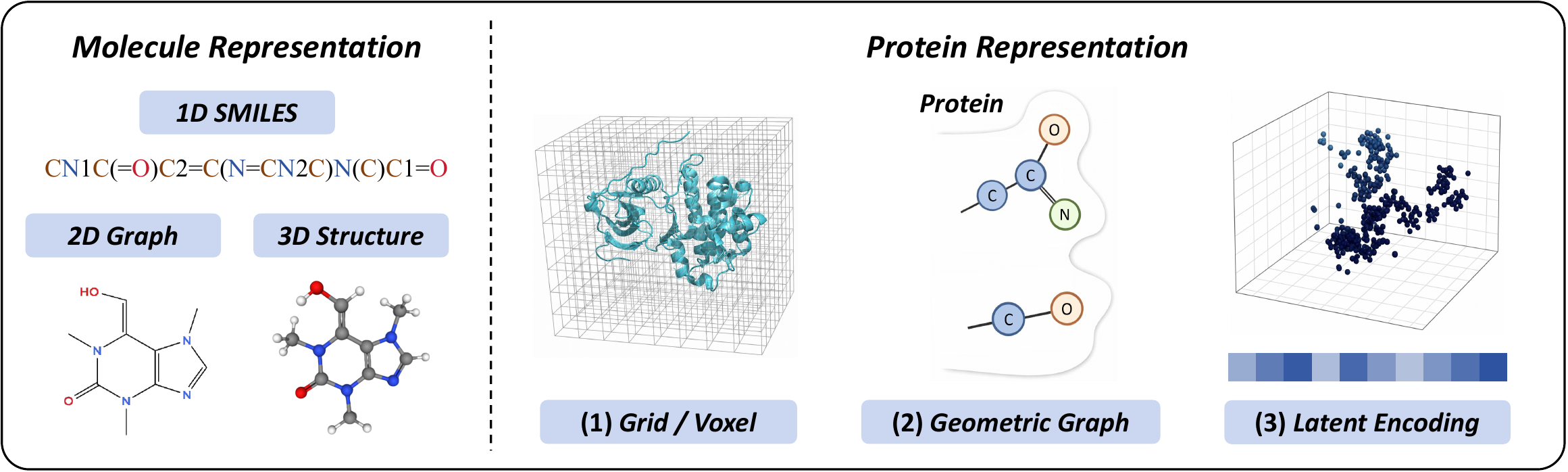}
		\caption{Overview of molecular and protein-pocket representations for molecule generation.}
		\label{fig:representation}
	\end{figure*}

	Molecular representation is the foundation of molecule generation~\cite{elton2019deep}. It determines how chemical structures are encoded for learning, what information is preserved or discarded, and whether generated candidates remain syntactically valid and chemically meaningful~\cite{xu2019deep}. Representation choices directly shape a model's ability to capture validity constraints, topological patterns, stereochemistry, and conformational behavior~\cite{krenn2020self,jaeger2018mol2vec,sumita2026molecular}. When protein-pocket information is incorporated, the selected representation also determines how geometric and physicochemical interaction cues are provided to the generative model~\cite{guan2023targetdiff,huang2024ipdiff}. A schematic overview of representative molecular and protein-pocket representations used across generative modeling pipelines is shown in Figure~\ref{fig:representation}.

	\subsection{Molecular Representations}

	Molecular representations determine how chemical structures are encoded and therefore strongly influence the learning objectives, model architectures, and chemical constraints used in molecule generation. At the one-dimensional level, molecules are represented as linear strings. SMILES encodes a molecular graph through atom symbols, bond symbols, branches, and ring-closure tokens, offering a compact representation that is well suited to sequence models but sensitive to syntax errors and atom-ordering choices~\cite{weininger1988smiles,bjerrum2017smiles}. Randomized SMILES alleviates atom-ordering dependence by providing multiple valid string encodings of the same molecular graph, whereas Self-Referencing Embedded Strings (SELFIES) uses semantic constraints to ensure that every SELFIES string can be decoded into a valid molecular structure, thereby substantially improving representation robustness~\cite{krenn2022selfies}. DeepSMILES modifies the treatment of branches and rings to reduce common syntax errors, while InChI provides a standardized, layered identifier designed primarily for canonical structure representation rather than direct molecule generation~\cite{o2018deepsmiles,heller2013inchi}. Beyond atom-level strings, fragment- or motif-based representations decompose molecules into chemically meaningful substructures, such as rings, functional groups, scaffolds, or junction-tree nodes. These higher-level units can reduce the number of low-level generation steps and help preserve local chemical validity, making them particularly useful for scaffold-aware and substructure-based generation~\cite{jin2020hierarchical,jin2018junction,david2020molecular}. Learned molecular language representations, including Mol2Vec embeddings and Transformer-based encodings, further map molecular tokens or substructures into continuous embedding spaces that capture statistical and semantic relationships across chemical corpora~\cite{jaeger2018mol2vec,chithrananda2020chemberta}.

	At the two-dimensional level, molecules are represented as graphs in which atoms form nodes and chemical bonds form edges. Node features may encode element type, formal charge, hybridization, aromaticity, and stereochemistry, whereas edge features typically describe bond type, conjugation, and ring membership. Graph neural networks (GNNs) propagate and aggregate information across neighboring atoms to learn local and global molecular features~\cite{gilmer2017neural,kipf2016semi,velivckovic2017graph,zhou2020graph,corso2024graph}. Graph-based generative models can then construct molecules atom by atom, bond by bond, or through larger substructures, with connectivity and valence constraints imposed during generation or post-processing~\cite{jin2018junction,jin2020hierarchical}. Three-dimensional representations extend molecular graphs by associating atoms with Cartesian coordinates and, in some cases, pairwise distances, bond angles, torsional angles, local frames, or molecular surfaces. Because molecular properties and intermolecular interactions depend on geometry, 3D generative models commonly adopt E(3)- or SE(3)-equivariant architectures so that their outputs transform consistently under rotation and translation~\cite{satorras2021n,bronstein2021geometric,hoogeboom2022equivariant}. These representations support conformation-aware generation, joint topology--geometry modeling, and protein--ligand interaction modeling. Continuous spatial representations provide an alternative to discrete atom-coordinate representations. Electron-density and electrostatic-potential fields describe electronic or interaction-relevant quantities distributed in three-dimensional space, whereas voxelized atomic-density grids encode molecular composition and shape on regular spatial lattices~\cite{xiang2025edbench,ragoza2022generating,o20233d}.

	\begin{table*}[!tp]
		\caption{Summary of Representative Deep Generative Frameworks for Molecule Generation.}
		\label{tbl:vae_gan_flow_diffusion_models_summary}
		\centering
		\footnotesize
		\setlength{\tabcolsep}{4pt}
		\begin{tabularx}{\textwidth}{l c c c c X}
			\toprule
			\multirow{2}{*}{Model} & \multirow{2}{*}{Year} & \multicolumn{3}{c}{Input} & \multirow{2}{*}{Code} \\
			\cmidrule(lr){3-5}
			& & 1D Seq. & 2D Graph & 3D Struct. & \\
			\midrule
			\rowcolor{cyan!5} \multicolumn{6}{l}{\textbf{RNN/Transformer-based Models}} \\
			\rowcolor{cyan!5} CharRNN~\cite{segler2018generating} & 2018 & \ding{51} & \ding{55} & \ding{55} & \url{https://github.com/molecularsets/moses/tree/master/moses/char_rnn} \\
			\rowcolor{cyan!5} REINVENT~\cite{blaschke2020reinvent,persico2026applying} & 2020 & \ding{51} & \ding{55} & \ding{55} & \url{https://github.com/MolecularAI/Reinvent} \\
			\rowcolor{cyan!5} MolGPT~\cite{bagal2021molgpt} & 2021 & \ding{51} & \ding{55} & \ding{55} & \url{https://github.com/devalab/molgpt} \\
			\rowcolor{cyan!5} GDL~\cite{li2022generative} & 2022 & \ding{51} & \ding{55} & \ding{55} & \url{https://doi.org/10.5281/zenodo.7074218} \\
			\rowcolor{cyan!5} cMolGPT~\cite{wang2023cmolgpt} & 2023 & \ding{51} & \ding{55} & \ding{55} & \url{https://github.com/VV123/cMolGPT} \\
			\rowcolor{cyan!5} GMTransformer~\cite{wei2023probabilistic} & 2023 & \ding{51} & \ding{55} & \ding{55} & \url{https://github.com/usccolumbia/GMTransformer} \\
			\rowcolor{cyan!5} FSM-DDTR~\cite{monteiro2023fsm} & 2023 & \ding{51} & \ding{55} & \ding{55} & \url{https://github.com/larngroup/FSM-DDTR} \\
			\rowcolor{cyan!5} CLM~\cite{moret2023leveraging,horne2023exploration} & 2023 & \ding{51} & \ding{55} & \ding{55} & \url{https://doi.org/10.5281/zenodo.7370858} \\
			\rowcolor{cyan!5} ACEGEN~\cite{bou2024acegen} & 2024 & \ding{51} & \ding{55} & \ding{55} & \url{https://github.com/acellera/acegen-open} \\
			\rowcolor{cyan!5} TamGen~\cite{wu2024tamgen} & 2024 & \ding{51} & \ding{55} & \ding{55} & \url{https://github.com/SigmaGenX/TamGen} \\
			\rowcolor{cyan!5} FragGPT~\cite{yue2024unlocking} & 2024 & \ding{51} & \ding{55} & \ding{55} & \url{https://github.com/pengbingxin/FragGPT-Interface} \\
			\rowcolor{cyan!5} CONSMI~\cite{qian2024consmi} & 2024 & \ding{51} & \ding{55} & \ding{55} & \url{https://github.com/WaterFlow1/CONSMI} \\
			\rowcolor{cyan!5} SMILES-RNN~\cite{thomas2025identification} & 2025 & \ding{51} & \ding{55} & \ding{55} & \url{https://github.com/MorganCThomas/SMILES-RNN} \\
			\rowcolor{cyan!5} TransPharmer~\cite{xie2025accelerating} & 2025 & \ding{51} & \ding{55} & \ding{55} & \url{https://github.com/iipharma/transpharmer-repo} \\
			\rowcolor{blue!5} \multicolumn{6}{l}{\textbf{VAE-based Models}} \\
			\rowcolor{blue!5} GrammarVAE~\cite{pmlr-v70-kusner17a} & 2017 & \ding{51} & \ding{55} & \ding{55} & \url{https://github.com/mkusner/grammarVAE} \\
			\rowcolor{blue!5} GraphVAE~\cite{simonovsky2018graphvae} & 2018 & \ding{55} & \ding{51} & \ding{55} & -- \\
			\rowcolor{blue!5} JT-VAE~\cite{jin2018junction} & 2018 & \ding{55} & \ding{51} & \ding{55} & \url{https://github.com/wengong-jin/icml18-jtnn} \\
			\rowcolor{blue!5} ARAE~\cite{hong2019molecular} & 2019 & \ding{51} & \ding{55} & \ding{55} & -- \\
			\rowcolor{blue!5} Fragment-based VAE~\cite{podda2020deep} & 2020 & \ding{51} & \ding{55} & \ding{55} & \url{https://github.com/marcopodda/fragment-based-dgm} \\
			\rowcolor{blue!5} RNN-Attention~\cite{dollar2021attention} & 2021 & \ding{51} & \ding{55} & \ding{55} & -- \\
			\rowcolor{blue!5} TransVAE~\cite{dollar2021attention} & 2021 & \ding{51} & \ding{55} & \ding{55} & \url{https://github.com/oriondollar/TransVAE} \\
			\rowcolor{blue!5} Graph Transformer VAE~\cite{mitton2021graph} & 2021 & \ding{55} & \ding{51} & \ding{55} & -- \\
			\rowcolor{blue!5} Conditional $\beta$-VAE~\cite{fil2021beta} & 2021 & \ding{51} & \ding{55} & \ding{55} & -- \\
			\rowcolor{blue!5} 3DLinker~\cite{huang2022threedlinker} & 2022 & \ding{55} & \ding{55} & \ding{51} & -- \\
			\rowcolor{blue!5} D-MolVAE~\cite{du2022small} & 2022 & \ding{55} & \ding{51} & \ding{55} & -- \\
			\rowcolor{blue!5} TrustMol~\cite{wijaya2025trustworthy} & 2025 & \ding{51} & \ding{55} & \ding{51} & \url{https://github.com/ktirta/TrustMol} \\
			\rowcolor{green!5} \multicolumn{6}{l}{\textbf{GAN-based Models}} \\
			\rowcolor{green!5} ORGAN~\cite{guimaraes2017objective} & 2017 & \ding{51} & \ding{55} & \ding{55} & \url{https://github.com/gablg1/ORGAN} \\
			\rowcolor{green!5} MolGAN~\cite{de2018molgan} & 2018 & \ding{55} & \ding{51} & \ding{55} & \url{https://github.com/nicola-decao/MolGAN} \\
			\rowcolor{green!5} TransORGAN~\cite{li2022transorgan} & 2022 & \ding{51} & \ding{55} & \ding{55} & -- \\
			\rowcolor{green!5} EarlGAN~\cite{tang2023earlgan} & 2023 & \ding{51} & \ding{55} & \ding{55} & \url{https://github.com/tang7777777/EarlGAN} \\
			\rowcolor{green!5} SpotGAN~\cite{li2023spotgan} & 2023 & \ding{51} & \ding{55} & \ding{55} & \url{https://github.com/naruto7283/SpotGAN} \\
			\rowcolor{green!5} SpotWGAN~\cite{li2023spotgan} & 2023 & \ding{51} & \ding{55} & \ding{55} & \url{https://github.com/naruto7283/SpotGAN} \\
			\rowcolor{green!5} InstGAN~\cite{tang2025instgan} & 2025 & \ding{51} & \ding{55} & \ding{55} & \url{https://github.com/tang777777/InstGAN} \\
			\rowcolor{green!5} TopMT-GAN~\cite{wang2025topmt} & 2025 & \ding{55} & \ding{55} & \ding{51} & \url{https://github.com/aeghnnsw/TopMT} \\
			\rowcolor{green!5} RL-MolGAN~\cite{li2026reinforcement} & 2026 & \ding{51} & \ding{55} & \ding{55} & \url{https://github.com/tang777777/MIR} \\
			\rowcolor{orange!5} \multicolumn{6}{l}{\textbf{Flow-based Models}} \\
			\rowcolor{orange!5} GraphAF~\cite{shi2020graphaf} & 2020 & \ding{55} & \ding{51} & \ding{55} & \url{https://github.com/DeepGraphLearning/GraphAF} \\
			\rowcolor{orange!5} MoFlow~\cite{zang2020moflow} & 2020 & \ding{55} & \ding{51} & \ding{55} & \url{https://github.com/calvin-zcx/moflow} \\
			\rowcolor{orange!5} GraphDF~\cite{luo2021graphdf} & 2021 & \ding{55} & \ding{51} & \ding{55} & \url{https://github.com/divelab/DIG/tree/dig-stable/dig/ggraph/GraphDF} \\
			\rowcolor{orange!5} GraphCNF~\cite{lippe2021categorical} & 2021 & \ding{55} & \ding{51} & \ding{55} & \url{https://github.com/lrjconan/GRF} \\
			\rowcolor{orange!5} GeoBFN~\cite{song2024unified} & 2024 & \ding{55} & \ding{55} & \ding{51} & \url{https://github.com/AlgoMole/GeoBFN} \\
			\rowcolor{orange!5} FlowMol~\cite{dunn2024mixed} & 2024 & \ding{55} & \ding{51} & \ding{51} & \url{https://github.com/dunni3/FlowMol} \\
			\rowcolor{orange!5} GGFlow~\cite{hou2024ggflow} & 2024 & \ding{55} & \ding{51} & \ding{55} & \url{https://github.com/Xiaoyang878/GGFlow} \\
			\rowcolor{orange!5} SemlaFlow~\cite{irwin2024semlaflow} & 2024 & \ding{55} & \ding{51} & \ding{51} & \url{https://github.com/rssrwn/semla-flow} \\
			\rowcolor{orange!5} VNFlow~\cite{hostavs2025vnflow} & 2025 & \ding{51} & \ding{55} & \ding{55} & \url{https://github.com/nrc-cnrc/VNFlow} \\
			\rowcolor{orange!5} Megalodon-flow~\cite{reidenbach2026modular} & 2026 & \ding{55} & \ding{51} & \ding{51} & \url{https://github.com/NVIDIA-Digital-Bio/megalodon} \\
			\rowcolor{purple!5} \multicolumn{6}{l}{\textbf{Diffusion-based Models}} \\
			\rowcolor{purple!5} GDSS~\cite{jo2022score} & 2022 & \ding{55} & \ding{51} & \ding{55} & \url{https://github.com/harryjo97/GDSS} \\
			\rowcolor{purple!5} EDM~\cite{hoogeboom2022equivariant} & 2022 & \ding{55} & \ding{55} & \ding{51} & \url{https://github.com/ehoogeboom/e3_diffusion_for_molecules} \\
			\rowcolor{purple!5} DiGress~\cite{vignac2022digress} & 2022 & \ding{55} & \ding{51} & \ding{55} & \url{https://github.com/cvignac/DiGress} \\
			\bottomrule
		\end{tabularx}
	\end{table*}

	\begin{table*}[!tp]
		\ContinuedFloat
		\caption{continued}
		\centering
		\footnotesize
		\setlength{\tabcolsep}{4pt}
		\begin{tabularx}{\textwidth}{l c c c c X}
			\toprule
			\multirow{2}{*}{Model} & \multirow{2}{*}{Year} & \multicolumn{3}{c}{Input} & \multirow{2}{*}{Code} \\
			\cmidrule(lr){3-5}
			& & 1D Seq. & 2D Graph & 3D Struct. & \\
			\midrule
			\rowcolor{purple!5} D2L-OMP~\cite{guo2024diffusing} & 2023 & \ding{55} & \ding{51} & \ding{55} & -- \\
			\rowcolor{purple!5} MDM~\cite{huang2023mdm} & 2023 & \ding{55} & \ding{55} & \ding{51} & \url{https://github.com/tencent-ailab/MDM} \\
			\rowcolor{purple!5} MiDi~\cite{vignac2023midi} & 2023 & \ding{55} & \ding{51} & \ding{51} & \url{https://github.com/cvignac/MiDi} \\
			\rowcolor{purple!5} VoxMol~\cite{o20233d} & 2023 & \ding{55} & \ding{55} & \ding{51} & \url{https://github.com/Genentech/voxmol} \\
			\rowcolor{purple!5} DiffLinker~\cite{igashov2024equivariant} & 2024 & \ding{55} & \ding{55} & \ding{51} & \url{https://github.com/igashov/DiffLinker} \\
			\rowcolor{purple!5} END~\cite{cornet2024equivariant} & 2024 & \ding{55} & \ding{55} & \ding{51} & \url{https://github.com/frcnt/equivariant-neural-diffusion} \\
			\rowcolor{purple!5} MUDiff~\cite{hua2024mudiff} & 2024 & \ding{55} & \ding{51} & \ding{51} & \url{https://github.com/WillHua127/mudiff} \\
			\rowcolor{purple!5} Twigs~\cite{mercatali2024diffusion} & 2024 & \ding{55} & \ding{55} & \ding{51} & \url{https://github.com/Aalto-QuML/Diffusion_twigs} \\
			\rowcolor{purple!5} 3D-EDiffMG~\cite{xu2025threedediffmg} & 2025 & \ding{55} & \ding{51} & \ding{51} & \url{https://github.com/ZheLi-Lab-Collaboration/3D-EDiffMG} \\
			\rowcolor{purple!5} ADiT~\cite{joshi2025allatom} & 2025 & \ding{55} & \ding{55} & \ding{51} & \url{https://github.com/facebookresearch/all-atom-diffusion-transformer} \\
			\rowcolor{purple!5} DiffMC-Gen~\cite{yang2025diffmcgen} & 2025 & \ding{55} & \ding{51} & \ding{51} & -- \\
			\rowcolor{purple!5} MDRL~\cite{yuan2025mdrl} & 2025 & \ding{55} & \ding{55} & \ding{51} & \url{https://github.com/Xinol1024/MDRL} \\
			\rowcolor{purple!5} TED~\cite{lu2026bridging} & 2026 & \ding{55} & \ding{51} & \ding{51} & \url{https://github.com/BaruaBee/TED} \\
			\bottomrule
		\end{tabularx}
	\end{table*}

	\subsection{Protein-Pocket and Interaction-Aware Representations}

	Protein information can be represented at multiple levels, ranging from amino acid sequences and pairwise residue relationships to explicit three-dimensional pocket structures. Sequence-derived representations encode residue identity and, when obtained from evolutionary profiles or pretrained protein language models, may also capture evolutionary and contextual information, but they only indirectly reflect the spatial and chemical environment of the binding pocket. Contact and distance maps provide coarse pairwise structural constraints but do not preserve full spatial orientation, atomic composition, or local physicochemical features. Therefore, pocket-conditioned molecule generation predominantly relies on local three-dimensional representations that characterize the geometry and chemistry surrounding the ligand-binding region. Protein pockets can be encoded through several complementary formalisms: grid- or voxel-based representations discretize atomic or physicochemical features in three-dimensional space for convolutional processing~\cite{ragoza2022generating,pinheiro2024voxbind}; geometric representations model pocket atoms or residues as point clouds or graphs processed by distance-aware GNNs or SE(3)-equivariant networks~\cite{guan2023targetdiff,lin2025diffbp,schneuing2024structure}; and latent pocket encodings compress sequence-derived, structural, or pharmacophoric information into conditional vectors for efficient target-specific generation~\cite{islam2025conditioned,shen2024tacogfn}. Interaction-aware conditioning may further incorporate explicit protein--ligand contacts or learned protein--ligand interaction priors~\cite{kang2026pocket,huang2024ipdiff}, as well as pharmacophore-based constraints~\cite{zhu2023pharmacophore,feng2024generation}, rather than relying solely on the receptor's geometric representation. In practical workflows, protein pockets are commonly defined during preprocessing by retaining protein atoms or residues within a predefined distance from the bound or reference ligand.

	\section{Deep Generative Frameworks for Molecule Generation}

	Molecule generation is built upon deep generative models that learn molecular distributions from large-scale chemical data and sample novel candidate structures under chemical, structural, or property-related constraints~\cite{lim2018molecular,sattarov2019novo}. Although RNN/Transformer-based, VAE-based, GAN-based, flow-based, and diffusion-based models differ in molecular representation, learning objective, and sampling mechanism, they provide reusable algorithmic foundations for autoregressive sequence modeling, latent-space sampling, distribution matching, invertible transformation, and iterative denoising. These frameworks can be applied to sequence-, graph-, and structure-based molecule generation, and can also be adapted to conditional or target-informed settings when additional structural or property information is available. This section reviews representative deep generative frameworks for molecule generation, as summarized in Table~\ref{tbl:vae_gan_flow_diffusion_models_summary}, discusses their major training objectives, and examines how these frameworks are extended to pocket-conditioned molecule generation.
	
	\subsection{RNN/Transformer-based Models}

\begin{figure*}[!t]
		\centering
		\includegraphics[width=\linewidth]{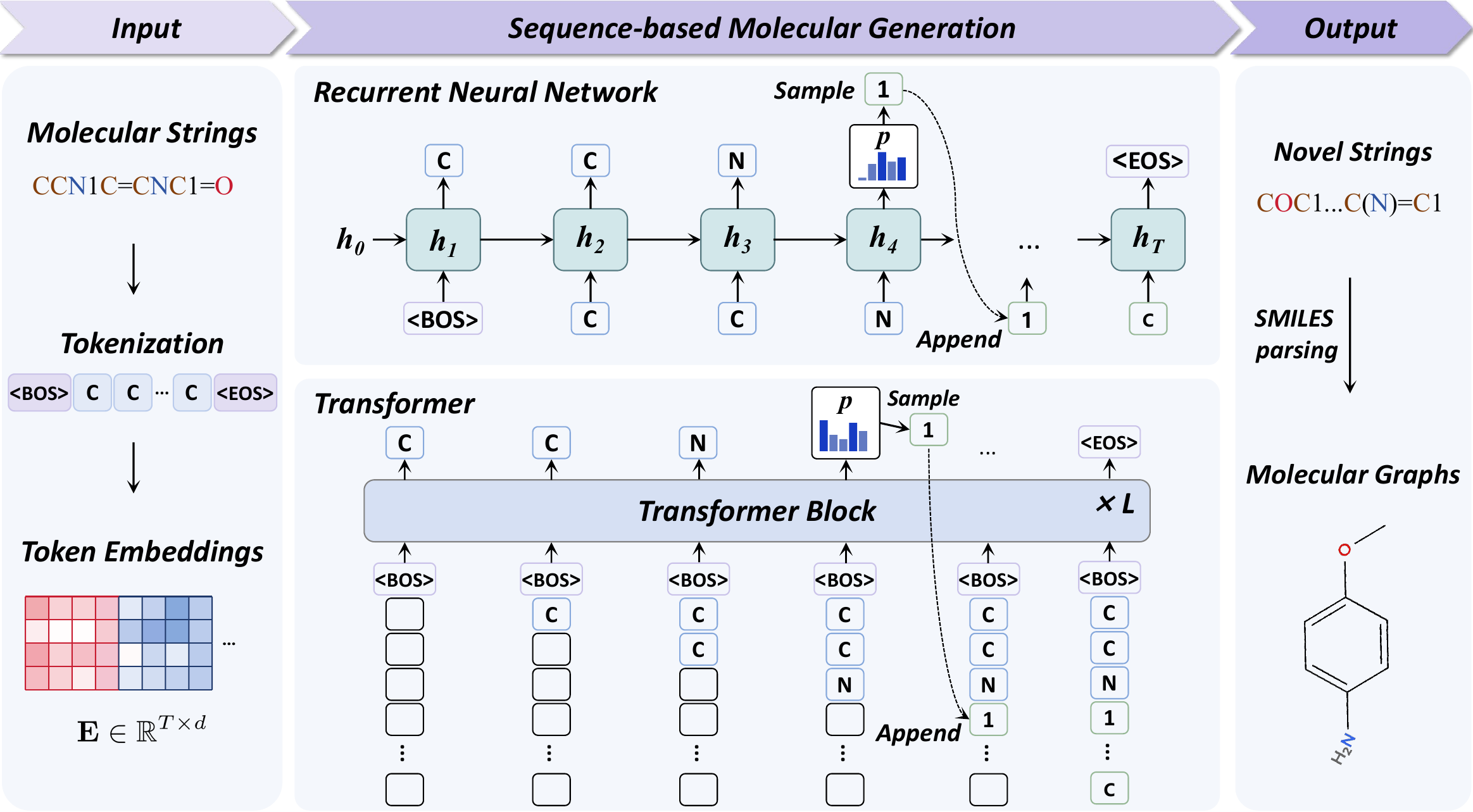}
		\caption{
		RNN- and Transformer-based molecule generation pipelines with molecular sequences, where tokenized molecular strings are autoregressively modeled using recurrent hidden states or causal self-attention to generate novel molecular strings that are parsed into molecular graphs.
		}
		\label{fig:rnn_transformer}
	\end{figure*}

	RNNs and Transformers provide autoregressive frameworks for sequence-based molecule generation by modeling molecules as ordered token sequences, most commonly SMILES strings. As illustrated in Figure~\ref{fig:rnn_transformer}, a molecular string is tokenized between beginning- and end-of-sequence tokens (\texttt{<BOS>} and \texttt{<EOS>}) and subsequently mapped to an embedding matrix $\mathbf{E}\in\mathbb{R}^{T\times d}$, where $T$ and $d$ denote the sequence length and embedding dimension, respectively. Given a molecular sequence $x=(x_1,\ldots,x_T)$, its probability distribution can be factorized autoregressively (Equation~\ref{eq:autoregressive_factorization}):
	\begin{equation}
	p(x)=\prod_{t=1}^{T} p(x_t\mid x_{<t}),
	\label{eq:autoregressive_factorization}
	\end{equation}
	where $x_{<t}$ denotes the preceding tokens. RNNs recurrently update hidden states from the initial state $h_0$ to $h_t$ to predict the next-token distribution $p$, whereas Transformers process the prefix through $L$ stacked causal self-attention blocks, enabling direct access to preceding tokens and more effective modeling of long-range dependencies~\cite{wei2023probabilistic}. At each step, a token is sampled from $p$ and appended to the sequence until \texttt{<EOS>} is generated. These architectures can further incorporate molecular properties, biological targets, fragments, or other conditions for more flexible and controllable molecule generation~\cite{wang2023cmolgpt,wu2024tamgen,yue2024unlocking}.

	Early sequence-based molecular generators were largely built on recurrent architectures. CharRNN represents an influential early example, demonstrating that character-level RNNs can learn SMILES distributions and generate focused molecular libraries~\cite{segler2018generating}. REINVENT introduced a production-ready framework that couples recurrent SMILES generation with reinforcement learning to steer sampling toward user-defined objectives~\cite{blaschke2020reinvent}. With the emergence of Transformer architectures, sequence-based generation benefited from more flexible context modeling and conditioning mechanisms. MolGPT applies a Transformer decoder with masked self-attention to next-token molecule generation and supports both property- and scaffold-conditioned generation~\cite{bagal2021molgpt}. Subsequent extensions further broadened the controllability of this paradigm: FSM-DDTR incorporates a feedback loop and multi-objective optimization into Transformer-based molecule generation~\cite{monteiro2023fsm}, whereas TransPharmer integrates pharmacophore information into sequence generation to enable pharmacophore-constrained molecule generation~\cite{xie2025accelerating}.

	\subsection{VAE-based Models}
	
\begin{figure*}[!t]
		\centering
		\includegraphics[width=\linewidth]{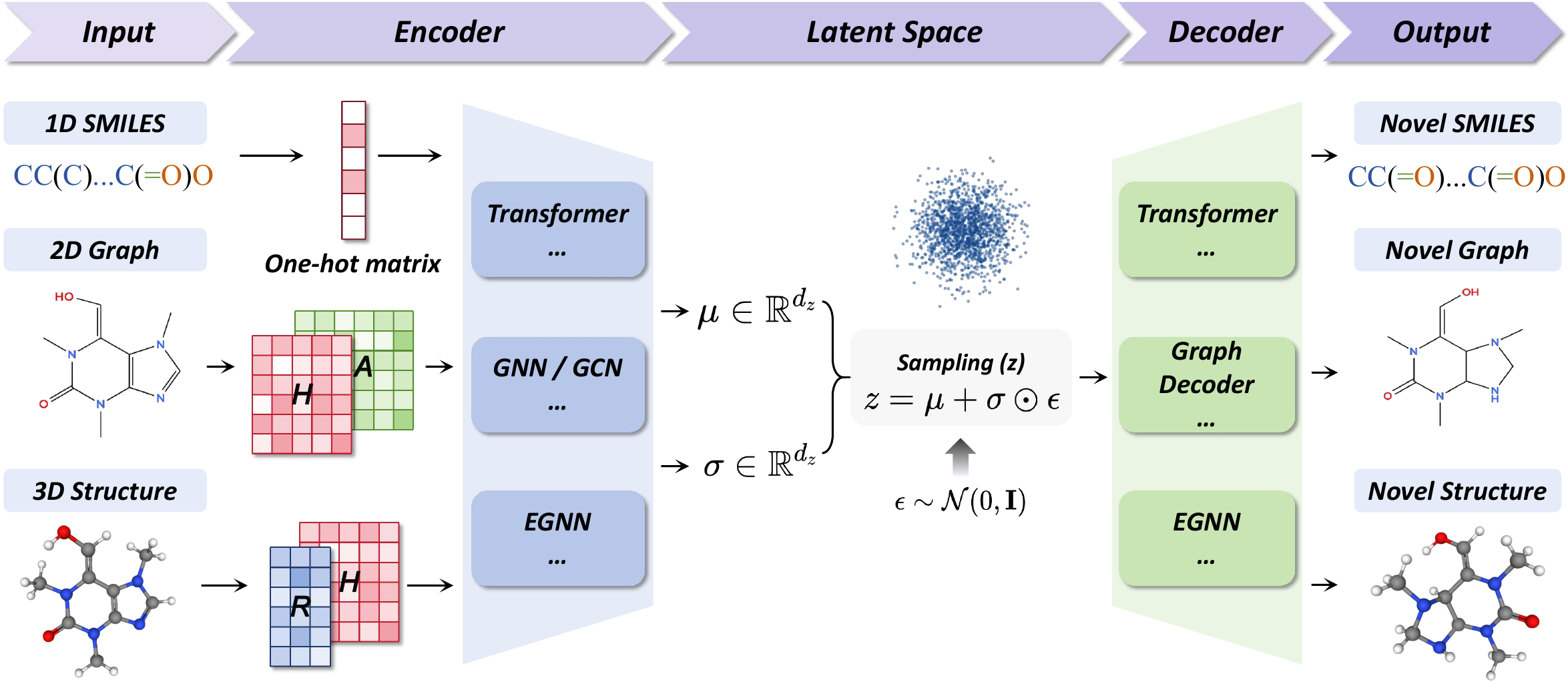}
		\caption{
		VAE-based molecule generation pipelines across 1D, 2D, and 3D molecular representations, where molecules are encoded into a Gaussian latent space and decoded to generate novel structures.
		}
		\label{fig:vae_pipeline}
	\end{figure*}

	VAEs provide a probabilistic framework for molecule generation by learning a continuous latent representation of discrete molecular structures, enabling sampling from a regularized latent space and decoding back to valid molecular instances~\cite{lim2018molecular,samanta2020vae,yan2020re}. As illustrated in Figure~\ref{fig:vae_pipeline}, molecules can be encoded as one-hot SMILES representations, 2D molecular graphs with atom-feature matrices $H$ and adjacency matrices $A$, or 3D structures with coordinate matrices $R$, which are processed by encoders such as Transformers~\cite{yoshikai2024novel}, GNNs/graph convolutional networks (GCNs)~\cite{kondratyev2023generative,ma2021gf}, and E(3)-equivariant graph neural networks (EGNNs), respectively. These representations are mapped into a latent Gaussian space parameterized by $\mu$ and $\sigma$, where latent variables are sampled via the reparameterization trick $z=\mu+\sigma\odot\epsilon$ with $\epsilon\sim\mathcal{N}(0,\mathbf{I})$~\cite{fil2021beta,xu2025exploring}. The decoder then reconstructs or generates molecules in the corresponding representation, using sequence decoders, graph decoders, or geometric decoders, thereby enabling end-to-end optimization of reconstruction fidelity and latent regularization~\cite{sattarov2019novo,dollar2021attention,zhou2023deep}.

	Based on this encoder--decoder formulation, VAE-based molecule generation models have explored different molecular representations and latent-space designs. Recent VAE-based studies have also explored latent-variable generation across complementary molecular representations~\cite{wijaya2025trustworthy}. GrammarVAE constrains SMILES generation with formal grammars, whereas GraphVAE directly decodes molecular graphs~\cite{pmlr-v70-kusner17a,simonovsky2018graphvae}. Sequence- and fragment-oriented variants such as ARAE, D-MolVAE, and Fragment-based VAE improve latent-space learning by combining adversarial regularization, graph decomposition, or fragment-level generation~\cite{hong2019molecular,du2022small,podda2020deep}. Graph Transformer VAE combines a graph-convolutional VAE with a Transformer-based module for molecular edge prediction, while 3DLinker employs an E(3)-equivariant graph VAE to predict anchor atoms and jointly generate linker graphs and three-dimensional coordinates conditioned on input molecular fragments~\cite{mitton2021graph,huang2022threedlinker}.

	\subsection{GAN-based Models}
	
\begin{figure*}[!t]
		\centering
		\includegraphics[width=\linewidth]{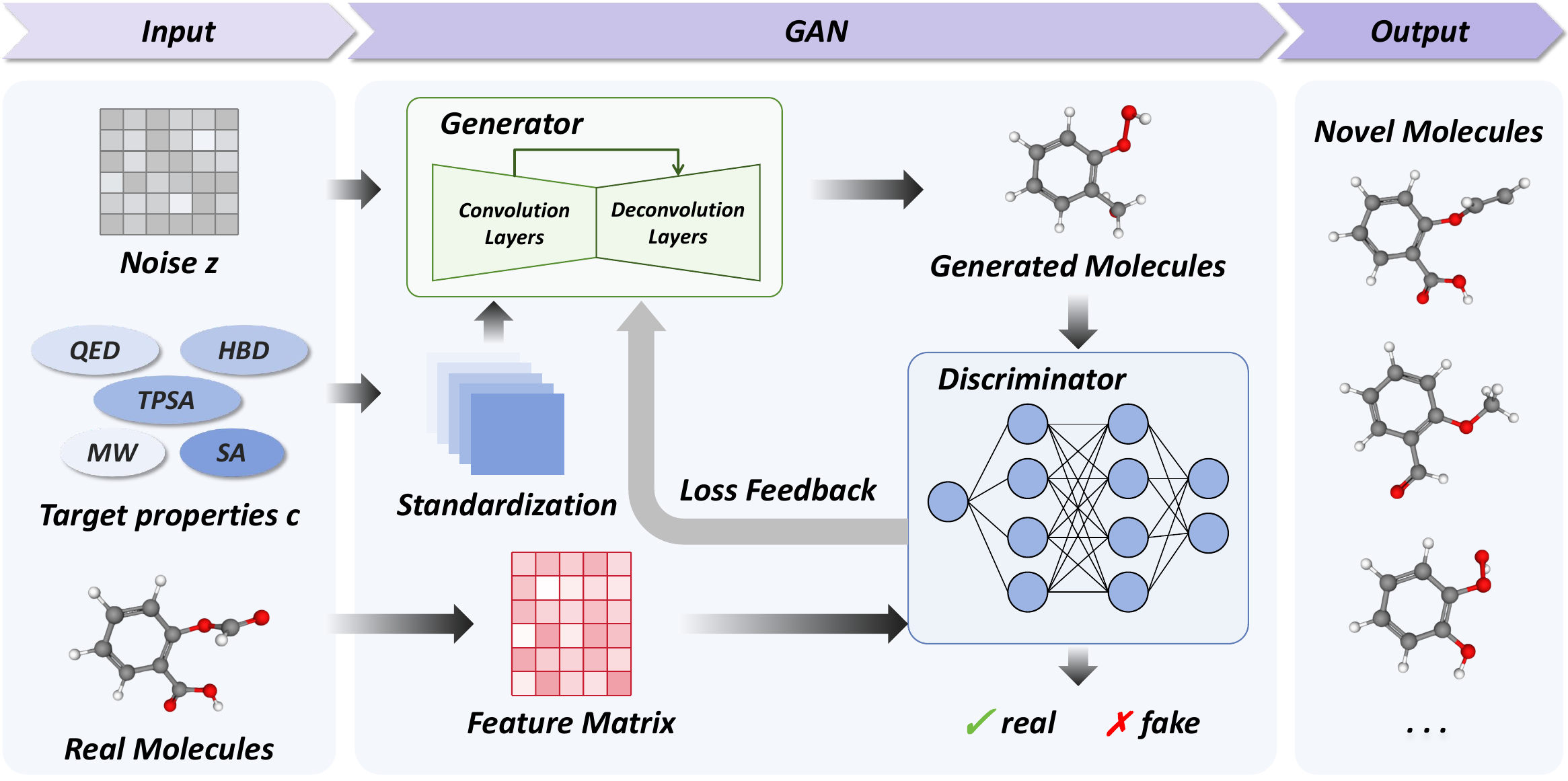}
		\caption{
		GAN-based molecule generation pipelines with 3D molecular representations, where a generator produces molecules from latent noise and a discriminator distinguishes real and generated samples.
		}
		\label{fig:gan_framework}
	\end{figure*}

	GANs formulate molecule generation as an adversarial learning problem between a generator $G$ and a discriminator $D$~\cite{guimaraes2017objective,NIPS2014_f033ed80}. As illustrated in Figure~\ref{fig:gan_framework}, in a representative property-guided GAN-based pipeline, the generator maps a latent noise vector $z$, optionally combined with property-related information $c$, into candidate molecular representations, while the discriminator distinguishes generated samples from real molecules~\cite{NIPS2014_f033ed80,mirza2014conditional}. Depending on the specific architecture, the generator may map random noise into molecular graphs, generate SMILES sequences, or produce scaffold-conditioned molecular decorations, while desired properties may be incorporated either through explicit conditioning or reinforcement-learning rewards~\cite{li2023spotgan,macedo2024medgan,li2024tengan}. Through this adversarial optimization process, the generator is trained to approximate the molecular distribution, enabling the generation of chemically plausible structures. In structure-based 3D generation, GANs have been applied to different molecular representations. Wang et al.\ generated ligand electron-density maps conditioned on protein-pocket electron density and subsequently interpreted them as molecular structures, whereas TopMT-GAN first constructs three-dimensional molecular topologies within protein pockets and then assigns atom and bond types using a second adversarial model~\cite{wang2022pocket,wang2025topmt}.

	Following this adversarial learning principle, GAN-based molecule generation models have been extended across sequence, graph, and property-oriented optimization settings. Sequence-based variants adapt adversarial learning to SMILES generation using different sequence architectures. TransORGAN employs a Transformer-based generator with policy-gradient training, whereas EarlGAN combines a long short-term memory (LSTM)-based generator and bidirectional LSTM discriminator within an actor--critic reinforcement-learning framework~\cite{li2022transorgan,tang2023earlgan}. Later extensions, including InstGAN and RL-MolGAN, incorporate instant and global reward signals or reinforcement-learning signals to improve property-oriented molecular optimization~\cite{tang2025instgan,li2026reinforcement}. MolGAN generates small molecular graphs using an implicit likelihood-free GAN with reinforcement learning~\cite{de2018molgan}.

	\subsection{Flow-based Models}
	
\begin{figure*}[!t]
		\centering
		\includegraphics[width=\linewidth]{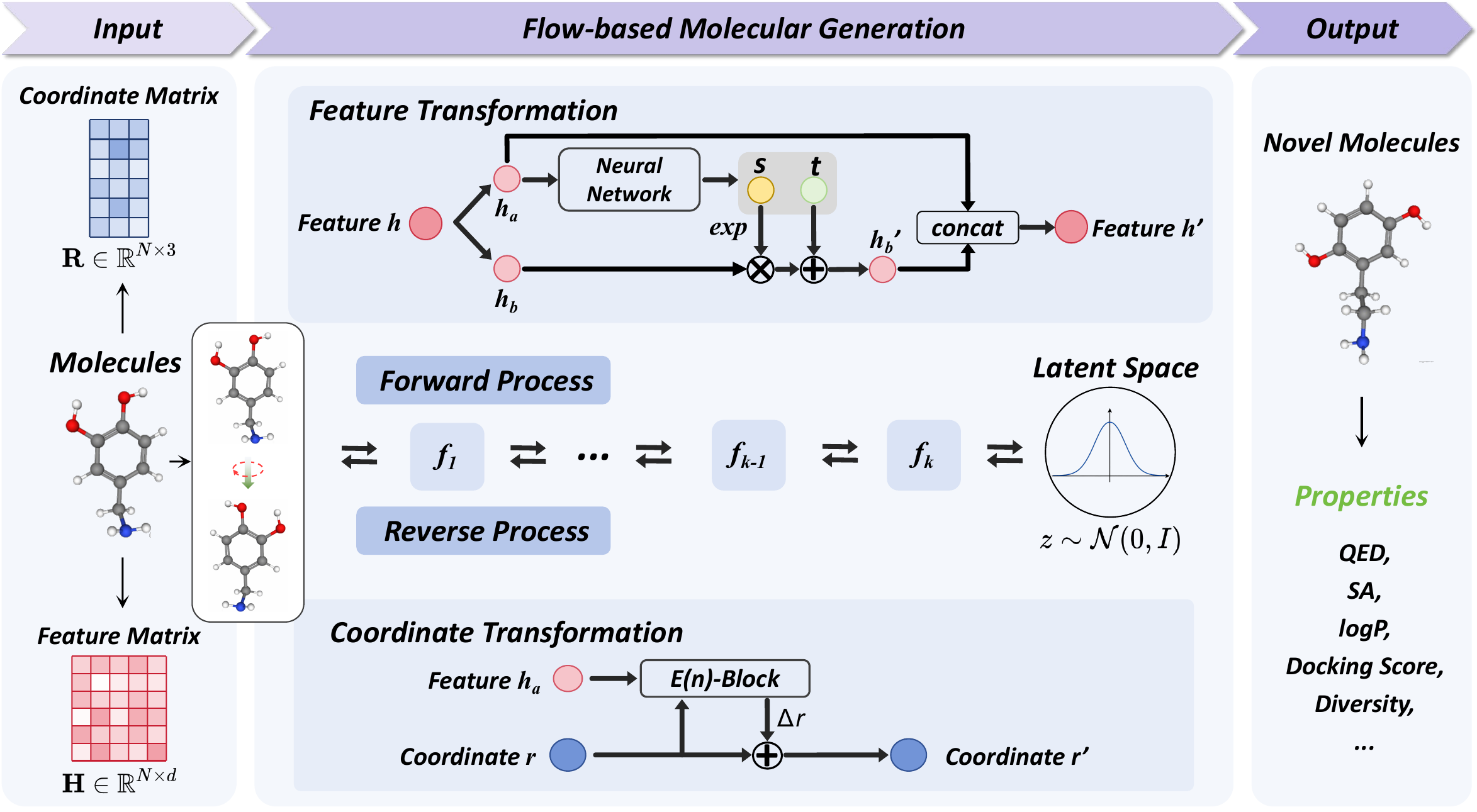}
		\caption{
		Normalizing-flow-based molecule generation pipelines with 3D molecular representations, where molecular structures are mapped through a sequence of invertible transformations from data space to a Gaussian latent space and reconstructed via the inverse process.
		}
		\label{fig:flow_framework}
	\end{figure*}

	Flow-based models encompass several mathematically distinct formulations, including normalizing flows, flow matching, and Bayesian flow networks. Normalizing-flow models construct an explicitly invertible mapping through a sequence of bijective transformations, enabling tractable likelihood evaluation and direct sampling~\cite{zang2020moflow}. Flow-matching models, exemplified by FlowMol, instead learn a continuous or discrete transport process without requiring each neural-network block to be explicitly invertible~\cite{dunn2024mixed}, whereas Bayesian flow networks, exemplified by GeoBFN, iteratively update the parameters of probability distributions over molecular variables through Bayesian inference~\cite{song2024unified}. In 3D molecular settings, a molecule is represented by an atomic feature matrix $H\in\mathbb{R}^{N\times d}$ and a coordinate matrix $R\in\mathbb{R}^{N\times 3}$, where $N$ denotes the number of atoms. As illustrated in Figure~\ref{fig:flow_framework}, a normalizing-flow model $f=f_k\circ\cdots\circ f_1$ transforms molecular representations into latent variables $z\sim\mathcal{N}(0,\mathbf{I})$, while the inverse mapping enables generation from latent space back to molecular structures~\cite{NEURIPS2021_21b5680d}.
	Each invertible block $f_i$ typically performs feature transformation and coordinate transformation jointly. For feature updates, an input representation $h$ is split and transformed via a neural network to produce scale and shift parameters $(s,t)$, followed by element-wise exponential scaling, translation, and concatenation to obtain updated features $h'$. For geometric updates, atomic coordinates $r$ are refined through equivariant $E(n)$-blocks that predict displacement $\Delta r$, yielding transformed coordinates $r'$. The forward and reverse processes are composed of stacked transformations $f_1,\dots,f_k$, enabling reversible mapping between data and latent spaces $z\sim\mathcal{N}(0,\mathbf{I})$.
	Within this reversible density-estimation framework, normalizing-flow models have mainly focused on molecular graph generation and geometric extensions. Autoregressive flow formulations provide a tractable route to stepwise molecular graph generation~\cite{shi2020graphaf}. At the graph level, GraphDF uses discrete latent variables and invertible modulo-shift transformations for molecular graph generation and property-oriented optimization, whereas GraphCNF constructs a permutation-invariant graph generator by mapping categorical graph variables into continuous representations~\cite{luo2021graphdf,lippe2021categorical}. VNFlow further integrates a variational autoencoder with normalizing flows for molecule generation~\cite{hostavs2025vnflow}. Beyond explicitly invertible normalizing flows, recent flow-matching methods such as GGFlow and Megalodon-flow further develop scalable graph and geometric transport architectures for molecule generation across 2D and 3D settings~\cite{hou2024ggflow,reidenbach2026modular}.

	\subsection{Diffusion-based Models}
	
\begin{figure*}[!t]
		\centering
		\includegraphics[width=\linewidth]{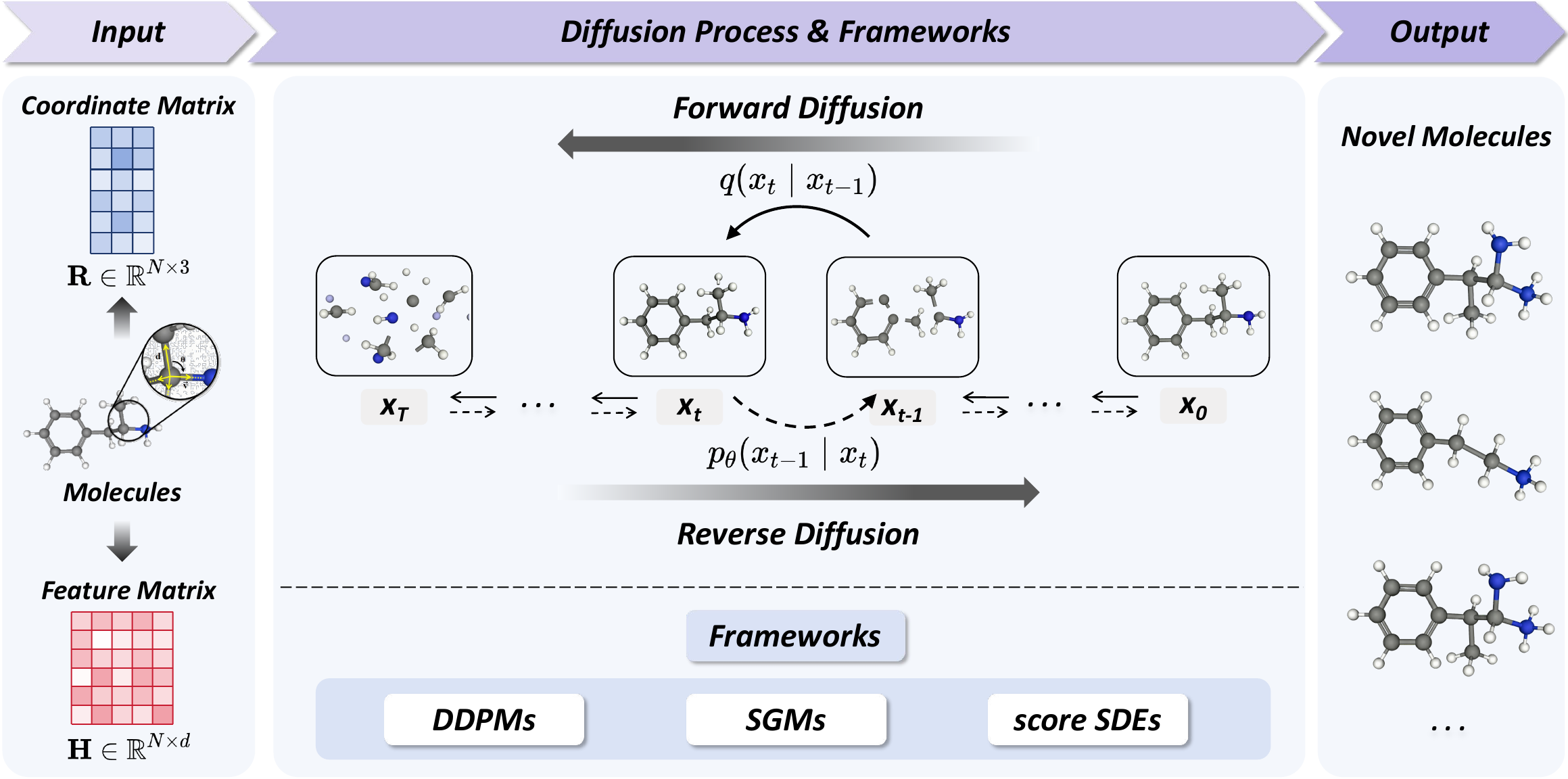}
		\caption{
		Diffusion-based molecule generation pipelines with 3D molecular representations, where molecules are progressively corrupted into noise through a forward diffusion process and reconstructed via a learned reverse denoising process.
		}
		\label{fig:diffusion_overview}
	\end{figure*}

	As illustrated in Figure~\ref{fig:diffusion_overview}, diffusion models formulate generative modeling as a stochastic denoising process that transforms data $x_0$ into noise $x_T \sim \mathcal{N}(0,\mathbf{I})$ and then reverses this process to generate molecular samples~\cite{sohl2015deep}. Classical diffusion frameworks include denoising diffusion probabilistic models (DDPMs) and score-based generative models (SGMs), both of which can be unified under the continuous-time score-based stochastic differential equation (score-SDE) framework.

	DDPMs define diffusion as a discrete-time Markov chain with a fixed Gaussian noise schedule $\{\beta_t\}$ in the forward process and a learnable reverse transition $p_\theta(x_{t-1}\mid x_t)$ for iterative denoising~\cite{ho2020denoising}.
	SGMs learn a noise-conditional score function $s_\theta(x_t,\sigma_t)\approx\nabla_{x_t}\log q_{\sigma_t}(x_t)$ across noise levels $\{\sigma_t\}$, and generate samples by following the estimated score field from $\sigma_{\max}$ to $\sigma_{\min}$~\cite{NEURIPS2019_3001ef25}.
	The score-SDE framework generalizes diffusion to continuous-time dynamics $\mathrm{d}x = f(x,t)\mathrm{d}t + g(t)\mathrm{d}\mathbf{w}$, where generation corresponds to solving a reverse-time SDE driven by the score function $\nabla_x \log p_t(x)$~\cite{song2020score}. The forward and reverse dynamics of the three frameworks can be interpreted under a unified perspective of iteratively mapping molecular structures between data and Gaussian noise distributions.

	Molecule generation studies have extended diffusion models to chemically specific generation tasks. At the molecular graph level, GDSS and DiGress formulate diffusion over atoms and bonds, enabling joint generation of molecular topology and chemical types~\cite{jo2022score,vignac2022digress}. Related models adapt diffusion to property-oriented optimization and 3D molecule generation, such as D2L-OMP and MDM~\cite{guo2024diffusing,huang2023mdm}. Several models have also explored the joint modeling of molecular graphs and three-dimensional geometry~\cite{vignac2023midi,yang2025diffmcgen}. For structure-aware generation, DiffLinker and END introduce equivariant diffusion mechanisms for molecular linker design and 3D coordinate generation~\cite{igashov2024equivariant,cornet2024equivariant}. More recent methods have explored distinct extensions of molecular diffusion: Twigs employs multiple co-evolving diffusion processes for conditional graph generation and molecular optimization~\cite{mercatali2024diffusion}; 3D-EDiffMG introduces an equivariant diffusion framework for three-dimensional molecule generation~\cite{xu2025threedediffmg}; ADiT develops a unified all-atom latent diffusion framework for molecules and periodic materials~\cite{joshi2025allatom}; and MDRL combines three-dimensional molecular diffusion with reinforcement learning for the joint optimization of multi-target binding and molecular properties~\cite{yuan2025mdrl}.

	\subsection{Methodological Trade-offs Across Generative Paradigms}

	Each generative paradigm is built upon distinct architectural assumptions and generation mechanisms, which lead to different methodological strengths and limitations. These differences are reflected not only in how molecular distributions are modeled, but also in sampling efficiency, representational flexibility, likelihood tractability, and the ability to accommodate complex molecular topology.

	\noindent\textbf{Sampling speed versus expressiveness.}
	A central trade-off among generative paradigms lies in the balance between sampling efficiency and the flexibility of the generative process. VAEs and GANs typically generate molecules through a single decoding or forward pass, enabling fast sampling after training~\cite{lim2018molecular,de2018molgan}. Normalizing flows also support efficient generation through invertible mappings, although continuous-flow variants may require numerical integration~\cite{zang2020moflow,dunn2024mixed}. RNN- and Transformer-based models instead generate molecular sequences autoregressively, introducing sequential decoding costs that increase with sequence length, although Transformers allow substantially greater parallelism during training~\cite{segler2018generating,bagal2021molgpt}. In contrast, diffusion models rely on iterative denoising, which further increases inference cost but allows molecular structures and coordinates to be progressively refined~\cite{hoogeboom2022equivariant}. Thus, direct mappings favor sampling efficiency, whereas autoregressive and diffusion-based generation trade additional inference cost for greater flexibility in sequential or iterative molecular construction.

	\noindent\textbf{Density estimation versus topological flexibility.}
	Normalizing flows provide tractable density estimation through invertible transformations, but the requirement of bijectivity can restrict architectural flexibility and complicate changes in molecular size or topology~\cite{zang2020moflow}. Autoregressive RNN- and Transformer-based models also allow explicit likelihood factorization while accommodating variable-length molecular sequences through sequential token generation, although their autoregressive sampling remains inherently sequential~\cite{segler2018generating,bagal2021molgpt}. Discrete diffusion models can likewise accommodate categorical atom and bond transitions without requiring invertible mappings, providing greater flexibility for variable molecular graphs, but generally rely on iterative generation rather than direct likelihood-based sampling~\cite{vignac2022digress}. Thus, likelihood tractability, sampling efficiency, and topological flexibility represent distinct and often competing design priorities across generative paradigms.

	\subsection{Training Objectives}
	
	The training objectives of molecular generative models can be generally formulated as the optimization of an expected objective over molecular samples $M$ drawn from the empirical data distribution $p_{\mathrm{data}}$. In unconditional molecule generation, deep generative models learn the empirical distribution of molecular structures from molecular samples alone.

	From a unified perspective, these objectives aim to jointly regulate three aspects of molecule generation: reconstruction fidelity, distribution alignment, and structural or physical consistency. Let $\theta$ denote the trainable model parameters, and let $\lambda_1$ and $\lambda_2$ be non-negative weighting hyperparameters that control the relative contributions of the distribution-alignment and consistency terms, respectively. Accordingly, the overall training objective can be expressed in a generic form (Equation~\ref{eq:overall_objective}):
	\begin{equation}
	\min_{\theta}\ \mathcal{L} = \mathbb{E}_{M \sim p_{\mathrm{data}}} \big[ \mathcal{L}_{\mathrm{rec}} + \lambda_1 \mathcal{L}_{\mathrm{dist}} + \lambda_2 \mathcal{L}_{\mathrm{cons}} \big],
	\label{eq:overall_objective}
	\end{equation}
	where $\mathcal{L}_{\mathrm{rec}}$ measures reconstruction fidelity,
	$\mathcal{L}_{\mathrm{dist}}$ aligns generated molecules with the molecular data distribution, and
	$\mathcal{L}_{\mathrm{cons}}$ encodes structural, physical, or binding-related constraints.
	Individual paradigms may omit or instantiate these terms differently depending on their specific training objective.
	For models without explicit pocket conditioning, optimization is performed over molecular samples alone.
	In pocket-conditioned models, the objectives are conditioned on the protein pocket $P$, encouraging generated ligands to satisfy pocket compatibility and interaction constraints.

	Autoregressive sequence models, including RNN- and Transformer-based models, are commonly trained by maximum likelihood, equivalently minimizing the token-level negative log-likelihood (categorical cross-entropy). For a molecular sequence $x=(x_1,\ldots,x_T)$, the objective can be written as (Equation~\ref{eq:autoregressive_objective}):
	\begin{equation}
	\mathcal{L}_{\mathrm{AR}}=-\mathbb{E}_{x\sim p_{\mathrm{data}}}\left[\sum_{t=1}^{T}\log p_{\theta}(x_t\mid x_{<t})\right],
	\label{eq:autoregressive_objective}
	\end{equation}
	where $x_{<t}$ denotes the tokens preceding position $t$. This objective trains the model to predict each molecular token from its preceding context, as used in recurrent SMILES generators and Transformer-based next-token generation~\cite{segler2018generating,bagal2021molgpt}. For conditional sequence generation, the same formulation can be extended as $p_{\theta}(x_t\mid x_{<t},c)$, where $c$ denotes a molecular property, scaffold, target, or other generation condition~\cite{wang2023cmolgpt}.

	In practice, reconstruction-based objectives are used in likelihood-based generative models. For example, variational autoencoders minimize the negative evidence lower bound (negative ELBO) (Equation~\ref{eq:vae_loss}):
	\begin{equation}
	\mathcal{L}_{\mathrm{VAE}} = - \mathbb{E}_{q_\phi(z|M)}[\log p_\theta(M|z)] + D_{\mathrm{KL}}(q_\phi(z|M)\|p(z)),
	\label{eq:vae_loss}
	\end{equation}
	which jointly enforces reconstruction fidelity and latent space regularization, as adopted in latent-variable molecule generation models~\cite{lim2018molecular} and graph-based molecule generation models such as JT-VAE~\cite{jin2018junction}.

	Distribution matching objectives are commonly realized through implicit adversarial learning, where the generator is trained to match the empirical molecular distribution (Equation~\ref{eq:gan_objective}):
	\begin{equation}
	\min_G \max_D\, \mathcal{V}(D,G) = \mathbb{E}_{M\sim p_{\mathrm{data}}} [\log D(M)] + \mathbb{E}_{z\sim p(z)} [\log(1-D(G(z)))].
	\label{eq:gan_objective}
	\end{equation}
	This formulation is widely used in GAN-based molecular generators such as MolGAN, where adversarial training ensures chemical realism of generated molecular graphs~\cite{de2018molgan}.

	Flow-based models encompass mathematically distinct formulations and therefore do not share a single universal training objective. As an illustrative example, continuous normalizing-flow models minimize the negative log-likelihood of a continuous or dequantized molecular representation $M$ (Equation~\ref{eq:flow_objective}):
	\begin{equation}
	\mathcal{L}_{\mathrm{NF}}=-\mathbb{E}_{M\sim p_{\mathrm{data}}}\left[\log p_Z\big(f_\theta(M)\big)+\log\left|\det J_{f_\theta}(M)\right|\right].
	\label{eq:flow_objective}
	\end{equation}
	Here, $p_Z$ denotes the latent prior, $J_{f_\theta}(M)=\partial f_\theta(M)/\partial M$ is the Jacobian matrix, and $M$ denotes a continuous or dequantized molecular representation. This objective enables exact likelihood evaluation and reversible generation, as used in normalizing-flow models such as MoFlow~\cite{zang2020moflow}. Other flow-based formulations, including discrete flows, flow matching, and Bayesian flow networks, employ different training objectives.

	Diffusion-based models minimize denoising or score-matching objectives to learn the reverse process. For continuous molecular variables, a commonly used noise-prediction objective can be written as (Equation~\ref{eq:diffusion_objective}):
	\begin{equation}
	\mathcal{L}_{\mathrm{diff}}=\mathbb{E}_{t,M,\epsilon}\left[w(t)\left\|\epsilon-\epsilon_\theta(M_t,t)\right\|^2\right],
	\label{eq:diffusion_objective}
	\end{equation}
	where $M_t$ denotes the noisy molecular representation at time step $t$, $\epsilon$ is the sampled noise, and $w(t)$ is an optional time-dependent weighting factor. This noise-prediction objective is used in EDM, where the model predicts the joint noise added to atom coordinates and atom features~\cite{hoogeboom2022equivariant}. TargetDiff instead models continuous atom coordinates with Gaussian diffusion and discrete atom types with categorical diffusion, combining an atom-coordinate mean squared error (MSE) loss with an atom-type KL-divergence loss~\cite{guan2023targetdiff}.

	\subsection{Pocket-Conditioned Molecule Generation}
	
	Building on the generative frameworks introduced above, pocket-conditioned molecule generation extends general molecule generation by incorporating the structural and chemical context of a target protein binding pocket~\cite{guan2023targetdiff,guan2024decompdiff,qian2024kgdiff}. In this setting, the pocket is not merely an external label but provides spatial and chemical constraints, including shape complementarity, residue composition, steric boundaries, and chemical environments that influence protein--ligand interactions. The objective is therefore not only to generate chemically valid and drug-like molecules but also to produce candidate ligands that are compatible with a specific target binding environment during generation. In this review, we focus on pocket-conditioned generation methods that treat the input protein pocket as a fixed structural context during molecule generation. Here, pocket conditioning refers to the direct incorporation of pocket information into the molecule generation process, and is distinguished from goal-directed molecular optimization, in which molecular properties are improved through external objectives or post-training optimization strategies~\cite{winter2019efficient,gao2020generative}.

	\subsubsection{Pocket--Ligand Conditioning Interfaces}

	Pocket--ligand conditioning interfaces convert a binding pocket into spatial and chemical constraints that guide atom placement, bond formation, fragment growth, or full-ligand refinement. An effective interface must align protein and ligand features within a shared three-dimensional frame, preserve spatial symmetries, and expose relevant local pocket information throughout generation.

	One approach represents the binding pocket as a three-dimensional spatial field. VoxBind, for example, uses voxelized representations of protein binding sites to encode occupied and unoccupied spatial regions around the pocket~\cite{pinheiro2024voxbind}. Although voxel-based conditioning makes pocket shape and steric exclusion explicit, its spatial precision depends on grid resolution. More commonly, pockets and ligands are represented as geometric objects in continuous 3D space. TargetDiff formulates molecule generation as an equivariant diffusion process in which noisy ligand coordinates and atom features are progressively refined under pocket constraints~\cite{guan2023targetdiff}. Later studies have extended this framework by introducing additional geometric, graph, or interaction-aware constraints to improve conformational quality, molecular connectivity, and protein--ligand interaction modeling~\cite{schneuing2024structure,huang2024pmdm,huang2024binddm,qu2024molcraft}.

	Other methods introduce chemically structured or interaction-aware interfaces. DecompDiff separates ligands into scaffold and arm components and assigns different priors to different molecular parts, allowing the model to better capture localized functional groups and interaction patterns~\cite{guan2024decompdiff}. Related methods introduce chemically meaningful intermediate units or knowledge-guided signals to improve pocket-compatible molecular construction. D3FG employs functional groups as intermediate building blocks, and TAGMol incorporates target-aware molecular fragments~\cite{lin2023d3fg,dorna2024tagmol}. KGDiff and BoKDiff introduce binding-aware knowledge constraints to guide molecule generation toward chemically and structurally compatible pocket interactions~\cite{qian2024kgdiff,yalabadi2025bokdiff}. Similar interaction- or affinity-aware conditioning strategies are adopted in IPDiff, AliDiff, and DiffGui~\cite{huang2024ipdiff,gu2024alidiff,hu2025diffgui}.

	Flow-based conditioning provides an alternative to iterative diffusion. These methods learn continuous transformations from simple prior distributions to pocket-conditioned ligand distributions~\cite{zhang2024flowsbdd,zhou2025paflow}. PAFlow combines flow matching with interaction guidance and atom-number prediction to adapt ligand size and generation trajectories to pocket geometry~\cite{zhou2025paflow}.

	\subsubsection{Pocket-Conditioned Ligand Construction Strategies}

	Pocket-conditioned models can be organized according to how they construct and refine ligands under pocket constraints, rather than solely according to their underlying generative architectures. Direct atom-level generation methods jointly model ligand coordinates, atom types, and molecular connectivity within the binding pocket. Early equivariant diffusion frameworks such as TargetDiff and DiffSBDD~\cite{guan2023targetdiff,schneuing2024structure} established this paradigm, while subsequent models such as MolCRAFT~\cite{qu2024molcraft}, SGEDiff~\cite{gong2025sgediff} and MSIDiff~\cite{zhang2025msidiff}, further improve denoising, geometric consistency, interaction modeling, and conformational quality. Related flow-based methods formulate this process as continuous transport toward a pocket-conditioned ligand distribution~\cite{zhang2024flowsbdd,jiang2024pocketflow,zhou2025paflow,zhang2025molform}.
	
	Another strategy constructs ligands through chemically or structurally meaningful intermediate units. DecompDiff separates scaffold and arm generation and assigns different priors to distinct molecular components~\cite{guan2024decompdiff}, while D3FG and TAGMol introduce functional-group- or motif-aware construction~\cite{lin2023d3fg,dorna2024tagmol}. TopMT-GAN adopts a topology-first strategy, first generating three-dimensional molecular topologies inside the pocket and then assigning atom and bond types through a second generative stage~\cite{wang2025topmt}. These approaches reduce the complexity of whole-molecule generation and better connect molecular construction with localized pocket interactions.

	Many pocket-conditioned models incorporate guidance signals during sampling or generation refinement. These signals differ from post hoc evaluation metrics: they act during generation to bias the model toward more plausible and target-compatible structures. DiffGui provides a representative example of guided generation, where pocket information and molecular properties jointly influence the generation process~\cite{hu2025diffgui}. Other studies also incorporate interaction priors, target-aware constraints, docking-related signals, or sampling-level optimization to improve pocket compatibility and molecular quality~\cite{qian2024kgdiff,yalabadi2025bokdiff,zhou2025paflow}.

	Beyond direct pocket-conditioned generation, protein structural information can also be incorporated through structure-based scoring or optimization loops, or through joint structural modeling. For example, molecular docking can be integrated into generative optimization as a scoring or reward component, thereby providing structure-based feedback to guide molecular optimization~\cite{guo2021dockstream}. Unlike direct pocket conditioning, docking typically provides an external scoring or reward signal rather than directly conditioning the molecular construction process. This distinction is further illustrated by a recent benchmark comparing a REINVENT-based workflow, in which external docking scores were incorporated into reinforcement-learning-based optimization, with direct pocket-conditioned 3D generative models under a common evaluation framework~\cite{liu2024good}. In parallel, co-folding and related joint modeling strategies can also be incorporated into molecule generation workflows to jointly model protein--ligand complex structures and provide structure-based information for molecule generation~\cite{krishna2024generalized,abramson2024accurate}.

	\section{Comparative Performance Analysis}
	
	A central question in molecule generation is how to reliably assess the chemical quality, structural plausibility, and target relevance of generated molecules. Accordingly, this section first summarizes commonly used evaluation metrics and then compares representative general and pocket-conditioned models across widely used benchmarks and commonly reported evaluation settings.

	\subsection{Evaluation Metrics}
	
	Evaluation of molecule generation should consider multiple distinct and complementary levels~\cite{liu2024good,nie2025durian}. Intrinsic chemical quality evaluates whether generated molecules are chemically meaningful as standalone structures. Geometric and conformational fidelity evaluates whether generated three-dimensional structures are physically plausible, whereas target-specific binding metrics assess pocket fit and protein--ligand interaction quality in pocket-conditioned settings.

	\subsubsection{Intrinsic Chemical Quality}
	
	Intrinsic chemical quality checks whether generated molecules are chemically plausible on their own before any target-specific evaluation. These scores are computed with toolkits such as RDKit, are inexpensive and reproducible, and are independent of downstream docking or affinity evaluation.

	\textbf{Validity} is usually reported as the percentage of generated molecules that can be successfully parsed and satisfy basic chemical rules.
	\textbf{Novelty} measures the fraction of generated molecules absent from a reference or training set.
	\textbf{Uniqueness} measures the fraction of non-duplicated molecules within the generated sample.
	\textbf{Diversity} is commonly quantified by pairwise molecular dissimilarity, internal diversity, or scaffold diversity, reflecting whether generated molecules cover varied chemical structures rather than repeated analogues.
	\textbf{Drug-likeness} is commonly estimated by quantitative estimate of drug-likeness (QED), which integrates multiple physicochemical descriptors such as molecular weight, logP, hydrogen-bond donors and acceptors, topological polar surface area, rotatable bonds, aromatic rings, and structural alerts~\cite{bickerton2012quantifying}.
	\textbf{Synthesizability} is often assessed by the synthetic accessibility (SA) score, which combines fragment contributions and molecular complexity penalties to estimate how difficult a molecule is to synthesize~\cite{ertl2009estimation,skoraczynski2023critical}.
	The original SA score is usually reported on a 1--10 scale, where lower values indicate easier synthesis, whereas some molecule-generation benchmarks report a normalized SA score in the range of 0--1, where higher values indicate better synthetic accessibility.

	\subsubsection{Geometric and Conformational Fidelity}
	
	Geometric and conformational fidelity assesses whether generated molecules adopt chemically plausible 3D arrangements, including pocket-conditioned poses or ensembles that sample accessible conformational states.
	Unlike graph-centric validity, reference-based conformational fidelity metrics assess spatial agreement between generated and reference structures. Root-mean-square deviation (RMSD)-based metrics usually require optimal rigid-body alignment, commonly using the Kabsch algorithm.
	Reported numbers are sensitive to hydrogen treatment, stereochemistry, symmetry matching, and whether structures are compared as generated or energy-relaxed under a consistent protocol, so cross-study comparisons require identical evaluation settings.

	\textbf{RMSD} measures the average spatial deviation between matched atoms of a generated structure and a reference structure after optimal rigid-body alignment~\cite{kabsch1976solution,10.1093/bioinformatics/btaa018}:
	\begin{equation}
		\mathrm{RMSD}(X,Y)=\min_{Q,\mathbf{t}}\sqrt{\frac{1}{N}\sum_{i=1}^{N}\left\|Q\mathbf{x}_i+\mathbf{t}-\mathbf{y}_i\right\|_2^2},
		\label{eq:rmsd}
	\end{equation}
	where $X$ and $Y$ denote the generated and reference atomic coordinates, $N$ is the number of matched atoms, and $Q$ and $\mathbf{t}$ are the optimal rotation and translation. Lower RMSD indicates that the generated pose or conformer is closer to the reference structure.

	\textbf{Coverage (COV) and matching (MAT)} are commonly used to compare generated and reference conformer sets~\cite{xu2022geodiff}. \textbf{COV} measures the fraction of reference conformers that can be matched by at least one generated conformer within an RMSD threshold $\delta$ (Equation~\ref{eq:cov}):
	\begin{equation}
		\mathrm{COV\text{-}R}(\mathcal{R},\mathcal{G})=
		\frac{1}{|\mathcal{R}|}
		\left|
		\left\{
		r\in\mathcal{R}\mid
		\min_{g\in\mathcal{G}}\mathrm{RMSD}(r,g)<\delta
		\right\}
		\right|.
		\label{eq:cov}
	\end{equation}
	A higher COV indicates broader coverage of the reference conformational space.

	\textbf{MAT} measures the average minimum root-mean-square deviation from each reference conformer to the closest generated conformer (Equation~\ref{eq:mat})~\cite{xu2022geodiff}:
	\begin{equation}
		\mathrm{MAT\text{-}R}(\mathcal{R},\mathcal{G})=
		\frac{1}{|\mathcal{R}|}
		\sum_{r\in\mathcal{R}}
		\min_{g\in\mathcal{G}}\mathrm{RMSD}(r,g).
		\label{eq:mat}
	\end{equation}
	A lower MAT indicates better conformer-level matching accuracy. Here, $\mathcal{R}$ and $\mathcal{G}$ denote the reference and generated conformer sets, respectively.

	\subsubsection{Target-Specific Binding Performance}
	In pocket-conditioned settings, target-specific binding metrics are used to assess whether generated ligands fit the binding site and form plausible protein--ligand interactions~\cite{thomas2021comparison}.
	\textbf{Docking-based scores}, including Vina Score, Vina Min, and Vina Dock, provide fast empirical assessments of binding quality by evaluating ligand poses within a receptor model. Vina Score usually refers to the score of the generated pose, Vina Min refers to the score after local energy minimization, and Vina Dock refers to the score after redocking the ligand into the pocket with AutoDock Vina. Lower, more negative docking scores are generally interpreted as better predicted binding affinity, but these values are sensitive to ligand preparation, receptor preparation, docking box definition, pose sampling, and scoring protocols.
	\textbf{High affinity} is commonly defined as the proportion of generated molecules whose predicted binding scores satisfy a predefined affinity criterion, such as outperforming a reference ligand or passing a docking-score threshold. Higher values indicate that a larger fraction of generated candidates are predicted to bind favorably to the target pocket.

	\begin{table*}[!tp]
		\caption{Summary of Performance of Molecule Generation Models on Benchmark Datasets. Validity, uniqueness, and novelty are reported as mean percentages. ``--'' indicates that the corresponding metric was not reported in the original study.}
		\label{tbl:generation_performance_by_dataset}
		\centering
		\footnotesize
		\setlength{\tabcolsep}{4pt}
		\renewcommand{\arraystretch}{1.05}

		\begin{tabular}{@{}>{\raggedright\arraybackslash}p{0.26\textwidth}
		>{\raggedright\arraybackslash}p{0.09\textwidth}
		>{\centering\arraybackslash}p{0.13\textwidth}
		>{\centering\arraybackslash}p{0.13\textwidth}
		>{\centering\arraybackslash}p{0.13\textwidth}
		>{\raggedright\arraybackslash}p{0.15\textwidth}@{}}
		\toprule
		Model & Year & Validity~(\%)$\uparrow$ & Uniqueness~(\%)$\uparrow$ & Novelty~(\%)$\uparrow$ & Category \\
		\specialrule{\lightrulewidth}{\aboverulesep}{0pt}

		\rowcolor{gray!15}
		\multicolumn{6}{@{}l}{\textbf{Dataset 1: QM9}} \\
		GrammarVAE~\cite{pmlr-v70-kusner17a} & 2017 & 60.20 & 9.30 & 80.90 & VAE-based \\
		ARAE~\cite{hong2019molecular} & 2019 & 86.20 & 93.50 & 37.10 & VAE-based \\
		Graph Transformer VAE~\cite{mitton2021graph} & 2021 & 74.60 & 22.50 & 93.90 & VAE-based \\
		D-MolVAE~\cite{du2022small} & 2022 & 100.00 & 97.80 & 97.36 & VAE-based \\
		TrustMol~\cite{wijaya2025trustworthy} & 2025 & -- & 88.00 & 87.70 & VAE-based \\
		MolGAN~\cite{de2018molgan} & 2018 & 98.10 & 10.40 & 94.20 & GAN-based \\
		EarlGAN~\cite{tang2023earlgan} & 2023 & 94.07 & 86.24 & 70.04 & GAN-based \\
		RL-MolGAN~\cite{li2026reinforcement} & 2025 & 84.80 & 89.30 & 99.60 & GAN-based \\
		MoFlow~\cite{zang2020moflow} & 2020 & 100.00 & 99.20 & 98.03 & Flow-based \\
		GraphAF~\cite{shi2020graphaf} & 2020 & 100.00 & 94.51 & 88.83 & Flow-based \\
		GraphDF~\cite{luo2021graphdf} & 2021 & 100.00 & 97.62 & 98.10 & Flow-based \\
		GeoBFN~\cite{song2024unified} & 2024 & 95.31 & 97.53 & 66.40 & Flow-based \\
		GGFlow~\cite{hou2024ggflow} & 2024 & 100.00 & -- & -- & Flow-based \\
		SemlaFlow~\cite{irwin2024semlaflow} & 2024 & 99.40 & 95.40 & -- & Flow-based \\
		Megalodon-flow~\cite{reidenbach2026modular} & 2026 & 97.60 & -- & -- & Flow-based \\
		EDM~\cite{hoogeboom2022equivariant} & 2022 & 91.90 & 90.50 & 59.90 & Diffusion-based \\
		GDSS~\cite{jo2022score} & 2022 & 100.00 & 97.82 & -- & Diffusion-based \\
		DiGress~\cite{vignac2022digress} & 2022 & 99.00 & 96.20 & -- & Diffusion-based \\
		MDM~\cite{huang2023mdm} & 2023 & 98.60 & 94.60 & 90.00 & Diffusion-based \\
		MiDi~\cite{vignac2023midi} & 2023 & 97.90 & 97.60 & 67.50 & Diffusion-based \\
		VoxMol~\cite{o20233d} & 2023 & 98.70 & -- & -- & Diffusion-based \\
		END~\cite{cornet2024equivariant} & 2024 & 94.80 & -- & -- & Diffusion-based \\
		D2L-OMP~\cite{guo2024diffusing} & 2024 & 100.00 & 99.80 & -- & Diffusion-based \\
		MUDiff~\cite{hua2024mudiff} & 2024 & 98.90 & 99.30 & -- & Diffusion-based \\
		3D-EDiffMG~\cite{xu2025threedediffmg} & 2025 & 99.30 & 90.00 & 98.60 & Diffusion-based \\
		ADiT~\cite{joshi2025allatom} & 2025 & 97.43 & 96.92 & -- & Diffusion-based \\
		TED (2D)~\cite{lu2026bridging} & 2026 & 98.00 & -- & -- & Diffusion-based \\

		\addlinespace[1.5pt]
		\rowcolor{gray!15}
		\multicolumn{6}{@{}l}{\textbf{Dataset 2: ZINC}} \\
		GrammarVAE~\cite{pmlr-v70-kusner17a} & 2017 & 76.36 & 99.55 & 100.00 & VAE-based \\
		JT-VAE~\cite{jin2018junction} & 2018 & 100.00 & 13.94 & 99.43 & VAE-based \\
		ARAE~\cite{hong2019molecular} & 2019 & 90.30 & 100.00 & 100.00 & VAE-based \\
		D-MolVAE~\cite{du2022small} & 2022 & 100.00 & 99.88 & 99.99 & VAE-based \\
		MolGAN~\cite{de2018molgan} & 2018 & 95.30 & 4.30 & 100.00 & GAN-based \\
		EarlGAN~\cite{tang2023earlgan} & 2023 & 96.14 & 99.06 & 99.74 & GAN-based \\
		SpotGAN~\cite{li2023spotgan} & 2023 & 93.26 & 92.78 & 92.75 & GAN-based \\
		RL-MolGAN~\cite{li2026reinforcement} & 2025 & 93.30 & 92.80 & 92.80 & GAN-based \\
		GraphAF~\cite{shi2020graphaf} & 2020 & 100.00 & 99.10 & 100.00 & Flow-based \\
		MoFlow~\cite{zang2020moflow} & 2020 & 100.00 & 99.99 & 100.00 & Flow-based \\
		GraphDF~\cite{luo2021graphdf} & 2021 & 100.00 & 99.16 & 100.00 & Flow-based \\
		GGFlow~\cite{hou2024ggflow} & 2024 & 100.00 & -- & -- & Flow-based \\
		GDSS~\cite{jo2022score} & 2022 & 100.00 & 99.64 & 100.00 & Diffusion-based \\
		D2L-OMP~\cite{guo2024diffusing} & 2023 & 100.00 & 99.80 & 100.00 & Diffusion-based \\
		DiffLinker~\cite{igashov2024equivariant} & 2024 & 93.80 & 24.00 & 30.30 & Diffusion-based \\

		\addlinespace[1.5pt]
		\rowcolor{gray!15}
		\multicolumn{6}{@{}l}{\textbf{Dataset 3: GEOM-Drugs}} \\
		GeoBFN~\cite{song2024unified} & 2024 & 92.08 & -- & -- & Flow-based \\
		SemlaFlow~\cite{irwin2024semlaflow} & 2024 & 93.90 & 100.00 & 99.60 & Flow-based \\
		Megalodon-flow~\cite{reidenbach2026modular} & 2026 & 94.40 & -- & -- & Flow-based \\
		EDM~\cite{hoogeboom2022equivariant} & 2022 & 68.60 & 68.60 & 68.60 & Diffusion-based \\
		MDM~\cite{huang2023mdm} & 2023 & 99.50 & 99.00 & 99.00 & Diffusion-based \\
		MiDi~\cite{vignac2023midi} & 2023 & 77.80 & 100.00 & 100.00 & Diffusion-based \\
		VoxMol~\cite{o20233d} & 2023 & 93.40 & 99.10 & -- & Diffusion-based \\
		DiffLinker~\cite{igashov2024equivariant} & 2024 & 93.50 & 36.70 & 70.70 & Diffusion-based \\
		END~\cite{cornet2024equivariant} & 2024 & 89.20 & -- & -- & Diffusion-based \\
		ADiT~\cite{joshi2025allatom} & 2025 & 95.30 & 100.00 & -- & Diffusion-based \\
		TED (2D)~\cite{lu2026bridging} & 2026 & 91.60 & -- & -- & Diffusion-based \\
		\bottomrule
		\end{tabular}
	\end{table*}
	\begin{table*}[!tp]
		\ContinuedFloat
		\caption{continued}
		\centering
		\footnotesize
		\setlength{\tabcolsep}{4pt}
		\renewcommand{\arraystretch}{1.05}

		\begin{tabular}{@{}>{\raggedright\arraybackslash}p{0.26\textwidth}
		>{\raggedright\arraybackslash}p{0.09\textwidth}
		>{\centering\arraybackslash}p{0.13\textwidth}
		>{\centering\arraybackslash}p{0.13\textwidth}
		>{\centering\arraybackslash}p{0.13\textwidth}
		>{\raggedright\arraybackslash}p{0.15\textwidth}@{}}
		\toprule
		Model & Year & Validity~(\%)$\uparrow$ & Uniqueness~(\%)$\uparrow$ & Novelty~(\%)$\uparrow$ & Category \\
		\specialrule{\lightrulewidth}{\aboverulesep}{0pt}
		\rowcolor{gray!15}
		\multicolumn{6}{@{}l}{\textbf{Dataset 4: MOSES}} \\
		CharRNN~\cite{segler2018generating} & 2018 & 97.50 & 99.90 & 84.20 & RNN-based \\
		REINVENT (Prior)~\cite{blaschke2020reinvent} & 2020 & 98.60 & 100.00 & 78.30 & RNN-based \\
		MolGPT~\cite{bagal2021molgpt} & 2021 & 99.40 & 100.00 & 79.70 & Transformer-based \\
		cMolGPT~\cite{wang2023cmolgpt} & 2023 & 98.80 & 99.90 & -- & Transformer-based \\
		GMTransformer~\cite{wei2023probabilistic} & 2023 & 85.87 & 99.98 & 95.31 & Transformer-based \\
		FSM-DDTR~\cite{monteiro2023fsm} & 2023 & 91.15 & 99.92 & 97.38 & Transformer-based \\
		FragGPT~\cite{yue2024unlocking} & 2024 & 98.30 & 99.90 & 99.40 & Transformer-based \\
		CONSMI~\cite{qian2024consmi} & 2024 & 99.10 & 100.00 & 83.40 & Transformer-based \\
		\bottomrule
		\end{tabular}
	\end{table*}
	\subsection{Performance Comparison}
	QM9 represents a relatively simple small-molecule space, ZINC covers a more complex drug-like chemical space, and GEOM-Drugs extends the comparison to three-dimensional conformational and geometric generation. Together, these benchmarks provide a progressive evaluation setting from basic small-molecule generation to drug-like and 3D molecule generation, while providing a common set of metrics for cross-model comparison. We therefore summarize the performance of representative deep generative models on QM9, ZINC, GEOM-Drugs and MOSES in Table~\ref{tbl:generation_performance_by_dataset}. The table is organized by dataset and reports commonly used generation metrics, including validity, uniqueness, and novelty, together with the model category. Since these values are collected from the original model publications and subsequent benchmarking or comparative studies, differences in preprocessing, dataset splits, and evaluation protocols may exist across papers. Therefore, this table is intended as a benchmark-level overview of basic generation capability rather than a direct ranking. When these basic generation metrics approach their upper bounds, small differences should be interpreted cautiously, as they may provide limited evidence of meaningful performance differences~\cite{ash2025practically}.
	
	On QM9 and ZINC, many flow-based and diffusion-based models report high validity, while several VAE- and GAN-based models also achieve competitive results, indicating that generative frameworks are capable of producing chemically valid molecules. For example, D-MolVAE, MoFlow, and other representative models report 100\% validity on QM9, while several models on ZINC achieve validity and novelty values close to or at 100\%. Similar high values are observed for uniqueness on MOSES. These metrics therefore remain useful for characterizing basic generation capability, but their ability to distinguish among well-performing models decreases as performance approaches the upper bound. Accordingly, small numerical differences among near-saturated results should be interpreted cautiously rather than as evidence of practically meaningful superiority~\cite{ash2025practically}. In contrast, larger deviations, such as the relatively low uniqueness of MolGAN and JT-VAE on ZINC, the lower uniqueness and novelty of DiffLinker on ZINC, and the reduced validity of EDM and MiDi on GEOM-Drugs, provide clearer evidence of differences in basic generation performance.

	CrossDocked2020 provides large-scale protein--ligand complex data that jointly capture binding-pocket structures and three-dimensional ligand poses, making it suitable for evaluating pocket-constrained generation and binding-related performance. As one of the most widely used benchmarks for structure-based molecule generation, it also provides broad model coverage and a set of commonly reported target-specific metrics for benchmark-level comparison. For pocket-conditioned molecule generation, we selected 21 representative models that were evaluated on CrossDocked2020 and reported a common set of metrics suitable for comparison. Among them, 5 models are flow-based and 16 are diffusion-based. Although pocket-conditioned VAE- and GAN-based models have also been proposed, they were not included in Table~\ref{tbl:pocket_conditioned_crossdocked2020_performance} because their reported evaluations on CrossDocked2020 did not provide benchmark results that could be incorporated into the quantitative comparison presented here. We summarize the performance of representative pocket-conditioned molecule generation models on CrossDocked2020 in Table~\ref{tbl:pocket_conditioned_crossdocked2020_performance}. Normalized SA scores were calculated from the synthetic accessibility scores proposed by Ertl and Schuffenhauer~\cite{ertl2009estimation} using the normalization scheme $Score_{SA}^{\mathrm{norm}}=(10-Score_{SA})/9$. All values are rounded to two decimal places. ``--'' indicates not reported. The reported values were compiled from original model publications and subsequent comparative studies. Compared with the general metrics in Table~\ref{tbl:generation_performance_by_dataset}, pocket-conditioned evaluation places greater emphasis on target-specific binding and drug-like properties, such as Vina Score, Vina Min, QED, and normalized SA\@. Although these models are commonly evaluated on CrossDocked2020, their reported results may still be affected by differences in data filtering, pocket definition, docking protocols, generated sample size, and post-processing strategies.
	
	The results suggest that recent pocket-conditioned models have improved binding-related performance while maintaining favorable drug-like properties. Among the compiled results, PAFlow reports the lowest Vina Score, Vina Min, and Vina Dock values (-8.31, -8.79, and -9.46, respectively), as well as the highest High Affinity ratio (80.80\%). VoxBind achieves the highest QED score (0.57) and the highest normalized SA score (0.70), indicating favorable drug-likeness and synthetic accessibility, respectively. PocketFlow achieves the highest diversity score (0.87). Collectively, these results highlight the inherent trade-offs among binding affinity, drug-likeness, synthetic accessibility, and molecular diversity in current pocket-conditioned generation models, where improvements in one aspect do not necessarily translate into simultaneous gains across all evaluation dimensions.

	Notably, docking metrics are influenced by multiple aspects of the evaluation protocol, including receptor preparation, search-box definition, and exhaustiveness~\cite{forli2016computational,eberhardt2021autodock}. As summarized in Table~\ref{tbl:pocket_conditioned_crossdocked2020_performance}, five types of docking protocols were used. Type~1 used AutoDock Vina~\cite{eberhardt2021autodock}, with receptors prepared using PDB2PQR and AutoDockTools, a search box defined by extending the generated-ligand bounding box by 5~\AA{} in each dimension, and an exhaustiveness of 16. Type~2 followed the same protocol as Type~1 but used an exhaustiveness of 32. Type~3 used QuickVina2~\cite{alhossary2015fast}, with receptors preprocessed using PyMOL and AutoDockTools, a fixed 20 $\times$ 20 $\times$ 20~\AA{} box centered at the generated molecule, and an exhaustiveness of 16. Type~4 used AutoDock Vina~\cite{eberhardt2021autodock}, but other docking settings were not reported. Type~5 used Glide~\cite{friesner2004glide}, with inner and outer boxes of 13 $\times$ 13 $\times$ 19~\AA{} and 27 $\times$ 27 $\times$ 33~\AA{}, respectively, defined around the ligand. Receptor preparation was not reported. Most of the compiled models followed Type~1, whereas DecompDiff and BoKDiff used Type~2, SGEDiff used Type~3, and FlowSBDD used Type~4. These differences can affect the docking values: receptor preparation alters the chemical representation of the binding site and may shift Vina Score, Vina Min, and Vina Dock values, and consequently affect the derived High Affinity ratio. Search-box definition determines the conformational space explored during docking, while higher exhaustiveness increases the search effort and may increase the likelihood of identifying lower-energy docked poses, thereby potentially leading to more favorable Vina Dock values. Therefore, protocol-dependent variations should be taken into account when interpreting absolute docking values, particularly small differences between models.

	\begin{table}[!t]
		\caption{Summary of Performance of Pocket-Conditioned Molecule Generation Models on CrossDocked2020.}
		\label{tbl:pocket_conditioned_crossdocked2020_performance}
		\centering
		\footnotesize
		\setlength{\tabcolsep}{2pt}
		\renewcommand{\arraystretch}{0.9}
		\begin{tabular}{@{}>{\raggedright\arraybackslash}p{0.16\textwidth}
		>{\centering\arraybackslash}p{0.05\textwidth}
		>{\centering\arraybackslash}p{0.11\textwidth}
		>{\centering\arraybackslash}p{0.10\textwidth}
		>{\centering\arraybackslash}p{0.11\textwidth}
		>{\centering\arraybackslash}p{0.08\textwidth}
		>{\centering\arraybackslash}p{0.08\textwidth}
		>{\centering\arraybackslash}p{0.13\textwidth}
		>{\centering\arraybackslash}p{0.09\textwidth}@{}}
		\toprule
		Model & Type & Vina Score~$\downarrow$ & Vina Min~$\downarrow$ & Vina Dock~$\downarrow$ & QED~$\uparrow$ & SA~$\uparrow$ & High Affinity~$\uparrow$ & Diversity~$\uparrow$ \\
		\midrule
		TargetDiff~\cite{guan2023targetdiff} & 1 & -5.47 & -6.64 & -7.80 & 0.48 & 0.58 & 58.10\% & 0.72 \\
		D3FG~\cite{lin2023d3fg} & 1 & -- & -2.59 & -6.78 & 0.49 & 0.66 & -- & -- \\
		DecompDiff~\cite{guan2024decompdiff} & 2 & -5.67 & -7.04 & -8.39 & 0.45 & 0.61 & 64.40\% & 0.68 \\
		IPDiff~\cite{huang2024ipdiff} & 1 & -6.42 & -7.45 & -8.57 & 0.52 & 0.61 & 69.50\% & 0.74 \\
		TAGMol~\cite{dorna2024tagmol} & 1 & -7.02 & -7.95 & -8.59 & 0.55 & 0.56 & 69.80\% & 0.69 \\
		AliDiff~\cite{gu2024alidiff} & 1 & -7.07 & -8.09 & -8.90 & 0.50 & 0.57 & 73.40\% & 0.73 \\
		BindDM~\cite{huang2024binddm} & 1 & -5.92 & -7.29 & -8.41 & 0.51 & 0.58 & 64.80\% & 0.75 \\
		DiffSBDD~\cite{schneuing2024structure} & 1 & -- & -2.15 & -5.53 & 0.49 & 0.34 & -- & -- \\
		VoxBind~\cite{pinheiro2024voxbind} & 1 & -6.94 & -7.54 & -8.30 & 0.57 & 0.70 & 71.30\% & 0.73 \\
		PMDM~\cite{huang2024pmdm} & 1 & -4.87 & -6.17 & -7.28 & 0.51 & 0.61 & -- & -- \\
		KGDiff~\cite{qian2024kgdiff} & 1 & -8.04 & -8.78 & -9.43 & 0.51 & 0.54 & 79.20\% & -- \\
		FlowSBDD~\cite{zhang2024flowsbdd} & 4 & -3.62 & -6.72 & -8.50 & 0.47 & 0.51 & 63.40\% & 0.75 \\
		MolCRAFT~\cite{qu2024molcraft} & 1 & -6.59 & -7.27 & -7.92 & 0.50 & 0.69 & 59.10\% & 0.73 \\
		PocketFlow~\cite{jiang2024pocketflow} & 5 & -- & -- & -- & 0.51 & -- & -- & 0.87 \\
		IRDiff~\cite{huang2024interaction} & 1 & -6.03 & -7.27 & -8.42 & 0.53 & 0.59 & 67.40\% & 0.72 \\
		DiffGui~\cite{hu2025diffgui} & 1 & -5.90 & -6.89 & -7.90 & 0.50 & 0.65 & -- & -- \\
		SGEDiff~\cite{gong2025sgediff} & 3 & -5.64 & -- & -7.40 & -- & -- & 57.94\% & -- \\
		MSIDiff~\cite{zhang2025msidiff} & 1 & -6.36 & -7.47 & -8.55 & 0.52 & 0.54 & 68.60\% & 0.71 \\
		BoKDiff~\cite{yalabadi2025bokdiff} & 2 & -5.92 & -7.50 & -8.58 & 0.48 & 0.60 & -- & -- \\
		PAFlow~\cite{zhou2025paflow} & 1 & -8.31 & -8.79 & -9.46 & 0.49 & 0.57 & 80.80\% & 0.71 \\
		MolFORM~\cite{zhang2025molform} & 1 & -5.42 & -6.42 & -7.50 & 0.48 & 0.60 & -- & 0.78 \\
		\bottomrule
		\end{tabular}
	\end{table}
	
	\subsection{Practical Implications}


	In practical applications, benchmark improvements should be interpreted together with synthetic accessibility, sampling cost, receptor assumptions, protocol reproducibility, and realistic experimental constraints. More favorable docking scores do not necessarily imply experimentally viable candidates, particularly when receptor preparation, docking settings, and post-processing differ across studies. Moreover, excessive reliance on empirical scoring functions such as AutoDock Vina may yield favorable docking scores for poses with unrealistic geometries or severe steric clashes, thereby creating a disconnect between better docking metrics and actual physical or biological plausibility~\cite{weller2024structure}. Benchmark results are therefore most informative when used to identify task-specific strengths and limitations rather than to establish a universal model ranking.

	\begin{sidewaystable}
	\caption{Summary of Representative Case Studies of Molecular Generative Models.}
	\label{tbl:representative_case_studies}
	\centering
	\scriptsize
	\setlength{\tabcolsep}{1.7pt}
	\renewcommand{\arraystretch}{1.15}

	\begin{tabular}{@{}
		>{\raggedright\arraybackslash}m{0.10\textheight}
		>{\centering\arraybackslash}m{0.11\textheight}
		>{\raggedright\arraybackslash}m{0.075\textheight}
		>{\centering\arraybackslash}m{0.180\textheight}
		>{\centering\arraybackslash}m{0.05\textheight}
		>{\centering\arraybackslash}m{0.05\textheight}
		>{\centering\arraybackslash}m{0.05\textheight}
		>{\centering\arraybackslash}m{0.05\textheight}
		>{\centering\arraybackslash}m{0.12\textheight}
		>{\raggedright\arraybackslash}m{0.15\textheight}
		@{}}
	\toprule

	\textbf{Model}
	& \textbf{Model Type}
	& \textbf{SMILES}
	& \textbf{Molecular Structure}
	& \textbf{Target\newline PDB ID}
	& \textbf{In vitro}
	& \textbf{X-ray}
	& \textbf{In vivo}
	& \textbf{Novelty Status}
	& \textbf{Metrics} \\
	\midrule

	PocketFlow~\cite{jiang2024pocketflow}
	& Flow
	& \seqsplit{OC(=O)c1c(Cl)cccc1Oc2ccc(F)cc2}
	& \includegraphics[width=2.3cm,height=2.3cm,keepaspectratio]{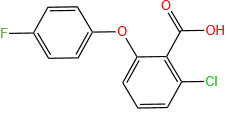}
	& 4R3I
	& \cmark
	& \cmark
	& \xmark
	& Novel molecule
	& IC$_{50}$ = 32.6 $\pm$ 2.72 $\mu$M\newline
	$\Delta T_m$ = 1.08 $^\circ$C\newline
	$K_D$ = 108 $\mu$M \\
	\midrule

	DiffLinker~\cite{igashov2024equivariant}
	& Diffusion
	& \seqsplit{C[C@H](Oc1ccc(cc1)c2cn(C)cn2)C(=O)Nc3[nH]c(cn3)c4ccc(Br)cc4}
	& \includegraphics[width=2.3cm,height=2.3cm,keepaspectratio]{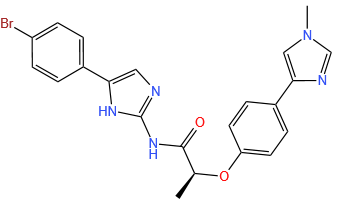}
	& 5OU2
	& \xmark
	& \xmark
	& \xmark
	& Reported molecule (PMID: 29547284)
	& -- \\
	\midrule

	DiffGui~\cite{hu2025diffgui}
	& Diffusion
	& \seqsplit{O=C1NCC2(CCCCC2)n2c1cc1cnc(Nc3cnc4[nH]ccc4c3)nc12}
	& \includegraphics[width=2.4cm,height=2.4cm,keepaspectratio]{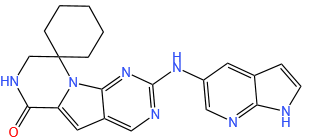}
	& 6G77
	& \cmark
	& \xmark
	& \xmark
	& Novel molecule
	& IC$_{50}$ $\approx$ 111.1 nM \\
	\midrule

	PMDM~\cite{huang2024pmdm}
	& Diffusion
	& \seqsplit{CC(C)NC(=O)O[C@@H]1CCC(C2=CC(NC3=NC(CO)=CN=C3)=NN2)C1}
	& \includegraphics[width=2.1cm,height=2.1cm,keepaspectratio]{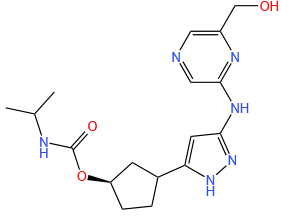}
	& 8H6T
	& \cmark
	& \xmark
	& \xmark
	& Novel molecule
	& IC$_{50}$ = 0.253 nM \\
	\midrule

	REINVENT~\cite{persico2026applying}
	& RNN
	& \seqsplit{NC1=NC(C2=NC=CC(C)=C2)=CC(C3=C(OC)C=CC=C3)=N1}
	& \includegraphics[width=2cm,height=2cm,keepaspectratio]{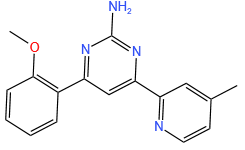}
	& 4EIY
	& \cmark
	& \xmark
	& \xmark
	& Novel molecule
	& IC$_{50}$ $>$ 30 $\mu$M\newline
	$K_i$ = 525 nM \\
	\midrule

	REINVENT~\cite{giblin2026generative}
	& RNN
	& \seqsplit{n41c(nc(c1C)-c3c2ncn(c2cnc3)C)c(cc(c4)-c5ccc(cc5)N6CCOCC6)C(=O)N}
	& \includegraphics[width=2.9cm,height=2.9cm,keepaspectratio]{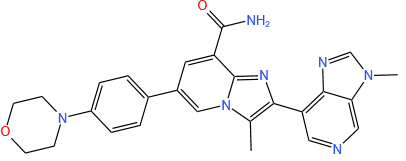}
	& ---
	& \cmark
	& \cmark
	& \xmark
	& Novel molecule
	& IC$_{50}$ = 33.3 nM \\
	\midrule

	REINVENT~\cite{quinn2024accelerated}
	& RNN
	& \seqsplit{O=C(C1=NC(C2CC2)=NC(C)=C1)NC3=CC=CC([C@]4(C)CCNC(O4)=O)=C3}
	& \includegraphics[width=2.5cm,height=2.5cm,keepaspectratio]{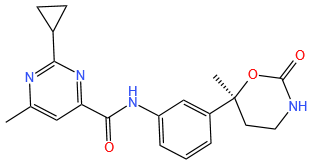}
	& 9FQI
	& \cmark
	& \cmark
	& \xmark
	& Novel molecule
	& IC$_{50}$ = 4,800 nM \\
	\midrule

	GDL~\cite{li2022generative}
	& RNN
	& \seqsplit{CC(C)C(=O)NC1=NN=C2C=C(C3=CC=C4C(=C3)C(C(=O)NC(C)C3=CC=CC=C3)=CN4C)C=CN12}
	& \includegraphics[width=2.5cm,height=2.5cm,keepaspectratio]{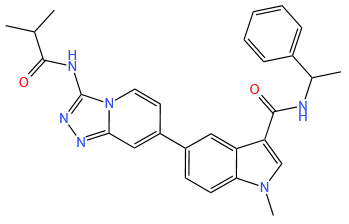}
	& 4ITH
	& \cmark
	& \cmark
	& \cmark
	& Novel molecule
	& IC$_{50}$ = 35.0 $\pm$ 4.8 nM ($^{33}$P);\newline
	5.9 $\pm$ 2.5 nM (ADP-Glo) \\
	\bottomrule

	\end{tabular}
	\end{sidewaystable}

	Beyond benchmark-level comparisons, we curated eight representative case studies to examine how molecular generative models have been applied in experimentally oriented drug-design workflows, as summarized in Table~\ref{tbl:representative_case_studies}. The cases were selected because they provide compound-level experimental evidence and cover different molecular design strategies, biological targets, and levels of experimental validation~\cite{jiang2024pocketflow,hu2025diffgui,huang2024pmdm}. Three REINVENT studies investigated distinct targets spanning a receptor implicated in neurodegenerative disease and two cancer-immunotherapy targets from kinase and E3 ubiquitin ligase families~\cite{persico2026applying,giblin2026generative,quinn2024accelerated}. GDL further provides an extensively validated case targeting RIPK1~\cite{li2022generative}, whereas DiffLinker represents a retrospective example in which the model recovered a previously reported IMPDH inhibitor~\cite{igashov2024equivariant}. For each case, the SMILES representation and two-dimensional molecular structure are provided to identify the selected compound. The corresponding Target PDB ID records the experimentally determined protein structure associated with the target in the original study. Experimental validation is categorized as \textit{in vitro}, X-ray analysis, or \textit{in vivo}. The \textit{in vitro} category includes biochemical assays that directly measure inhibition of purified proteins or enzymes, biophysical assays that assess how strongly a compound binds to its target, cell-based assays that evaluate functional effects in cultured cells, and receptor-binding assays that quantify ligand affinity for a receptor. X-ray crystallography provides direct structural evidence by determining protein--ligand complex structures from diffraction data~\cite{maveyraud2020protein}. The \textit{in vivo} category refers to evaluation in animal models, such as pharmacokinetic, tolerability, or efficacy studies. Target-specific metrics quantify experimentally measured activity or binding. For example, IC$_{50}$ is estimated from concentration--response data as the concentration producing 50\% inhibition~\cite{sebaugh2011guidelines}.

	Experimental validation of generative molecular design has progressed beyond computational scoring, but remains largely at the early preclinical stage. Seven of the eight cases include prospective \textit{in vitro} evaluation, and four also underwent X-ray structural characterization~\cite{jiang2024pocketflow,giblin2026generative,quinn2024accelerated,li2022generative}. Among these cases, GDL provides the most extensive validation pipeline. The generated RIPK1 inhibitor RI-962 was first synthesized and tested in \textit{in vitro} biochemical assays, followed by X-ray structural analysis and further validation in cellular necroptosis models. It was then evaluated \textit{in vivo} for pharmacokinetics and tolerability, and subsequently tested for therapeutic efficacy in mouse models of inflammatory diseases~\cite{li2022generative}. In contrast, the remaining cases largely stop at biochemical, biophysical, cellular, or structural validation. Thus, although generative models have produced experimentally active and structurally validated molecules, progression to \textit{in vivo} efficacy remains uncommon, and none of the cases summarized here has been reported to enter human clinical evaluation. Further medicinal-chemistry optimization, pharmacokinetic and safety assessment, and preclinical development are still needed before such molecules can advance toward clinical candidates.

	\section{Future Directions}
	
	Looking ahead, progress in molecule generation will depend not only on model architecture but also on data quality, task definition, and methodological consistency. Future datasets should move beyond static single-pocket snapshots to better represent receptor conformational heterogeneity~\cite{wankowicz2022ligand}, while incorporating curated protein--ligand complexes with binding-related annotations~\cite{wagle2023sunsetting}. Standardized benchmarks should also adopt consistent data splits and binding-relevant evaluation protocols to improve generalization and cross-study comparability~\cite{francoeur2020three}. Richer three-dimensional corpora integrating crystal structures, computed conformers, and field-based descriptors such as electron density may strengthen the connection between ligand geometry and binding context~\cite{axelrod2022geom,xiang2025edbench,wang2022pocket}. Methodologically, future pipelines should align representation, generation, docking, scoring, and filtering under shared geometric assumptions, while incorporating pharmacophore, shape, electrostatic, and interaction-aware constraints~\cite{huang2024ipdiff,chen2025generating,ragoza2022generating}. Another direction is modeling receptor flexibility. Treating the receptor as fixed simplifies pocket-conditioned generation but limits modeling of ligand-induced conformational changes, including induced-fit effects.
	Extending this direction may enable realistic modeling of receptor adaptation during generation. At the same time, pocket-conditioned generation is likely to expand from single-pocket design to selective polypharmacology, related pocket families, and multi-objective lead optimization, requiring models to balance binding compatibility, drug-likeness, and synthetic accessibility across targets~\cite{flynn2025pharmacoforge,lee2025bind}. Achieving this goal will require geometry-consistent datasets, interaction-aware modeling, integrated generation pipelines, and task-aware evaluation that more effectively connect binding-relevant three-dimensional structures with experimentally actionable molecules.
	
	\section{Conclusions}
	
	In this review, we provided a systematic evaluation of molecule generation models for \textit{de novo} drug design. We summarized representative data resources and molecular and protein-pocket representations, organized RNN/Transformer-based, VAE-based, GAN-based, flow-based, and diffusion-based models according to their generation mechanisms, and reviewed their major training objectives and pocket-conditioned extensions. We further synthesized and comparatively analyzed literature-reported benchmark results using intrinsic chemical, geometric, conformational, and target-specific metrics. The comparison shows that model performance varies across representations, datasets, task settings, and evaluation protocols, and that improvements in one metric do not necessarily translate into consistent gains across other aspects of molecular quality. Reliable assessment therefore requires jointly considering chemical validity, structural plausibility, target relevance, and evaluation consistency.

	Despite this progress, studies still face heterogeneous benchmark settings, simplified receptor and interaction modeling, and insufficient comparability across reported results. Future work should emphasize standardized benchmark construction, consistent data-splitting and reporting protocols, receptor-flexible and interaction-aware generation, synthesis- and property-aware constraints, interpretable modeling, and reproducible cross-dataset evaluation.
	Addressing these challenges will improve cross-study comparability and support the development of generative models that produce chemically plausible, structurally reliable, and target-relevant molecules. More broadly, closer integration of molecule generation with medicinal chemistry, synthesis, and experimental validation will be important for translating methodological advances into practical drug-design applications.

	\section*{Data and Software Availability}
	
	The data supporting the findings of this Review were collected from the
	publications and publicly available resources cited in the manuscript.
	No dedicated software was developed for this Review.
	The compiled benchmark resources, evaluation metrics, model references,
	and available code repositories are accessible at
	\href{https://github.com/JacklinGroup/molecule-generation-review}
	{https://github.com/JacklinGroup/molecule-generation-review}.
	
	\section*{Acknowledgements}
	
	The work is supported in part by the National Natural Science Foundation of China (Grant No. 62573372) and the Hunan Provincial Key Research and Development Program Project (Grant No. 2025JK2003).

	\begingroup
	\sloppy
	\printbibliography

@article{reymond2015chemical,
  title={The chemical space project},
  author={Reymond, Jean-Louis},
  journal={Accounts of chemical research},
  volume={48},
  number={3},
  pages={722--730},
  year={2015},
  publisher={ACS Publications}
}

@article{pang2023deep,
  title={Deep generative models in de novo drug molecule generation},
  author={Pang, Chao and Qiao, Jianbo and Zeng, Xiangxiang and Zou, Quan and Wei, Leyi},
  journal={Journal of Chemical Information and Modeling},
  volume={64},
  number={7},
  pages={2174--2194},
  year={2023},
  publisher={ACS Publications}
}

@article{alakhdar2024diffusion,
  title={Diffusion models in de novo drug design},
  author={Alakhdar, Amira and Poczos, Barnabas and Washburn, Newell},
  journal={Journal of Chemical Information and Modeling},
  volume={64},
  number={19},
  pages={7238--7256},
  year={2024},
  publisher={ACS Publications}
}

@article{bronstein2021geometric,
  title={Geometric deep learning: Grids, groups, graphs, geodesics, and gauges},
  author={Bronstein, Michael M and Bruna, Joan and Cohen, Taco and Veli{\v{c}}kovi{\'c}, Petar},
  journal={arXiv preprint arXiv:2104.13478},
  year={2021}
}

@inproceedings{jin2020hierarchical,
  title={Hierarchical generation of molecular graphs using structural motifs},
  author={Jin, Wengong and Barzilay, Regina and Jaakkola, Tommi},
  booktitle={International conference on machine learning},
  pages={4839--4848},
  year={2020},
  organization={PMLR}
}

@article{guimaraes2017objective,
  title={Objective-reinforced generative adversarial networks (organ) for sequence generation models},
  author={Guimaraes, Gabriel Lima and Sanchez-Lengeling, Benjamin and Outeiral, Carlos and Farias, Pedro Luis Cunha and Aspuru-Guzik, Al{\'a}n},
  journal={arXiv preprint arXiv:1705.10843},
  year={2017}
}

@inproceedings{NEURIPS2021_31445061,
 author = {Luo, Shitong and Guan, Jiaqi and Ma, Jianzhu and Peng, Jian},
 booktitle = {Advances in Neural Information Processing Systems},
 pages = {6229--6239},
 title = {A 3D Generative Model for Structure-Based Drug Design},
 volume = {34},
 year = {2021}
}

@article{xie2022advances,
  title={Advances and challenges in de novo drug design using three-dimensional deep generative models},
  author={Xie, Weixin and Wang, Fanhao and Li, Yibo and Lai, Luhua and Pei, Jianfeng},
  journal={Journal of Chemical Information and Modeling},
  volume={62},
  number={10},
  pages={2269--2279},
  year={2022},
  publisher={ACS Publications}
}

@inproceedings{ganea2021geomol,
  author={Octavian Ganea and Lagnajit Pattanaik and Connor Coley and Regina Barzilay and Klavs F. Jensen and William H. Green and Tommi Jaakkola},
  title={GeoMol: Torsional Geometric Generation of Molecular 3D Conformer Ensembles},
  booktitle={Advances in Neural Information Processing Systems},
  volume={34},
  pages={13757--13769},
  year={2021},
  publisher={Curran Associates, Inc.}
}

@article{sousa2021generative,
  title={Generative deep learning for targeted compound design},
  author={Sousa, Tiago and Correia, Jo{\~a}o and Pereira, V{\'\i}tor and Rocha, Miguel},
  journal={Journal of Chemical Information and Modeling},
  volume={61},
  number={11},
  pages={5343--5361},
  year={2021},
  publisher={ACS Publications}
}

@article{shi2020graphaf,
  title={Graphaf: a flow-based autoregressive model for molecular graph generation},
  author={Shi, Chence and Xu, Minkai and Zhu, Zhaocheng and Zhang, Weinan and Zhang, Ming and Tang, Jian},
  journal={arXiv preprint arXiv:2001.09382},
  year={2020}
}

@inproceedings{zang2020moflow,
  title={Moflow: an invertible flow model for generating molecular graphs},
  author={Zang, Chengxi and Wang, Fei},
  booktitle={Proceedings of the 26th ACM SIGKDD international conference on knowledge discovery \& data mining},
  pages={617--626},
  year={2020}
}

@article{dunn2024mixed,
  title={Mixed continuous and categorical flow matching for 3d de novo molecule generation},
  author={Dunn, Ian and Koes, David Ryan},
  journal={ArXiv},
  pages={arXiv--2404},
  year={2024}
}

@article{song2024unified,
  title={Unified generative modeling of 3d molecules via bayesian flow networks},
  author={Song, Yuxuan and Gong, Jingjing and Qu, Yanru and Zhou, Hao and Zheng, Mingyue and Liu, Jingjing and Ma, Wei-Ying},
  journal={arXiv preprint arXiv:2403.15441},
  year={2024}
}

@article{irwin2024semlaflow,
  title={{{SemlaFlow}: efficient {3D} molecular generation with latent attention and equivariant flow matching}},
  author={Irwin, Ross and Tibo, Alessandro and Janet, Jon Paul and Olsson, Simon},
  journal={arXiv preprint arXiv:2406.07266},
  year={2024}
}

@inproceedings{sohl2015deep,
  title={{Deep unsupervised learning using nonequilibrium thermodynamics}},
  author={Sohl-Dickstein, Jascha and Weiss, Eric and Maheswaranathan, Niru and Ganguli, Surya},
  booktitle={International Conference on Machine Learning},
  volume={37},
  pages={2256--2265},
  year={2015},
  organization={pmlr}
}

@inproceedings{ho2020denoising,
  title={{Denoising diffusion probabilistic models}},
  author={Ho, Jonathan and Jain, Ajay and Abbeel, Pieter},
  booktitle={Advances in Neural Information Processing Systems},
  volume={33},
  pages={6840--6851},
  year={2020}
}

@article{song2020score,
  title={Score-based generative modeling through stochastic differential equations},
  author={Song, Yang and Sohl-Dickstein, Jascha and Kingma, Diederik P and Kumar, Abhishek and Ermon, Stefano and Poole, Ben},
  journal={arXiv preprint arXiv:2011.13456},
  year={2020}
}

@article{weininger1988smiles,
  title={SMILES, a chemical language and information system. 1. Introduction to methodology and encoding rules},
  author={Weininger, David},
  journal={Journal of chemical information and computer sciences},
  volume={28},
  number={1},
  pages={31--36},
  year={1988},
  publisher={ACS Publications}
}

@article{bjerrum2017smiles,
  title={{{SMILES} enumeration as data augmentation for neural network modeling of molecules}},
  author={Bjerrum, Esben Jannik},
  journal={arXiv preprint arXiv:1703.07076},
  year={2017}
}

@article{krenn2022selfies,
  title={{{SELFIES} and the future of molecular string representations}},
  author={Krenn, Mario and Ai, Qianxiang and Barthel, Senja and Carson, Nessa and Frei, Angelo and Frey, Nathan C and Friederich, Pascal and Gaudin, Th{\'e}ophile and Gayle, Alberto Alexander and Jablonka, Kevin Maik and others},
  journal={Patterns},
  volume={3},
  number={10},
  year={2022},
  pages = {100588},
}

@article{o2018deepsmiles,
  title={DeepSMILES: an adaptation of SMILES for use in machine-learning of chemical structures},
  author={O'Boyle, Noel and Dalke, Andrew},
  year={2018},
  journal={ChemRxiv}
}

@article{heller2013inchi,
  title={InChI-the worldwide chemical structure identifier standard},
  author={Heller, Stephen and McNaught, Alan and Stein, Stephen and Tchekhovskoi, Dmitrii and Pletnev, Igor},
  journal={Journal of cheminformatics},
  volume={5},
  number={1},
  pages={7},
  year={2013},
  publisher={Springer}
}

@inproceedings{gilmer2017neural,
  title={{Neural message passing for quantum chemistry}},
  author={Gilmer, Justin and Schoenholz, Samuel S and Riley, Patrick F and Vinyals, Oriol and Dahl, George E},
  booktitle={International Conference on Machine Learning},
  volume={70},
  pages={1263--1272},
  year={2017},
  organization={Pmlr}
}

@article{zhou2020graph,
  title={{Graph neural networks: a review of methods and applications}},
  author={Zhou, Jie and Cui, Ganqu and Hu, Shengding and Zhang, Zhengyan and Yang, Cheng and Liu, Zhiyuan and Wang, Lifeng and Li, Changcheng and Sun, Maosong},
  journal={AI Open},
  volume={1},
  pages={57--81},
  year={2020},
}

@article{velivckovic2017graph,
  title={Graph attention networks},
  author={Veli{\v{c}}kovi{\'c}, Petar and Cucurull, Guillem and Casanova, Arantxa and Romero, Adriana and Lio, Pietro and Bengio, Yoshua},
  journal={arXiv preprint arXiv:1710.10903},
  year={2017}
}

@inproceedings{xiang2025edbench,
  title={{{EDBench}: large-scale electron density data for molecular modeling}},
  author={Xiang, Hongxin and Li, Ke and Liu, Mingquan and Cheng, Zhixiang and Yao, Bin and Du, Wenjie and Xia, Jun and Zeng, Li and Jin, Xin and others},
  booktitle={Advances in Neural Information Processing Systems},
  volume={38},
  year={2025}
}

@article{francoeur2020three,
  title={Three-dimensional convolutional neural networks and a cross-docked data set for structure-based drug design},
  author={Francoeur, Paul G and Masuda, Tomohide and Sunseri, Jocelyn and Jia, Andrew and Iovanisci, Richard B and Snyder, Ian and Koes, David R},
  journal={Journal of Chemical Information and Modeling},
  volume={60},
  number={9},
  pages={4200--4215},
  year={2020},
  publisher={ACS Publications}
}

@article{polykovskiy2020molecular,
  title={Molecular sets (MOSES): a benchmarking platform for molecular generation models},
  author={Polykovskiy, Daniil and Zhebrak, Alexander and Sanchez-Lengeling, Benjamin and Golovanov, Sergey and Tatanov, Oktai and Belyaev, Stanislav and Kurbanov, Rauf and Artamonov, Aleksey and Aladinskiy, Vladimir and Veselov, Mark and others},
  journal={Frontiers in pharmacology},
  volume={11},
  pages={565644},
  year={2020},
  publisher={Frontiers}
}

@article{ragoza2022generating,
  title={Generating 3D molecules conditional on receptor binding sites with deep generative models},
  author={Ragoza, Matthew and Masuda, Tomohide and Koes, David Ryan},
  journal={Chemical Science},
  volume={13},
  number={9},
  pages={2701--2713},
  year={2022},
  publisher={Royal Society of Chemistry}
}

@inproceedings{o20233d,
  title={{{3D} molecule generation by denoising voxel grids}},
  author={O Pinheiro, Pedro O and Rackers, Joshua and Kleinhenz, Joseph and Maser, Michael and Mahmood, Omar and Watkins, Andrew and Ra, Stephen and Sresht, Vishnu and Saremi, Saeed},
  booktitle={Advances in Neural Information Processing Systems},
  volume={36},
  pages={69077--69097},
  year={2023}
}

@article{islam2025conditioned,
  title={{Conditioned generative modeling of molecular glues: a realistic {AI} approach for synthesizable drug-like molecules}},
  author={Islam, Naeyma N and Caulfield, Thomas R},
  journal={Biomolecules},
  volume={15},
  number={6},
  pages={849},
  year={2025},
}

@article{de2018molgan,
  title={{{MolGAN}: an implicit generative model for small molecular graphs}},
  author={De Cao, Nicola and Kipf, Thomas},
  journal={arXiv preprint arXiv:1805.11973},
  doi = {10.48550/arXiv.1805.11973},
  year={2018}
}

@inproceedings{hoogeboom2022equivariant,
  title={{Equivariant diffusion for molecule generation in {3D}}},
  author={Hoogeboom, Emiel and Satorras, V{\i}ctor Garcia and Vignac, Cl{\'e}ment and Welling, Max},
  booktitle={International Conference on Machine Learning},
  volume={162},
  pages={8867--8887},
  year={2022},
  organization={PMLR}
}

@article{xu2022geodiff,
  title={Geodiff: A geometric diffusion model for molecular conformation generation},
  author={Xu, Minkai and Yu, Lantao and Song, Yang and Shi, Chence and Ermon, Stefano and Tang, Jian},
  journal={arXiv preprint arXiv:2203.02923},
  year={2022}
}

@inproceedings{hua2024mudiff,
  title={{{MUDiff}: unified diffusion for complete molecule generation}},
  author={Hua, Chenqing and Luan, Sitao and Xu, Minkai and Ying, Zhitao and Fu, Jie and Ermon, Stefano and Precup, Doina},
  booktitle={Learning on Graphs Conference},
  volume={231},
  pages={33--1},
  year={2024},
  organization={PMLR}
}

@inproceedings{vignac2023midi,
  title={{{MiDi}: mixed graph and {3D} denoising diffusion for molecule generation}},
  author={Vignac, Clement and Osman, Nagham and Toni, Laura and Frossard, Pascal},
  booktitle={Joint European Conference on Machine Learning and Knowledge Discovery in Databases},
  volume={14170},
  pages={560--576},
  year={2023},
  organization={Springer}
}

@article{lin2025diffbp,
  title={Diffbp: Generative diffusion of 3d molecules for target protein binding},
  author={Lin, Haitao and Huang, Yufei and Zhang, Odin and Ma, Siqi and Liu, Meng and Li, Xuanjing and Wu, Lirong and Wang, Jishui and Hou, Tingjun and Li, Stan Z},
  journal={Chemical Science},
  volume={16},
  number={3},
  pages={1417--1431},
  year={2025},
  publisher={Royal Society of Chemistry}
}

@article{qian2024kgdiff,
  title={KGDiff: towards explainable target-aware molecule generation with knowledge guidance},
  author={Qian, Hao and Huang, Wenjing and Tu, Shikui and Xu, Lei},
  journal={Briefings in Bioinformatics},
  volume={25},
  number={1},
  pages={bbad435},
  year={2024},
  publisher={Oxford University Press}
}

@inproceedings{huang2024ipdiff,
  title={{Protein-ligand interaction prior for binding-aware {3D} molecule diffusion models}},
  author={Huang, Zhilin and Yang, Ling and Zhou, Xiangxin and Zhang, Zhilong and Zhang, Wentao and Zheng, Xiawu and Chen, Jie and Wang, Yu and Cui, Bin and Yang, Wenming},
  booktitle={The Twelfth International Conference on Learning Representations},
  year={2024}
}

@article{schneuing2024structure,
  title={Structure-based drug design with equivariant diffusion models},
  author={Schneuing, Arne and Harris, Charles and Du, Yuanqi and Didi, Kieran and Jamasb, Arian and Igashov, Ilia and Du, Weitao and Gomes, Carla and Blundell, Tom L and Lio, Pietro and others},
  journal={Nature Computational Science},
  volume={4},
  number={12},
  pages={899--909},
  year={2024},
  publisher={Nature Publishing Group US New York}
}

@article{chen2025generating,
  title={Generating 3D small binding molecules using shape-conditioned diffusion models with guidance},
  author={Chen, Ziqi and Peng, Bo and Zhai, Tianhua and Adu-Ampratwum, Daniel and Ning, Xia},
  journal={Nature Machine Intelligence},
  volume={7},
  number={5},
  pages={758--770},
  year={2025},
  publisher={Nature Publishing Group UK London}
}

@article{flynn2025pharmacoforge,
  title={PharmacoForge: pharmacophore generation with diffusion models},
  author={Flynn, Emma L and Shah, Riya and Dunn, Ian and Aggarwal, Rishal and Koes, David Ryan},
  journal={Frontiers in Bioinformatics},
  volume={5},
  pages={1628800},
  year={2025},
  publisher={Frontiers Media SA}
}

@article{lee2025bind,
  title={BInD: Bond and Interaction-Generating Diffusion Model for Multi-Objective Structure-Based Drug Design},
  author={Lee, Joongwon and Zhung, Wonho and Seo, Jisu and Kim, Woo Youn},
  journal={Advanced Science},
  volume={12},
  number={35},
  pages={e02702},
  year={2025},
  publisher={Wiley Online Library}
}

@article{lu2026bridging,
  title={Bridging 2D topology and 3D geometry via Transformer-enhanced Diffusion for Drug Molecule Generation},
  author={Lu, Yunhua and Zhang, Junhao and Zhang, Qingwei and Zhang, Chao and Zhang, Junan},
  journal={Biomedical Signal Processing and Control},
  volume={113},
  pages={109101},
  year={2026},
  publisher={Elsevier}
}

@article{hostavs2025vnflow,
  title={VNFlow: integration of variational autoencoders and normalizing flows for novel molecular design},
  author={Hosta{\v{s}}, Ji{\v{r}}{\'\i} and Ghaemi, Mohammad S and Hu, Hang and Lin, Junan and Hu, Anguang and Ooi, Hsu K},
  journal={Journal of Cheminformatics},
  volume={17},
  number={1},
  pages={1--17},
  year={2025},
  publisher={Springer}
}

@inproceedings{satorras2021n,
  title={{{E(n)} equivariant graph neural networks}},
  author={Satorras, V{\i}ctor Garcia and Hoogeboom, Emiel and Welling, Max},
  booktitle={International Conference on Machine Learning},
  volume={139},
  pages={9323--9332},
  year={2021},
  organization={PMLR}
}

@article{skoraczynski2023critical,
  title={Critical assessment of synthetic accessibility scores in computer-assisted synthesis planning},
  author={Skoraczy{\'n}ski, Grzegorz and Kitlas, Mateusz and Miasojedow, B{\l}a{\.z}ej and Gambin, Anna},
  journal={Journal of Cheminformatics},
  volume={15},
  number={1},
  pages={6},
  year={2023},
  publisher={Springer}
}

@article{10.1093/bioinformatics/btaa018,
  author = {Velázquez-Libera, José Luis and Durán-Verdugo, Fabio and Valdés-Jiménez, Alejandro and Núñez-Vivanco, Gabriel and Caballero, Julio},
  title = {{{LigRMSD}: a web server for automatic structure matching and {RMSD} calculations among identical and similar compounds in protein-ligand docking}},
  journal = {Bioinformatics},
  volume = {36},
  number = {9},
  pages = {2912-2914},
  year = {2020},
  month = {05},
}

@article{thomas2021comparison,
  title={Comparison of structure-and ligand-based scoring functions for deep generative models: a GPCR case study},
  author={Thomas, Morgan and Smith, Robert T and O’Boyle, Noel M and de Graaf, Chris and Bender, Andreas},
  journal={Journal of cheminformatics},
  volume={13},
  number={1},
  pages={39},
  year={2021},
  publisher={Springer}
}

@article{wankowicz2022ligand,
  title={{Ligand binding remodels protein side-chain conformational heterogeneity}},
  author={Wankowicz, Stephanie A and de Oliveira, Saulo H and Hogan, Daniel W and van den Bedem, Henry and Fraser, James S},
  journal={eLife},
  volume={11},
  pages={e74114},
  year={2022},
}

@inproceedings{jin2018junction,
  title={{Junction tree variational autoencoder for molecular graph generation}},
  author={Jin, Wengong and Barzilay, Regina and Jaakkola, Tommi},
  booktitle={International Conference on Machine Learning},
  volume={80},
  pages={2323--2332},
  year={2018},
  organization={PMLR}
}

@article{kipf2016semi,
  title={Semi-supervised classification with graph convolutional networks},
  author={Kipf, Thomas N and Welling, Max},
  journal={arXiv preprint arXiv:1609.02907},
  year={2016}
}

@article{elton2019deep,
  title={Deep learning for molecular design-a review of the state of the art},
  author={Elton, Daniel C and Boukouvalas, Zois and Fuge, Mark D and Chung, Peter W},
  journal={Molecular Systems Design \& Engineering},
  volume={4},
  number={4},
  pages={828--849},
  year={2019},
  publisher={Royal Society of Chemistry}
}

@article{xu2019deep,
  title={Deep learning for molecular generation},
  author={Xu, Youjun and Lin, Kangjie and Wang, Shiwei and Wang, Lei and Cai, Chenjing and Song, Chen and Lai, Luhua and Pei, Jianfeng},
  journal={Future medicinal chemistry},
  volume={11},
  number={6},
  pages={567--597},
  year={2019},
  publisher={Taylor \& Francis}
}

@article{fil2021beta,
  title={{{beta-VAE} reproducibility: challenges and extensions}},
  author={Fil, Miroslav and Mesinovic, Munib and Morris, Matthew and Wildberger, Jonas},
  journal={arXiv preprint arXiv:2112.14278},
  year={2021}
}

@article{wang2022pocket,
  title={A pocket-based 3D molecule generative model fueled by experimental electron density},
  author={Wang, Lvwei and Bai, Rong and Shi, Xiaoxuan and Zhang, Wei and Cui, Yinuo and Wang, Xiaoman and Wang, Cheng and Chang, Haoyu and Zhang, Yingsheng and Zhou, Jielong and others},
  journal={Scientific Reports},
  volume={12},
  number={1},
  pages={15100},
  year={2022},
  publisher={Nature Publishing Group UK London}
}

@inproceedings{li2023spotgan,
  title={SpotGAN: A reverse-transformer GAN generates scaffold-constrained molecules with property optimization},
  author={Li, Chen and Yamanishi, Yoshihiro},
  booktitle={Joint European Conference on Machine Learning and Knowledge Discovery in Databases},
  pages={323--338},
  year={2023},
  organization={Springer}
}

@article{macedo2024medgan,
  title={MedGAN: optimized generative adversarial network with graph convolutional networks for novel molecule design},
  author={Macedo, Bruno and Ribeiro Vaz, In{\^e}s and Taveira Gomes, Tiago},
  journal={Scientific Reports},
  volume={14},
  number={1},
  pages={1212},
  year={2024},
  publisher={Nature Publishing Group UK London}
}

@inproceedings{li2024tengan,
  title={{{TenGAN}: pure transformer encoders make an efficient discrete {GAN} for de novo molecular generation}},
  author={Li, Chen and Yamanishi, Yoshihiro},
  booktitle={International Conference on Artificial Intelligence and Statistics},
  volume={238},
  pages={361--369},
  year={2024},
  organization={PMLR}
}

@article{wang2025topmt,
  title={TopMT-GAN: a 3D topology-driven generative model for efficient and diverse structure-based ligand design},
  author={Wang, Shen and Lin, Tong and Peng, Tianyi and Xing, Enming and Chen, Sijie and Kara, Levent Burak and Cheng, Xiaolin},
  journal={Chemical Science},
  volume={16},
  number={6},
  pages={2796--2809},
  year={2025},
  publisher={Royal Society of Chemistry}
}

@article{samanta2020vae,
  title={VAE-Sim: a novel molecular similarity measure based on a variational autoencoder},
  author={Samanta, Soumitra and O'Hagan, Steve and Swainston, Neil and Roberts, Timothy J and Kell, Douglas B},
  journal={Molecules},
  volume={25},
  number={15},
  pages={3446},
  year={2020},
}

@inproceedings{yan2020re,
  title={Re-balancing variational autoencoder loss for molecule sequence generation},
  author={Yan, Chaochao and Wang, Sheng and Yang, Jinyu and Xu, Tingyang and Huang, Junzhou},
  booktitle={Proceedings of the 11th ACM international conference on bioinformatics, computational biology and health informatics},
  pages={1--7},
  year={2020}
}

@inproceedings{ma2021gf,
  title={GF-VAE: a flow-based variational autoencoder for molecule generation},
  author={Ma, Changsheng and Zhang, Xiangliang},
  booktitle={Proceedings of the 30th ACM international conference on information \& knowledge management},
  pages={1181--1190},
  year={2021}
}

@article{dollar2021attention,
  title={Attention-based generative models for de novo molecular design},
  author={Dollar, Orion and Joshi, Nisarg and Beck, David AC and Pfaendtner, Jim},
  journal={Chemical Science},
  volume={12},
  number={24},
  pages={8362--8372},
  year={2021},
  publisher={Royal Society of Chemistry}
}

@article{zhou2023deep,
  title={Deep generative design of porous organic cages via a variational autoencoder},
  author={Zhou, Jiajun and Mroz, Austin and Jelfs, Kim E},
  journal={Digital Discovery},
  volume={2},
  number={6},
  pages={1925--1936},
  year={2023},
  publisher={Royal Society of Chemistry}
}

@article{yoshikai2024novel,
  title={{A novel molecule generative model of {VAE} combined with transformer for unseen structure generation}},
  author={Yoshikai, Yasuhiro and Mizuno, Tadahaya and Nemoto, Shumpei and Kusuhara, Hiroyuki},
  journal={arXiv preprint arXiv:2402.11950},
  year={2024}
}

@article{xu2025exploring,
  title={Exploring Optimized Organic Fluorophore Search through Experimental Data-Driven Adaptive $\beta$-VAE},
  author={Xu, Yuzhi and Luo, Yongrui and Li, Bo and Jiang, Weikang and Zhang, Jinyu and Wei, Jiangbo and Bai, Hanzhi and Wang, Zhiqiang and Ge, Jiankai and Lin, Ruiming and others},
  journal={Jacs Au},
  volume={5},
  number={7},
  pages={3082--3091},
  year={2025},
  publisher={ACS Publications}
}

@article{wijaya2025trustworthy,
  title={Trustworthy inverse molecular design via alignment with molecular dynamics},
  author={Wijaya, Kevin Tirta and Ansari, Navid and Seidel, Hans-Peter and Babaei, Vahid},
  journal={Advanced Science},
  volume={12},
  number={27},
  pages={2416356},
  year={2025},
  publisher={Wiley Online Library}
}

@article{krenn2020self,
  title={Self-referencing embedded strings (SELFIES): A 100\% robust molecular string representation},
  author={Krenn, Mario and H{\"a}se, Florian and Nigam, AkshatKumar and Friederich, Pascal and Aspuru-Guzik, Alan},
  journal={Machine Learning: Science and Technology},
  volume={1},
  number={4},
  pages={045024},
  year={2020},
  publisher={IOP Publishing}
}

@article{jaeger2018mol2vec,
  title={Mol2vec: unsupervised machine learning approach with chemical intuition},
  author={Jaeger, Sabrina and Fulle, Simone and Turk, Samo},
  journal={Journal of Chemical Information and Modeling},
  volume={58},
  number={1},
  pages={27--35},
  year={2018},
  publisher={ACS Publications}
}

@article{sumita2026molecular,
  title={Molecular Design with Artificial Intelligence: Progress and Perspectives for Small Molecules},
  author={Sumita, Masato and Ishida, Shoichi and Yoshizoe, Kazuki and Tamura, Ryo and Terayama, Kei and Tsuda, Koji},
  journal={Chemical Reviews},
  volume={126},
  number={5},
  pages={3007--3054},
  year={2026},
  publisher={ACS Publications}
}

@article{david2020molecular,
  title={Molecular representations in AI-driven drug discovery: a review and practical guide},
  author={David, Laurianne and Thakkar, Amol and Mercado, Roc{\'\i}o and Engkvist, Ola},
  journal={Journal of cheminformatics},
  volume={12},
  number={1},
  pages={56},
  year={2020},
  publisher={Springer}
}

@article{corso2024graph,
  title={Graph neural networks},
  author={Corso, Gabriele and Stark, Hannes and Jegelka, Stefanie and Jaakkola, Tommi and Barzilay, Regina},
  journal={Nature Reviews Methods Primers},
  volume={4},
  number={1},
  pages={17},
  year={2024},
  publisher={Nature Publishing Group UK London}
}

@article{chithrananda2020chemberta,
  title={ChemBERTa: large-scale self-supervised pretraining for molecular property prediction},
  author={Chithrananda, Seyone and Grand, Gabriel and Ramsundar, Bharath},
  journal={arXiv preprint arXiv:2010.09885},
  year={2020}
}

@article{ramakrishnan2014quantum,
  title={Quantum chemistry structures and properties of 134 kilo molecules},
  author={Ramakrishnan, Raghunathan and Dral, Pavlo O and Rupp, Matthias and Von Lilienfeld, O Anatole},
  journal={Scientific data},
  volume={1},
  number={1},
  pages={1--7},
  year={2014},
  publisher={Nature Publishing Group}
}

@article{axelrod2022geom,
  title={GEOM, energy-annotated molecular conformations for property prediction and molecular generation},
  author={Axelrod, Simon and Gomez-Bombarelli, Rafael},
  journal={Scientific data},
  volume={9},
  number={1},
  pages={185},
  year={2022},
  publisher={Nature Publishing Group UK London}
}

@article{zdrazil2024chembl,
  title={The ChEMBL Database in 2023: a drug discovery platform spanning multiple bioactivity data types and time periods},
  author={Zdrazil, Barbara and Felix, Eloy and Hunter, Fiona and Manners, Emma J and Blackshaw, James and Corbett, Sybilla and De Veij, Marleen and Ioannidis, Harris and Lopez, David Mendez and Mosquera, Juan F and others},
  journal={Nucleic Acids Research},
  volume={52},
  number={D1},
  pages={D1180--D1192},
  year={2024},
  publisher={Oxford University Press}
}

@article{davies2015chembl,
  title={ChEMBL web services: streamlining access to drug discovery data and utilities},
  author={Davies, Mark and Nowotka, Micha{\l} and Papadatos, George and Dedman, Nathan and Gaulton, Anna and Atkinson, Francis and Bellis, Louisa and Overington, John P},
  journal={Nucleic Acids Research},
  volume={43},
  number={W1},
  pages={W612--W620},
  year={2015},
  publisher={Oxford University Press}
}

@article{tingle2023zinc,
  title={ZINC-22- A free multi-billion-scale database of tangible compounds for ligand discovery},
  author={Tingle, Benjamin I and Tang, Khanh G and Castanon, Mar and Gutierrez, John J and Khurelbaatar, Munkhzul and Dandarchuluun, Chinzorig and Moroz, Yurii S and Irwin, John J},
  journal={Journal of Chemical Information and Modeling},
  volume={63},
  number={4},
  pages={1166--1176},
  year={2023},
  publisher={ACS Publications}
}

@article{kim2025pubchem,
  title={PubChem 2025 update},
  author={Kim, Sunghwan and Chen, Jie and Cheng, Tiejun and Gindulyte, Asta and He, Jia and He, Siqian and Li, Qingliang and Shoemaker, Benjamin A and Thiessen, Paul A and Yu, Bo and others},
  journal={Nucleic Acids Research},
  volume={53},
  number={D1},
  pages={D1516--D1525},
  year={2025},
  publisher={Oxford University Press}
}

@article{liu2017forging,
  title={Forging the basis for developing protein--ligand interaction scoring functions},
  author={Liu, Zhihai and Su, Minyi and Han, Li and Liu, Jie and Yang, Qifan and Li, Yan and Wang, Renxiao},
  journal={Accounts of chemical research},
  volume={50},
  number={2},
  pages={302--309},
  year={2017},
  publisher={ACS Publications}
}

@article{knox2024drugbank,
  title={DrugBank 6.0: the DrugBank knowledgebase for 2024},
  author={Knox, Craig and Wilson, Mike and Klinger, Christen M and Franklin, Mark and Oler, Eponine and Wilson, Alex and Pon, Allison and Cox, Jordan and Chin, Na Eun and Strawbridge, Seth A and others},
  journal={Nucleic Acids Research},
  volume={52},
  number={D1},
  pages={D1265--D1275},
  year={2024},
  publisher={Oxford University Press}
}

@article{kondratyev2023generative,
  title={Generative model based on junction tree variational autoencoder for HOMO value prediction and molecular optimization},
  author={Kondratyev, Vladimir and Dryzhakov, Marian and Gimadiev, Timur and Slutskiy, Dmitriy},
  journal={Journal of Cheminformatics},
  volume={15},
  number={1},
  pages={11},
  year={2023},
  publisher={Springer}
}

@article{wagle2023sunsetting,
  title={Sunsetting binding MOAD with its last data update and the addition of 3D-ligand polypharmacology tools},
  author={Wagle, Swapnil and Smith, Richard D and Dominic III, Anthony J and DasGupta, Debarati and Tripathi, Sunil Kumar and Carlson, Heather A},
  journal={Scientific Reports},
  volume={13},
  number={1},
  pages={3008},
  year={2023},
  publisher={Nature Publishing Group UK London}
}

@article{zhang2025graph,
  title   = {{Graph Neural Networks in Modern AI-Aided Drug Discovery}},
  author  = {Zhang, Odin and Lin, Haitao and Zhang, Xujun and Wang, Xiaorui and Wu, Zhenxing and Ye, Qing and Zhao, Weibo and Wang, Jike and Ying, Kejun and Kang, Yu and Hsieh, Chang-Yu and Hou, Tingjun},
  journal = {Chemical Reviews},
  year    = {2025},
  volume  = {125},
  pages   = {10001--10103},
}

@article{wu2018moleculenet,
  title = {{MoleculeNet: A Benchmark for Molecular Machine Learning}},
  author = {Wu, Zhenqin and Ramsundar, Bharath and Feinberg, Evan N. and Gomes, Joseph and Geniesse, Caleb and Pappu, Aneesh S. and Leswing, Karl and Pande, Vijay},
  journal = {Chemical Science},
  volume = {9},
  number = {2},
  pages = {513--530},
  year = {2018},
}

@article{yang2019analyzing,
  title = {{Analyzing Learned Molecular Representations for Property Prediction}},
  author = {Yang, Kevin and Swanson, Kyle and Jin, Wengong and Coley, Connor and Eiden, Philipp and Gao, Hua and Guzman-Perez, Angel and Hopper, Timothy and Kelley, Brian and Mathea, Miriam and Palmer, Andrew and Settels, Volker and Jaakkola, Tommi and Jensen, Klavs and Barzilay, Regina},
  journal = {Journal of Chemical Information and Modeling},
  volume = {59},
  number = {8},
  pages = {3370--3388},
  year = {2019},
}

@article{lim2018molecular,
  title = {{Molecular Generative Model Based on Conditional Variational Autoencoder for de Novo Molecular Design}},
  author = {Lim, Jaechang and Ryu, Seongok and Kim, Jin Woo and Kim, Woo Youn},
  journal = {Journal of Cheminformatics},
  volume = {10},
  number = {1},
  pages = {31},
  year = {2018},
}

@article{sattarov2019novo,
  title = {{De Novo Molecular Design by Combining Deep Autoencoder Recurrent Neural Networks with Generative Topographic Mapping}},
  author = {Sattarov, Boris and Baskin, Igor I. and Horvath, Dragos and Marcou, Gilles and Bjerrum, Esben Jannik and Varnek, Alexandre},
  journal = {Journal of Chemical Information and Modeling},
  volume = {59},
  number = {3},
  pages = {1182--1196},
  year = {2019},
}

@article{nie2025durian,
  title = {{Durian: A Comprehensive Benchmark for Structure-Based 3D Molecular Generation}},
  author = {Nie, Dou and Zhao, Huifeng and Zhang, Odin and Weng, Gaoqi and Zhang, Hui and Jin, Jieyu and Lin, Haitao and Huang, Yufei and Liu, Liwei and Li, Dan and Hou, Tingjun and Kang, Yu},
  journal = {Journal of Chemical Information and Modeling},
  volume = {65},
  number = {1},
  pages = {173--186},
  year = {2025},
}

@inProceedings{pmlr-v70-kusner17a,
  title = 	 {{Grammar Variational Autoencoder}},
  author =       {Matt J. Kusner and Brooks Paige and Jos{\'e} Miguel Hern{\'a}ndez-Lobato},
  booktitle = 	 {Proceedings of the 34th International Conference on Machine Learning},
  pages = 	 {1945--1954},
  year = 	 {2017},
  volume = 	 {70},
  month     = {08}
}

@inproceedings{simonovsky2018graphvae,
  title = {{{GraphVAE}: Towards Generation of Small Graphs Using Variational Autoencoders}},
  author = {Simonovsky, Martin and Komodakis, Nikos},
  booktitle = {Artificial Neural Networks and Machine Learning -- {ICANN} 2018},
  series = {Lecture Notes in Computer Science},
  volume = {11139},
  pages = {412--422},
  year = {2018},
}

@article{hong2019molecular,
  title={Molecular generative model based on an adversarially regularized autoencoder},
  author={Hong, Seung Hwan and Ryu, Seongok and Lim, Jaechang and Kim, Woo Youn},
  journal={Journal of Chemical Information and Modeling},
  volume={60},
  number={1},
  pages={29--36},
  year={2019},
  publisher={ACS Publications}
}

@article{du2022small,
  title = {{Small Molecule Generation via Disentangled Representation Learning}},
  author = {Du, Yuanqi and others},
  journal = {Bioinformatics},
  volume = {38},
  number = {12},
  pages = {3200--3208},
  year = {2022},
}

@inproceedings{podda2020deep,
  title = {{A Deep Generative Model for Fragment-Based Molecule Generation}},
  author = {Podda, Marco and Bacciu, Davide and Micheli, Alessio},
  booktitle = {Proceedings of the Twenty Third International Conference on Artificial Intelligence and Statistics},
  series = {Proceedings of Machine Learning Research},
  volume = {108},
  pages = {2240--2250},
  year = {2020}
}

@article{mitton2021graph,
  title = {{A Graph {VAE} and Graph Transformer Approach to Generating Molecular Graphs}},
  author = {Mitton, Joshua and Senn, Hans M. and Wynne, Klaas and Murray-Smith, Roderick},
  journal = {arXiv preprint arXiv:2104.04345},
  year = {2021}
}

@inproceedings{huang2022threedlinker,
  title = {{{3DLinker}: An {E(3)} Equivariant Variational Autoencoder for Molecular Linker Design}},
  author = {Huang, Yinan and Peng, Xingang and Ma, Jianzhu and Zhang, Muhan},
  booktitle = {Proceedings of the 39th International Conference on Machine Learning},
  series = {Proceedings of Machine Learning Research},
  volume = {162},
  pages = {9280--9294},
  year = {2022}
}

@inproceedings{li2022transorgan,
  title = {{Transformer-Based Objective-Reinforced Generative Adversarial Network to Generate Desired Molecules}},
  author = {Li, Chen and Yamanaka, Chikashige and Kaitoh, Kazuma and Yamanishi, Yoshihiro},
  booktitle = {Proceedings of the Thirty-First International Joint Conference on Artificial Intelligence},
  pages = {3884--3890},
  year = {2022},
  doi = {10.24963/ijcai.2022/539}
}

@article{tang2023earlgan,
  title={EarlGAN: an enhanced actor--critic reinforcement learning agent-driven GAN for de novo drug design},
  author={Tang, Huidong and Li, Chen and Jiang, Shuai and Yu, Huachong and Kamei, Sayaka and Yamanishi, Yoshihiro and Morimoto, Yasuhiko},
  journal={Pattern Recognition Letters},
  volume={175},
  pages={45--51},
  year={2023},
  publisher={Elsevier}
}

@inproceedings{tang2025instgan,
  title={InstGAN: Instant Actor-Critic-Driven GAN for De Novo Molecule Generation and Property Optimization},
  author={Tang, Huidong and Li, Chen and Kamei, Sayaka and Yamanishi, Yoshihiro and Morimoto, Yasuhiko},
  booktitle={Proceedings of the Thirty-Fourth International Joint Conference on Artificial Intelligence},
  pages={6236--6244},
  year={2025}
}

@article{li2026reinforcement,
  title={A reinforcement learning-driven transformer GAN for molecular generation},
  author={Li, Chen and Tang, Huidong and Zhu, Ye and Yamanishi, Yoshihiro},
  journal={Machine Intelligence Research},
  pages={1--22},
  year={2026},
  publisher={Springer}
}

@inproceedings{luo2021graphdf,
  title = {{{GraphDF}: A Discrete Flow Model for Molecular Graph Generation}},
  author = {Luo, Youzhi and Yan, Keqiang and Ji, Shuiwang},
  booktitle = {Proceedings of the 38th International Conference on Machine Learning},
  series = {Proceedings of Machine Learning Research},
  volume = {139},
  year = {2021}
}

@article{lippe2021categorical,
  title = {{Categorical Normalizing Flows via Continuous Transformations}},
  author = {Lippe, Phillip and Nielsen, Efstratios Gavves and Welling, Max},
  journal = {arXiv preprint arXiv:2006.09790},
  year = {2021}
}

@article{hou2024ggflow,
  title = {{{GGFlow}: Generated Graph Flow for Molecular Graph Generation}},
  author = {Hou, Xiaoyang and Zhu, Tian and Ren, Milong and Bu, Dongbo and Gao, Xin and Zhang, Chunming and Sun, Shiwei},
  journal = {arXiv preprint arXiv:2411.05676},
  year = {2024}
}

@article{reidenbach2026modular,
  title = {{Applications of Modular Co-Design for De Novo {3D} Molecule Generation}},
  author = {Reidenbach, Danny and Nikitin, Filipp and Isayev, Olexandr and Paliwal, Saee Gopal},
  journal = {Digital Discovery},
  volume = {5},
  pages = {754},
  year = {2026}
}

@inproceedings{jo2022score,
  title = {{Score-Based Generative Modeling of Graphs via the System of Stochastic Differential Equations}},
  author = {Jo, Jaehyeong and Lee, Seul and Hwang, Sung Ju},
  booktitle = {Proceedings of the 39th International Conference on Machine Learning},
  series = {Proceedings of Machine Learning Research},
  volume = {162},
  pages = {10362--10383},
  year = {2022}
}

@article{guo2024diffusing,
  title={Diffusing on two levels and optimizing for multiple properties: A novel approach to generating molecules with desirable properties},
  author={Guo, Siyuan and Guan, Jihong and Zhou, Shuigeng},
  journal={IEEE/ACM transactions on computational biology and bioinformatics},
  volume={21},
  number={6},
  pages={2050--2063},
  year={2024},
  publisher={IEEE}
}

@article{vignac2022digress,
  title={Digress: Discrete denoising diffusion for graph generation},
  author={Vignac, Clement and Krawczuk, Igor and Siraudin, Antoine and Wang, Bohan and Cevher, Volkan and Frossard, Pascal},
  journal={arXiv preprint arXiv:2209.14734},
  year={2022}
}

@inproceedings{huang2023mdm,
  title = {{{MDM}: Molecular Diffusion Model for {3D} Molecule Generation}},
  author = {Huang, Lei and Zhang, Hengtong and Xu, Tingyang and Wong, Ka-Chun},
  booktitle = {Proceedings of the {AAAI} Conference on Artificial Intelligence},
  volume = {37},
  number = {4},
  pages = {5105--5112},
  year = {2023}
}

@article{igashov2024equivariant,
  title={Equivariant 3D-conditional diffusion model for molecular linker design},
  author={Igashov, Ilia and St{\"a}rk, Hannes and Vignac, Cl{\'e}ment and Schneuing, Arne and Satorras, Victor Garcia and Frossard, Pascal and Welling, Max and Bronstein, Michael and Correia, Bruno},
  journal={Nature Machine Intelligence},
  volume={6},
  number={4},
  pages={417--427},
  year={2024},
  publisher={Nature Publishing Group UK London}
}

@inproceedings{cornet2024equivariant,
  title={Equivariant neural diffusion for molecule generation},
  author={Cornet, Fran{\c{c}}ois and Bartosh, Grigory and Schmidt, Mikkel N and Naesseth, Christian A},
  booktitle={Advances in Neural Information Processing Systems},
  volume={37},
  pages={49429--49460},
  year={2024}
}

@inproceedings{mercatali2024diffusion,
  title={Diffusion twigs with loop guidance for conditional graph generation},
  author={Mercatali, Giangiacomo and Verma, Yogesh and Freitas, Andre and Garg, Vikas},
  booktitle={Advances in Neural Information Processing Systems},
  volume={37},
  pages={137741--137767},
  year={2024}
}

@article{xu2025threedediffmg,
  title={3D-EDiffMG: 3D equivariant diffusion-driven molecular generation to accelerate drug discovery},
  author={Xu, Chao and Liu, Runduo and Yao, Yufen and Huang, Wanyi and Li, Zhe and Luo, Hai-Bin},
  journal={Journal of Pharmaceutical Analysis},
  volume={15},
  number={6},
  pages={101257},
  year={2025},
  publisher={Elsevier}
}

@article{joshi2025allatom,
  title = {{All-Atom Diffusion Transformers: Unified Generative Modelling of Molecules and Materials}},
  author = {Joshi, Chaitanya K. and Fu, Xiang and Liao, Yi-Lun and Gharakhanyan, Vahe and Miller, Benjamin Kurt and Sriram, Anuroop and Ulissi, Zachary W.},
  journal = {arXiv preprint arXiv:2503.03965},
  year = {2025}
}

@article{yuan2025mdrl,
  title={A 3D generation framework using diffusion model and reinforcement learning to generate multi-target compounds with desired properties},
  author={Yuan, Yongna and Pan, Xiaohang and Li, Xiaohong and Zhang, Ruisheng and Su, Wei},
  journal={Journal of Cheminformatics},
  volume={17},
  number={1},
  pages={93},
  year={2025},
  publisher={Springer}
}

@article{yang2025diffmcgen,
  title        = {{{DiffMC-Gen}: A Dual Denoising Diffusion Model for Multi-Conditional Molecular Generation}},
  author       = {Yang, Yuwei and Gu, Shukai and Liu, Bo and Gong, Xiaoqing and Lu, Ruiqiang and Qiu, Jiayue and Yao, Xiaojun and Liu, Huanxiang},
  journal      = {Advanced Science},
  year         = {2025},
  volume       = {12},
  number       = {22},
  pages        = {2417726}
}

@article{guan2023targetdiff,
  title = {{{TargetDiff}: {3D} Equivariant Diffusion for Target-Aware Molecule Generation and Affinity Prediction}},
  author = {Guan, Jiaqi and Qian, Wesley Wei and Peng, Xingang and Su, Yufeng and Peng, Jian and Ma, Jianzhu},
  journal = {arXiv preprint arXiv:2303.03543},
  year = {2023}
}

@article{lin2023d3fg,
  title = {{{D3FG}: Functional-Group-Based Diffusion for Pocket-Specific Molecule Generation and Elaboration}},
  author = {Lin, Haitao and Huang, Yufei and Zhang, Odin and Wu, Lirong and Li, Siyuan and Chen, Zhiyuan and Li, Stan Z.},
  journal = {arXiv preprint arXiv:2306.13769},
  year = {2023}
}

@article{guan2024decompdiff,
  title = {{{DecompDiff}: Diffusion Models with Decomposed Priors for Structure-Based Drug Design}},
  author = {Guan, Jiaqi and Zhou, Xiangxin and Yang, Yuwei and Bao, Yu and Peng, Jian and Ma, Jianzhu and Liu, Qiang and Wang, Liang and Gu, Quanquan},
  journal = {arXiv preprint arXiv:2403.07902},
  year = {2024}
}

@article{dorna2024tagmol,
  title = {{{TAGMol}: Target-Aware Gradient-Guided Molecule Generation}},
  author = {Dorna, Vineeth and Subhalingam, D. and Kolluru, Keshav and Tuli, Shreshth and Singh, Mrityunjay and Singal, Saurabh and Krishnan, N. M. Anoop and Ranu, Sayan},
  journal = {arXiv preprint arXiv:2406.01650},
  year = {2024}
}

@article{gu2024alidiff,
  title = {{{ALiDiff}: Aligning Target-Aware Molecule Diffusion Models with Exact Energy Optimization}},
  author = {Gu, Siyi and Xu, Minkai and Powers, Alexander and Nie, Weili and Geffner, Tomas and Kreis, Karsten and Leskovec, Jure and Vahdat, Arash and Ermon, Stefano},
  journal = {arXiv preprint arXiv:2407.01648},
  year = {2024}
}

@inproceedings{huang2024binddm,
  title = {{{BindDM}: Binding-Adaptive Diffusion Models for Structure-Based Drug Design}},
  author = {Huang, Zhilin and Yang, Ling and Zhang, Zaixi and Zhou, Xiangxin and Bao, Yu and Zheng, Xiawu and Yang, Yuwei and Wang, Yu and Yang, Wenming},
  booktitle = {Proceedings of the {AAAI} Conference on Artificial Intelligence},
  volume = {38},
  number = {11},
  pages = {12621--12629},
  year = {2024}
}

@article{pinheiro2024voxbind,
  title = {{{VoxBind}: Structure-Based Drug Design by Denoising Voxel Grids}},
  author = {Pinheiro, Pedro O. and Jamasb, Arian and Mahmood, Omar and Sresht, Vishnu and Saremi, Saeed},
  journal = {arXiv preprint arXiv:2405.03961},
  year = {2024}
}

@article{huang2024pmdm,
  title = {{{PMDM}: A Dual Diffusion Model Enables {3D} Molecule Generation and Lead Optimization Based on Target Pockets}},
  author = {Huang, Lei and Zhang, Hengtong and Xu, Tingyang and Wong, Ka-Chun},
  journal = {Nature Communications},
  volume = {15},
  pages = {2657},
  year = {2024}
}

@article{zhang2024flowsbdd,
  title = {{{FlowSBDD}: Rectified Flow for Structure-Based Drug Design}},
  author = {Zhang, Daiheng and Gong, Chengyue and Liu, Qiang},
  journal = {arXiv preprint arXiv:2412.01174},
  year = {2024}
}

@article{qu2024molcraft,
  title = {{{MolCRAFT}: Structure-Based Drug Design in Continuous Parameter Space}},
  author = {Qu, Yanru and Qiu, Keyue and Song, Yuxuan and Gong, Jingjing and Han, Jiawei and Zheng, Mingyue and Zhou, Hao and Ma, Wei-Ying},
  journal = {arXiv preprint arXiv:2404.12141},
  year = {2024}
}

@article{jiang2024pocketflow,
  title = {{{PocketFlow}: A Data-and-Knowledge-Driven Structure-Based Molecular Generative Model}},
  author = {Jiang, Yuanyuan and Zhang, Guo and You, Jing and Zhang, Hailin and Yao, Rui and Xie, Huanzhang and Zhang, Liyun and Xia, Ziyi and Dai, Mengzhe and Wu, Yunjie and Li, Linli and Yang, Sheng-Yong},
  journal = {Nature Machine Intelligence},
  volume = {6},
  pages = {326--337},
  year = {2024}
}

@article{hu2025diffgui,
  title = {{{DiffGui}: Target-Aware {3D} Molecular Generation Based on Guided Equivariant Diffusion}},
  author = {Hu, Qiaoyu and Sun, Changzhi and He, Huan and Xu, Jiazheng and Liu, Danlin and Zhang, Wenqing and Shi, Sumeng and Zhang, Kai and Li, Honglin},
  journal = {Nature Communications},
  volume = {16},
  pages = {7928},
  year = {2025}
}

@article{gong2025sgediff,
  title = {{{SGEDiff}: A Subgraph-Enriched Diffusion Model for Structure-Based {3D} Molecular Generation}},
  author = {Gong, Changda and Fang, Jiaojiao and Tang, Yan and Liu, Guixia and Tang, Yun and Li, Weihua},
  journal = {Journal of Cheminformatics},
  volume = {17},
  pages = {175},
  year = {2025}
}

@article{zhang2025msidiff,
  title = {{{MSIDiff}: Multi-Stage Interaction-Aware Diffusion Model for Protein-Specific {3D} Molecule Generation}},
  author = {Zhang, Yaoxiang and Ma, Junteng and Zhang, Ze and Dong, Zhaoyang and Wang, Shuang},
  journal = {Expert Systems with Applications},
  volume = {298},
  pages = {129820},
  year = {2025}
}

@article{yalabadi2025bokdiff,
  title = {{{BoKDiff}: Best-of-{K} Diffusion Alignment for Target-Specific {3D} Molecule Generation}},
  author = {Yalabadi, Ali Khodabandeh and Yazdani-Jahromi, Mehdi and Garibay, Ozlem Ozmen},
  journal = {Bioinformatics Advances},
  volume = {5},
  number = {1},
  pages = {vbaf137},
  year = {2025}
}

@article{zhou2025paflow,
  title = {{{PAFlow}: Prior-Guided Flow Matching for Target-Aware Molecule Design with Learnable Atom Number}},
  author = {Zhou, Jingyuan and Qian, Hao and Tu, Shikui and Xu, Lei},
  journal = {arXiv preprint arXiv:2509.01486},
  year = {2025}
}

@article{zhang2025molform,
  title={MolFORM: Preference-Aligned Multimodal Flow Matching for Structure-Based Drug Design},
  author={Zhang, Daiheng and Zhang, Zhao},
  journal={arXiv preprint arXiv:2507.05503},
  year={2025}
}

@article{10.1093/nar/gkae1091,
  author  = {Burley, Stephen K. and Bhatt, Rusham and Bhikadiya, Charmi and Bi, Chunxiao and Biester, Alison and Biswas, Pratyoy and Bittrich, Sebastian and Blaumann, Santiago and Brown, Ronald and Chao, Henry and Chithari, Vivek Reddy and Craig, Paul A. and Crichlow, Gregg V. and Duarte, Jose M. and Dutta, Shuchismita and Feng, Zukang and Flatt, Justin W. and Ghosh, Sutapa and Goodsell, David S. and Green, Rachel Kramer and Guranovic, Vladimir and Henry, Jeremy and Hudson, Brian P. and Joy, Michael and Kaelber, Jason T. and Khokhriakov, Igor and Lai, Jhih-Siang and Lawson, Catherine L. and Liang, Yuhe and Myers-Turnbull, Douglas and Peisach, Ezra and Persikova, Irina and Piehl, Dennis W. and Pingale, Aditya and Rose, Yana and Sagendorf, Jared and Sali, Andrej and Segura, Joan and Sekharan, Monica and Shao, Chenghua and Smith, James and Trumbull, Michael and Vallat, Brinda and Voigt, Maria and Webb, Ben and Whetstone, Shamara and Wu-Wu, Amy and Xing, Tongji and Young, Jasmine Y. and Zalevsky, Arthur and Zardecki, Christine},
  title   = {{Updated Resources for Exploring Experimentally-Determined {PDB} Structures and Computed Structure Models at the {RCSB} Protein Data Bank}},
  journal = {Nucleic Acids Research},
  year    = {2025},
  volume  = {53},
  number  = {D1},
  pages   = {D564--D574}
}

@article{isert2023structure,
  author  = {Isert, Clemens and Atz, Kenneth and Schneider, Gisbert},
  title   = {{Structure-Based Drug Design with Geometric Deep Learning}},
  journal = {Current Opinion in Structural Biology},
  year    = {2023},
  volume  = {79},
  pages   = {102548}
}

@article{shen2024tacogfn,
  title={Taco{GFN}: Target-conditioned {GF}lowNet for Structure-based Drug Design},
  author={Tony Shen and Seonghwan Seo and Grayson Lee and Mohit Pandey and Jason R Smith and Artem Cherkasov and Woo Youn Kim and Martin Ester},
  journal={Transactions on Machine Learning Research},
  issn={2835-8856},
  year={2024}
}

@article{kang2026pocket,
  title={In-Pocket 3D Graphs Enhance Ligand--Target Compatibility in Generative Small-Molecule Creation: A Dopamine D2 Receptor Model System},
  author={Kang, Seung-gu and Weber, Jeffrey K and Morrone, Joseph A and Zhang, Leili and Huynh, Tien and Cornell, Wendy D},
  journal={The Journal of Physical Chemistry B},
  volume={130},
  number={11},
  pages={2965--2973},
  year={2026}
}

@article{zhu2023pharmacophore,
  title={A pharmacophore-guided deep learning approach for bioactive molecular generation},
  author={Zhu, Huimin and Zhou, Renyi and Cao, Dongsheng and Tang, Jing and Li, Min},
  journal={Nature Communications},
  volume={14},
  number={1},
  pages={6234},
  year={2023}
}

@article{feng2024generation,
  title={Generation of 3D molecules in pockets via a language model},
  author={Feng, Wei and Wang, Lvwei and Lin, Zaiyun and Zhu, Yanhao and Wang, Han and Dong, Jianqiang and Bai, Rong and Wang, Huting and Zhou, Jielong and Peng, Wei and others},
  journal={Nature Machine Intelligence},
  volume={6},
  number={1},
  pages={62--73},
  year={2024}
}

@inproceedings{NEURIPS2021_21b5680d,
  author = {Garcia Satorras, Victor and Hoogeboom, Emiel and Fuchs, Fabian and Posner, Ingmar and Welling, Max},
  booktitle = {Advances in Neural Information Processing Systems},
  pages = {4181--4192},
  title = {E(n) Equivariant Normalizing Flows},
  volume = {34},
  year = {2021}
}

@inproceedings{NIPS2014_f033ed80,
  author = {Goodfellow, Ian J. and Pouget-Abadie, Jean and Mirza, Mehdi and Xu, Bing and Warde-Farley, David and Ozair, Sherjil and Courville, Aaron and Bengio, Yoshua},
  booktitle = {Advances in Neural Information Processing Systems},
  pages = {2672--2680},
  title = {Generative Adversarial Nets},
  volume = {27},
  year = {2014}
}

@article{mirza2014conditional,
  title={Conditional generative adversarial nets},
  author={Mirza, Mehdi and Osindero, Simon},
  journal={arXiv preprint arXiv:1411.1784},
  year={2014}
}

@inproceedings{NEURIPS2019_3001ef25,
  author = {Song, Yang and Ermon, Stefano},
  booktitle = {Advances in Neural Information Processing Systems},
  pages={11895--11907},
  publisher = {Curran Associates, Inc.},
  title = {Generative Modeling by Estimating Gradients of the Data Distribution},
  volume = {32},
  year = {2019}
}

@article{bickerton2012quantifying,
  title   = {Quantifying the Chemical Beauty of Drugs},
  author  = {Bickerton, G. Richard and Paolini, Gaia V. and Besnard, J{\'e}r{\'e}my and Muresan, Sorel and Hopkins, Andrew L.},
  journal = {Nature Chemistry},
  year    = {2012},
  volume  = {4},
  number  = {2},
  pages   = {90--98}
}

@article{ertl2009estimation,
  title={Estimation of synthetic accessibility score of drug-like molecules based on molecular complexity and fragment contributions},
  author={Ertl, Peter and Schuffenhauer, Ansgar},
  journal={Journal of cheminformatics},
  volume={1},
  number={1},
  pages={8},
  year={2009},
  publisher={Springer}
}

@article{kabsch1976solution,
  title={A solution for the best rotation to relate two sets of vectors},
  author={Kabsch, Wolfgang},
  journal={Foundations of Crystallography},
  volume={32},
  number={5},
  pages={922--923},
  year={1976},
  publisher={International Union of Crystallography}
}

@inproceedings{huang2024interaction,
  title={Interaction-based retrieval-augmented diffusion models for protein-specific 3d molecule generation},
  author={Huang, Zhilin and Yang, Ling and Zhou, Xiangxin and Qin, Chujun and Yu, Yijie and Zheng, Xiawu and Zhou, Zikun and Zhang, Wentao and Wang, Yu and Yang, Wenming},
  booktitle={Proceedings of the 41st International Conference on Machine Learning},
  volume={235},
  pages={20348--20364},
  year={2024}
}

@article{forli2016computational,
  title={Computational protein--ligand docking and virtual drug screening with the AutoDock suite},
  author={Forli, Stefano and Huey, Ruth and Pique, Michael E and Sanner, Michel F and Goodsell, David S and Olson, Arthur J},
  journal={Nature protocols},
  volume={11},
  number={5},
  pages={905--919},
  year={2016},
  publisher={Nature Publishing Group UK London}
}

@article{eberhardt2021autodock,
  title={AutoDock Vina 1.2.0: new docking methods, expanded force field, and python bindings},
  author={Eberhardt, Jerome and Santos-Martins, Diogo and Tillack, Andreas F and Forli, Stefano},
  journal={Journal of Chemical Information and Modeling},
  volume={61},
  number={8},
  pages={3891},
  year={2021}
}

@article{alhossary2015fast,
  title={Fast, accurate, and reliable molecular docking with QuickVina 2},
  author={Alhossary, Amr and Handoko, Stephanus Daniel and Mu, Yuguang and Kwoh, Chee-Keong},
  journal={Bioinformatics},
  volume={31},
  number={13},
  pages={2214--2216},
  year={2015},
  publisher={Oxford University Press}
}

@article{weller2024structure,
  title={Structure-based drug design with a deep hierarchical generative model},
  author={Weller, Jesse A and Rohs, Remo},
  journal={Journal of Chemical Information and Modeling},
  volume={64},
  number={16},
  pages={6450},
  year={2024}
}

@article{gao2020generative,
  title={Generative network complex for the automated generation of drug-like molecules},
  author={Gao, Kaifu and Nguyen, Duc Duy and Tu, Meihua and Wei, Guo-Wei},
  journal={Journal of Chemical Information and Modeling},
  volume={60},
  number={12},
  pages={5682--5698},
  year={2020},
  publisher={ACS Publications}
}

@article{winter2019efficient,
  title={Efficient multi-objective molecular optimization in a continuous latent space},
  author={Winter, Robin and Montanari, Floriane and Steffen, Andreas and Briem, Hans and No{\'e}, Frank and Clevert, Djork-Arn{\'e}},
  journal={Chemical Science},
  volume={10},
  number={34},
  pages={8016--8024},
  year={2019},
  publisher={The Royal Society of Chemistry}
}

@article{brown2019guacamol,
  title={GuacaMol: benchmarking models for de novo molecular design},
  author={Brown, Nathan and Fiscato, Marco and Segler, Marwin HS and Vaucher, Alain C},
  journal={Journal of Chemical Information and Modeling},
  volume={59},
  number={3},
  pages={1096--1108},
  year={2019},
  publisher={ACS Publications}
}

@inproceedings{gao2022sample,
  title={Sample efficiency matters: a benchmark for practical molecular optimization},
  author={Gao, Wenhao and Fu, Tianfan and Sun, Jimeng and Coley, Connor},
  booktitle={Advances in neural information processing systems},
  volume={35},
  pages={21342--21357},
  year={2022}
}

@article{buttenschoen2024posebusters,
  title={PoseBusters: AI-based docking methods fail to generate physically valid poses or generalise to novel sequences},
  author={Buttenschoen, Martin and Morris, Garrett M and Deane, Charlotte M},
  journal={Chemical Science},
  volume={15},
  number={9},
  pages={3130--3139},
  year={2024},
  publisher={The Royal Society of Chemistry}
}

@article{harris2023benchmarking,
  title={Benchmarking generated poses: How rational is structure-based drug design with generative models?},
  author={Harris, Charles and Didi, Kieran and Jamasb, Arian R and Joshi, Chaitanya K and Mathis, Simon V and Lio, Pietro and Blundell, Tom},
  journal={arXiv preprint arXiv:2308.07413},
  year={2023}
}

@article{baillif2024benchmarking,
  title={Benchmarking structure-based three-dimensional molecular generative models using GenBench3D: ligand conformation quality matters},
  author={Baillif, Benoit and Cole, Jason and McCabe, Patrick and Bender, Andreas},
  journal={arXiv preprint arXiv:2407.04424},
  year={2024}
}

@article{segler2018generating,
  title={Generating focused molecule libraries for drug discovery with recurrent neural networks},
  author={Segler, Marwin HS and Kogej, Thierry and Tyrchan, Christian and Waller, Mark P},
  journal={ACS central science},
  volume={4},
  number={1},
  pages={120--131},
  year={2018},
  publisher={ACS Publications}
}

@article{blaschke2020reinvent,
  title={REINVENT 2.0: an AI tool for de novo drug design},
  author={Blaschke, Thomas and Ar{\'u}s-Pous, Josep and Chen, Hongming and Margreitter, Christian and Tyrchan, Christian and Engkvist, Ola and Papadopoulos, Kostas and Patronov, Atanas},
  journal={Journal of Chemical Information and Modeling},
  volume={60},
  number={12},
  pages={5918--5922},
  year={2020},
  publisher={ACS Publications}
}

@article{li2022generative,
  title={Generative deep learning enables the discovery of a potent and selective RIPK1 inhibitor},
  author={Li, Yueshan and Zhang, Liting and Wang, Yifei and Zou, Jun and Yang, Ruicheng and Luo, Xinling and Wu, Chengyong and Yang, Wei and Tian, Chenyu and Xu, Haixing and others},
  journal={Nature Communications},
  volume={13},
  number={1},
  pages={6891},
  year={2022},
  publisher={Nature Publishing Group UK London}
}

@article{moret2023leveraging,
  title={Leveraging molecular structure and bioactivity with chemical language models for de novo drug design},
  author={Moret, Michael and Pachon Angona, Irene and Cotos, Leandro and Yan, Shen and Atz, Kenneth and Brunner, Cyrill and Baumgartner, Martin and Grisoni, Francesca and Schneider, Gisbert},
  journal={Nature Communications},
  volume={14},
  number={1},
  pages={114},
  year={2023},
  publisher={Nature Publishing Group UK London}
}

@article{horne2023exploration,
  title={Exploration and exploitation approaches based on generative machine learning to identify potent small molecule inhibitors of $\alpha$-synuclein secondary nucleation},
  author={Horne, Robert I and Murtada, Mhd Hussein and Huo, Donghui and Brotzakis, Z Faidon and Gregory, Rebecca C and Possenti, Andrea and Chia, Sean and Vendruscolo, Michele},
  journal={Journal of chemical theory and computation},
  volume={19},
  number={14},
  pages={4701},
  year={2023}
}

@article{bou2024acegen,
  title={ACEGEN: Reinforcement learning of generative chemical agents for drug discovery},
  author={Bou, Albert and Thomas, Morgan and Dittert, Sebastian and Navarro, Carles and Majewski, Maciej and Wang, Ye and Patel, Shivam and Tresadern, Gary and Ahmad, Mazen and Moens, Vincent and others},
  journal={Journal of Chemical Information and Modeling},
  volume={64},
  number={15},
  pages={5900},
  year={2024}
}

@article{wu2024tamgen,
  title={TamGen: drug design with target-aware molecule generation through a chemical language model},
  author={Wu, Kehan and Xia, Yingce and Deng, Pan and Liu, Renhe and Zhang, Yuan and Guo, Han and Cui, Yumeng and Pei, Qizhi and Wu, Lijun and Xie, Shufang and others},
  journal={Nature Communications},
  volume={15},
  number={1},
  pages={9360},
  year={2024},
  publisher={Nature Publishing Group UK London}
}

@article{thomas2025identification,
  title={Identification of nanomolar adenosine A2A receptor ligands using reinforcement learning and structure-based drug design},
  author={Thomas, Morgan and Matricon, Pierre G and Gillespie, Robert J and Napi{\'o}rkowska, Maja and Neale, Hannah and Mason, Jonathan S and Brown, Jason and Harwood, Kaan and Fieldhouse, Charlotte and Swain, Nigel A and others},
  journal={Nature Communications},
  volume={16},
  number={1},
  pages={5485},
  year={2025},
  publisher={Nature Publishing Group UK London}
}

@article{xie2025accelerating,
  title={Accelerating discovery of bioactive ligands with pharmacophore-informed generative models},
  author={Xie, Weixin and Zhang, Jianhang and Xie, Qin and Gong, Chaojun and Ren, Yuhao and Xie, Jin and Sun, Qi and Xu, Youjun and Lai, Luhua and Pei, Jianfeng},
  journal={Nature Communications},
  volume={16},
  number={1},
  pages={2391},
  year={2025},
  publisher={Nature Publishing Group UK London}
}

@article{persico2026applying,
  title={Applying Deep-Learning-Driven De Novo Design to Hit Identification: A Case Study on A2A Adenosine Receptor Antagonists},
  author={Persico, Margherita and Micoli, Alessandra and Salmaso, Veronica and Cianciulli, Agostino and Moro, Stefano and Spalluto, Giampiero and Buccioni, Michela and Marucci, Gabriella and Volpini, Rosaria and Pozzan, Alfonso and others},
  journal={Journal of Medicinal Chemistry},
  volume={69},
  number={13},
  pages={15403--15422},
  year={2026},
  publisher={ACS Publications}
}

@article{bagal2021molgpt,
  title={MolGPT: molecular generation using a transformer-decoder model},
  author={Bagal, Viraj and Aggarwal, Rishal and Vinod, PK and Priyakumar, U Deva},
  journal={Journal of Chemical Information and Modeling},
  volume={62},
  number={9},
  pages={2064--2076},
  year={2021},
  publisher={ACS Publications}
}

@article{wang2023cmolgpt,
  title={cMolGPT: a conditional generative pre-trained transformer for target-specific de novo molecular generation},
  author={Wang, Ye and Zhao, Honggang and Sciabola, Simone and Wang, Wenlu},
  journal={Molecules},
  volume={28},
  number={11},
  pages={4430},
  year={2023},
  publisher={MDPI}
}

@article{wei2023probabilistic,
  title={Probabilistic generative transformer language models for generative design of molecules},
  author={Wei, Lai and Fu, Nihang and Song, Yuqi and Wang, Qian and Hu, Jianjun},
  journal={Journal of Cheminformatics},
  volume={15},
  number={1},
  pages={88},
  year={2023},
  publisher={Springer}
}

@article{monteiro2023fsm,
  title={FSM-DDTR: End-to-end feedback strategy for multi-objective De Novo drug design using transformers},
  author={Monteiro, Nelson RC and Pereira, Tiago O and Machado, Ana Catarina D and Oliveira, Jos{\'e} L and Abbasi, Maryam and Arrais, Joel P},
  journal={Computers in Biology and Medicine},
  volume={164},
  pages={107285},
  year={2023},
  publisher={Elsevier}
}

@article{yue2024unlocking,
  title={Unlocking comprehensive molecular design across all scenarios with large language model and unordered chemical language},
  author={Yue, Jie and Peng, Bingxin and Chen, Yu and Jin, Jieyu and Zhao, Xinda and Shen, Chao and Ji, Xiangyang and Hsieh, Chang-Yu and Song, Jianfei and Hou, Tingjun and others},
  journal={Chemical Science},
  volume={15},
  number={34},
  pages={13727--13740},
  year={2024},
  publisher={The Royal Society of Chemistry}
}

@article{qian2024consmi,
  title={Consmi: Contrastive learning in the simplified molecular input line entry system helps generate better molecules},
  author={Qian, Ying and Shi, Minghua and Zhang, Qian},
  journal={Molecules},
  volume={29},
  number={2},
  pages={495},
  year={2024},
  publisher={MDPI}
}

@article{guo2021dockstream,
  title={DockStream: a docking wrapper to enhance de novo molecular design},
  author={Guo, Jeff and Janet, Jon Paul and Bauer, Matthias R and Nittinger, Eva and Giblin, Kathryn A and Papadopoulos, Kostas and Voronov, Alexey and Patronov, Atanas and Engkvist, Ola and Margreitter, Christian},
  journal={Journal of cheminformatics},
  volume={13},
  number={1},
  pages={89},
  year={2021},
  publisher={Springer}
}

@article{giblin2026generative,
  title={Generative AI-Assisted Discovery of HPK1 Inhibitors},
  author={Giblin, Kathryn A and Song, Kun and Chen, Hongming and Chen, Weijie and Dong, Zhiqiang and Escobar, Randolph A and Grebe, Tyler P and Grimster, Neil P and Hird, Alexander W and Hughes, Samantha J and others},
  journal={Journal of Medicinal Chemistry},
  volume={69},
  number={15},
  pages={18723--18738},
  year={2026},
  publisher={ACS Publications}
}

@article{quinn2024accelerated,
  title={Accelerated discovery of carbamate Cbl-b inhibitors using generative AI models and structure-based drug design},
  author={Quinn, Taylor R and Giblin, Kathryn A and Thomson, Clare and Boerth, Jeffrey A and Bommakanti, Gayathri and Braybrooke, Erin and Chan, Christina and Chinn, Alex J and Code, Erin and Cui, Caifeng and others},
  journal={Journal of Medicinal Chemistry},
  volume={67},
  number={16},
  pages={14210--14233},
  year={2024},
  publisher={ACS Publications}
}

@article{ash2025practically,
  title={Practically significant method comparison protocols for machine learning in small molecule drug discovery},
  author={Ash, Jeremy R and Wognum, Cas and Rodr{\'\i}guez-P{\'e}rez, Raquel and Aldeghi, Matteo and Cheng, Alan C and Clevert, Djork-Arn{\'e} and Engkvist, Ola and Fang, Cheng and Price, Daniel J and Hughes-Oliver, Jacqueline M and others},
  journal={Journal of Chemical Information and Modeling},
  volume={65},
  number={18},
  pages={9398--9411},
  year={2025},
  publisher={ACS Publications}
}

@article{liu2024good,
  title={How good are current pocket-based 3D generative models?: The benchmark set and evaluation of protein pocket-based 3D molecular generative models},
  author={Liu, Haoyang and Qin, Yifei and Niu, Zhangming and Xu, Mingyuan and Wu, Jiaqiang and Xiao, Xianglu and Lei, Jinping and Ran, Ting and Chen, Hongming},
  journal={Journal of Chemical Information and Modeling},
  volume={64},
  number={24},
  pages={9260--9275},
  year={2024},
  publisher={ACS Publications}
}

@article{krishna2024generalized,
  title={Generalized biomolecular modeling and design with RoseTTAFold All-Atom},
  author={Krishna, Rohith and Wang, Jue and Ahern, Woody and Sturmfels, Pascal and Venkatesh, Preetham and Kalvet, Indrek and Lee, Gyu Rie and Morey-Burrows, Felix S and Anishchenko, Ivan and Humphreys, Ian R and others},
  journal={Science},
  volume={384},
  number={6693},
  pages={eadl2528},
  year={2024},
  publisher={American Association for the Advancement of Science}
}

@article{abramson2024accurate,
  title={Accurate structure prediction of biomolecular interactions with AlphaFold 3},
  author={Abramson, Josh and Adler, Jonas and Dunger, Jack and Evans, Richard and Green, Tim and Pritzel, Alexander and Ronneberger, Olaf and Willmore, Lindsay and Ballard, Andrew J and Bambrick, Joshua and others},
  journal={Nature},
  volume={630},
  number={8016},
  pages={493--500},
  year={2024},
  publisher={Nature Publishing Group UK London}
}

@article{friesner2004glide,
  title={Glide: a new approach for rapid, accurate docking and scoring. 1. Method and assessment of docking accuracy},
  author={Friesner, Richard A and Banks, Jay L and Murphy, Robert B and Halgren, Thomas A and Klicic, Jasna J and Mainz, Daniel T and Repasky, Matthew P and Knoll, Eric H and Shelley, Mee and Perry, Jason K and others},
  journal={Journal of Medicinal Chemistry},
  volume={47},
  number={7},
  pages={1739--1749},
  year={2004},
  publisher={ACS Publications}
}

@article{sebaugh2011guidelines,
  title={Guidelines for accurate EC50/IC50 estimation},
  author={Sebaugh, JL},
  journal={Pharmaceutical statistics},
  volume={10},
  number={2},
  pages={128--134},
  year={2011},
  publisher={Wiley Online Library}
}

@article{maveyraud2020protein,
  title={Protein X-ray crystallography and drug discovery},
  author={Maveyraud, Laurent and Mourey, Lionel},
  journal={Molecules},
  volume={25},
  number={5},
  pages={1030},
  year={2020},
  publisher={MDPI}
}
	\endgroup

\end{document}